\pdfoutput=1

\documentclass[11pt]{article}

\usepackage[]{acl}

\usepackage{times}
\usepackage{latexsym}
\usepackage{bbm}

\usepackage[T1]{fontenc}

\usepackage[utf8]{inputenc}

\usepackage{microtype}

\usepackage{amssymb}
\usepackage{amsmath}
\usepackage{algorithm,algcompatible}
\usepackage{graphicx}
\usepackage{multirow}
\usepackage{rotating}
\usepackage{caption}
\usepackage{subcaption}
\usepackage{todonotes}
\usepackage{bm,empheq}
\newcommand{\vctr}[1]{\bm{#1}}
\newcommand{\mtrx}[1]{\bm{#1}}

\newcommand\blfootnote[1]{%
  \begingroup
  \renewcommand\thefootnote{}\footnote{#1}%
  \addtocounter{footnote}{-1}%
  \endgroup
}

\usepackage{caption}
\title{Dynamic Dialogue Policy for Continual Reinforcement Learning}

\author{Christian Geishauser, Carel van Niekerk, Hsien-Chin Lin, \\ {\bf Nurul Lubis}, {\bf Michael Heck}, {\bf Shutong Feng}, {\bf Milica Gašić} \\
Heinrich Heine University Düsseldorf, Germany \\
         \texttt{\small \{geishaus, niekerk, linh, lubis, heckmi, fengs, gasic\}@hhu.de}}

\begin{document}

\maketitle
\begin{abstract}
Continual learning is one of the key components of human learning and a necessary requirement of artificial intelligence. As dialogue can potentially span infinitely many topics and tasks, a task-oriented dialogue system must have the capability to continually learn, dynamically adapting to new challenges while preserving the knowledge it already acquired. Despite the importance, continual reinforcement learning of the dialogue policy has remained largely unaddressed. The lack of a framework with training protocols, baseline models and suitable metrics, has so far hindered research in this direction. In this work we fill precisely this gap, enabling research in dialogue policy optimisation to go from static to dynamic learning. We provide a continual learning algorithm, baseline architectures and metrics for assessing continual learning models. Moreover, we propose the dynamic dialogue policy transformer (DDPT), a novel dynamic architecture that can integrate new knowledge seamlessly, is capable of handling large state spaces and obtains significant zero-shot performance when being exposed to unseen domains, without any growth in network parameter size. We validate the strengths of DDPT in simulation with two user simulators as well as with humans.


\end{abstract}

\section{Introduction}
\blfootnote{To be published in {\it \ COLING2022, October 12-17, 2022, Gyeongju, Republic of Korea}.}
Task-oriented dialogue systems are characterised by an underlying task or a goal that needs to be achieved during the conversation, such as managing a schedule or finding and booking a restaurant. Modular dialogue systems have a tracking component that maintains information about the dialogue in a belief state, and a planning component that models the underlying policy, i.e., the selection of actions~\cite{Levin97astochastic, probabilistic-roy, POMDP-Williams, zhang2020recent}. The spectrum of what a task-oriented dialogue system can understand and talk about is defined by an ontology. The ontology defines domains such as restaurants or hotels, slots within a domain such as the area or price, and values that a slot can take, such as the area being west and the price being expensive. As dialogue systems become more popular and powerful, they should not be restricted by a static ontology. Instead, they should be dynamic and grow as the ontology grows, allowing them to comprehend new information and talk about new topics -- just like humans do.

In the literature, this is referred to as continual learning \cite{clNLP, clRL, hadsell20}. A learner is typically exposed to a sequence of tasks that have to be learned in a sequential order. When faced with a new task, the learner should leverage its past knowledge (\emph{forward transfer}) and be flexible enough to rapidly learn how to solve the new task (\emph{maintain plasticity}). On the other hand, we must ensure that the learner does not forget how to solve previous tasks while learning the new one (prevent \emph{catastrophic forgetting}).  Rather, a learner should actually improve its behaviour on previous tasks after learning a new task, if possible (\emph{backward transfer}). 


Despite progress in continual learning \cite{lange21, clNN, clNLP, clRL, hadsell20}, there is -- to the best of our knowledge -- no work that addresses continual reinforcement learning (continual RL) of the dialogue policy, even though the policy constitutes a key component of dialogue systems. Research in this direction is hindered by the lack of a framework that provides suitable models, evaluation metrics and training protocols.

In modular task-oriented dialogue systems the input to the dialogue policy can be modelled in many different ways \cite{bbq, acer2, gdpl, dip, feudal1, metapolicy}. An appropriate choice of state representation is key to the success of any form of RL \cite{madureira20}.  In continual RL for the dialogue policy, this choice is even more essential. Different dialogue domains typically share structure and behaviour that should be reflected in the state and action representations. The  architecture needs to exploit such common structure, to the benefit of any algorithm applied to the model. In this work, we therefore centre our attention on this architecture. We contribute~\footnote{\url{https://doi.org/10.5281/zenodo.7075192}}

\begin{itemize}
    \item the first framework for continual RL to optimise the dialogue policy of a task-oriented dialogue system, 
    two baseline architectures, an implementation of the state-of-the-art continual RL algorithm \cite{clear} and continual learning metrics for evaluation based on \citet{powers2021cora}, and
    \item a further, more sophisticated, new continual learning architecture based on the transformer encoder-decoder \cite{vaswani17} and description embeddings, which we call dynamic dialogue policy transformer (DDPT). Our architecture can seamlessly integrate new information,
    has significant zero-shot performance and can cope with large state spaces that naturally arise from a growing number of domains
    while maintaining a  fixed number of network parameters.
\end{itemize}

\section{Related Work}
\subsection{Continual Learning in Task-oriented Dialogue Systems}
Despite progress in continual learning, task-oriented dialogue systems have been barely touched by the topic. \citet{lee18} proposed a task-independent neural architecture with an action selector. The action selector is a ranking model that calculates similarity between state and candidate actions. Other works concentrated on dialogue state tracking \cite{wu19} or natural language generation \cite{mi20, geng21}. \citet{geng21} proposed a network pruning and expanding strategy for natural language generation. \citet{madotto20} introduced an architecture called AdapterCL and trained it in a supervised fashion for intent prediction, state tracking, generation and end-to-end learning. However, that work focused on preventing catastrophic forgetting and did not address the dialogue policy. As opposed to the above-mentioned approaches, we consider continual RL to optimise a dialogue policy.

\subsection{Dialogue Policy State Representation}
In the absence of works that directly address continual learning for the dialogue policy, it is worth looking at approaches that allow dialogue policy adaptation to new domains and examining them in the context of continual learning requirements.

The first group among these methods introduces new parameters to the model when the domain of operation changes. The approaches directly vectorise the belief state, hence the size of the input vector depends on the domain (as different domains for instance have different numbers of slots) \cite{su16, bbq, acer2, gdpl, convlab2}. 
In the context of continual learning 
such approaches would likely preserve the plasticity of the underlying RL algorithm but would score poorly on forward and backward transfer. 

Another group of methods utilises a hand-coded domain-independent feature set that allows the policy to be transferred to different domains \cite{dip, feudal1, chen18, strac, lin-etal-2021-domain}. This is certainly more promising for continual learning, especially if the requirement is to keep the number of parameters bounded. However, while such models might score well on forward and backward transfer, it is possible that the plasticity of the underlying RL algorithm is degraded. Moreover, developing such features requires manual work and it is unclear if they would be adequate for any domain.


\citet{metapolicy} go a step further in that direction. They propose the usage of embeddings for domains, intents, slots and values in order to allow cross-domain transfer. To deal with the problem of a growing state space with an increased number of domains, they propose a simple averaging mechanism. However, as the number of domains becomes larger, averaging will likely result in information loss. Moreover, their architecture still largely depends on predefined feature categories.

A third option is to exploit similarities between different domains while learning about a new domain. \citet{gasic-committee-2015} use a committee of Gaussian processes together with designed kernel functions in order to define these similarities and therefore allow domain extension and training on new domains. A similarity-based approach could in principle score well on all three continual learning measures. However, it is desirable to minimise the amount of manual work needed to facilitate continual learning. 

\subsection{Dialogue Policy Action Prediction}

In the realm of domain adaptation, works assume a fixed number of actions that are slot-independent, and focus on the inclusion of slot-dependent actions when the domain changes \cite{dip, feudal1, chen18, strac, lin-etal-2021-domain}. This allows seamless addition of new slots, but the integration of new intents or slot-independent actions requires an expansion of the model. 

Works that allow new actions to be added to the action set compare the encoded state and action embeddings with each other \cite{lee18, metapolicy, vlasov19}, suggesting that exploiting similarities is key not only for state representations but also for action prediction.

With multi-domain dialogues it becomes necessary to be able to produce more than one action in a turn, which is why researchers started to use recurrent neural network (RNN) models to produce a sequence of actions in a single turn \cite{shu19, zhang20}. RNNs are known however to only provide a limited context dependency.


\begin{center}
\begin{figure}[t]
\includegraphics[trim=0cm 0cm 0.0cm 0.0cm, width=0.5\textwidth]{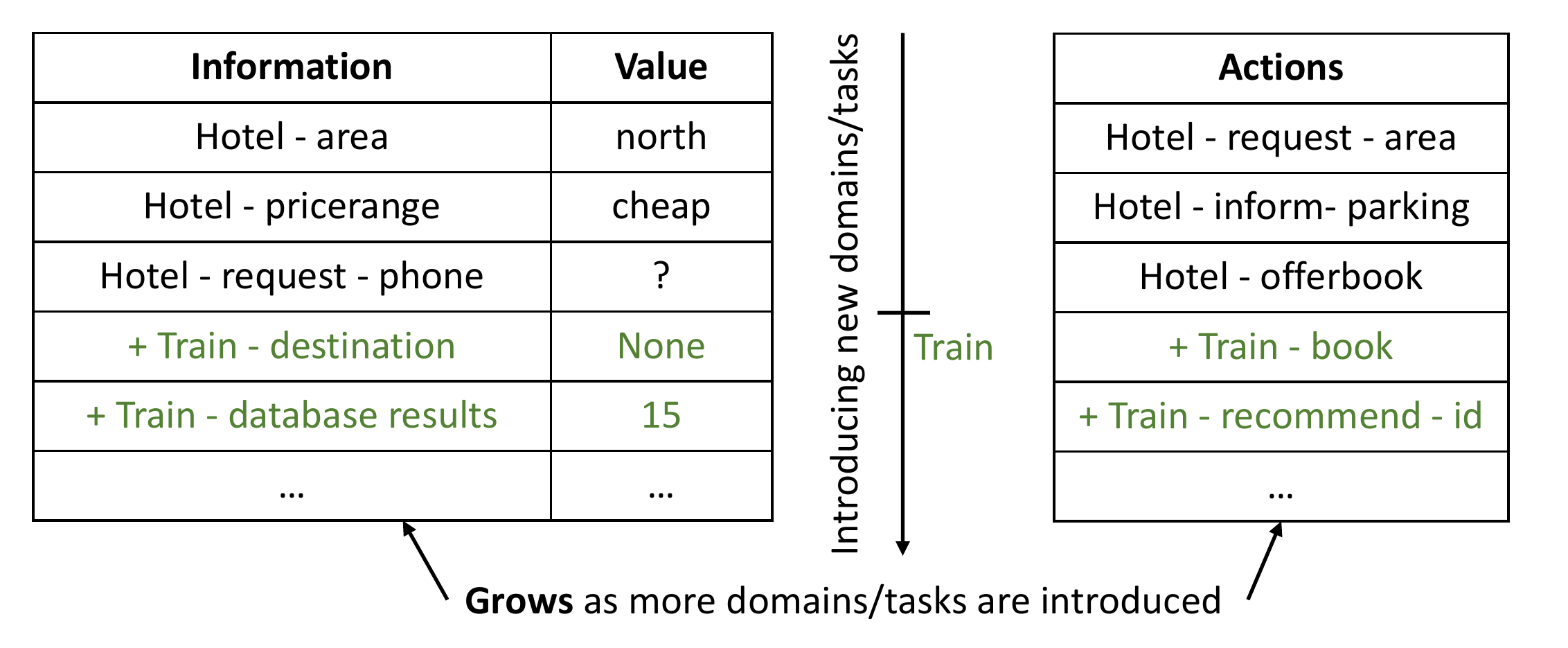}
\caption{The amount of information that the dialogue agent must comprehend and the possible actions it can take increases as new domains/tasks are introduced.}
\label{state-action-growth}
\end{figure}
\end{center}

\begin{figure*}[h!]
    \centering
    \small
    \begin{tabular}{ccc}
        \includegraphics[scale=0.41]{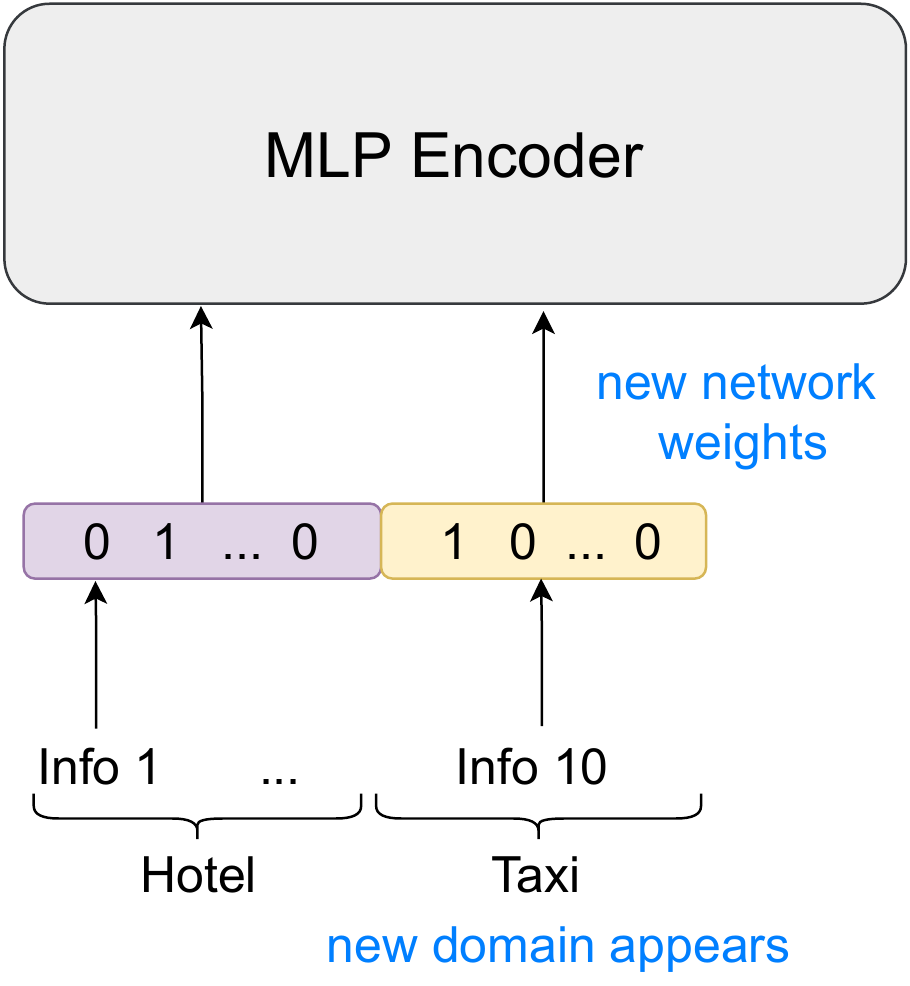} &   \includegraphics[scale=0.41]{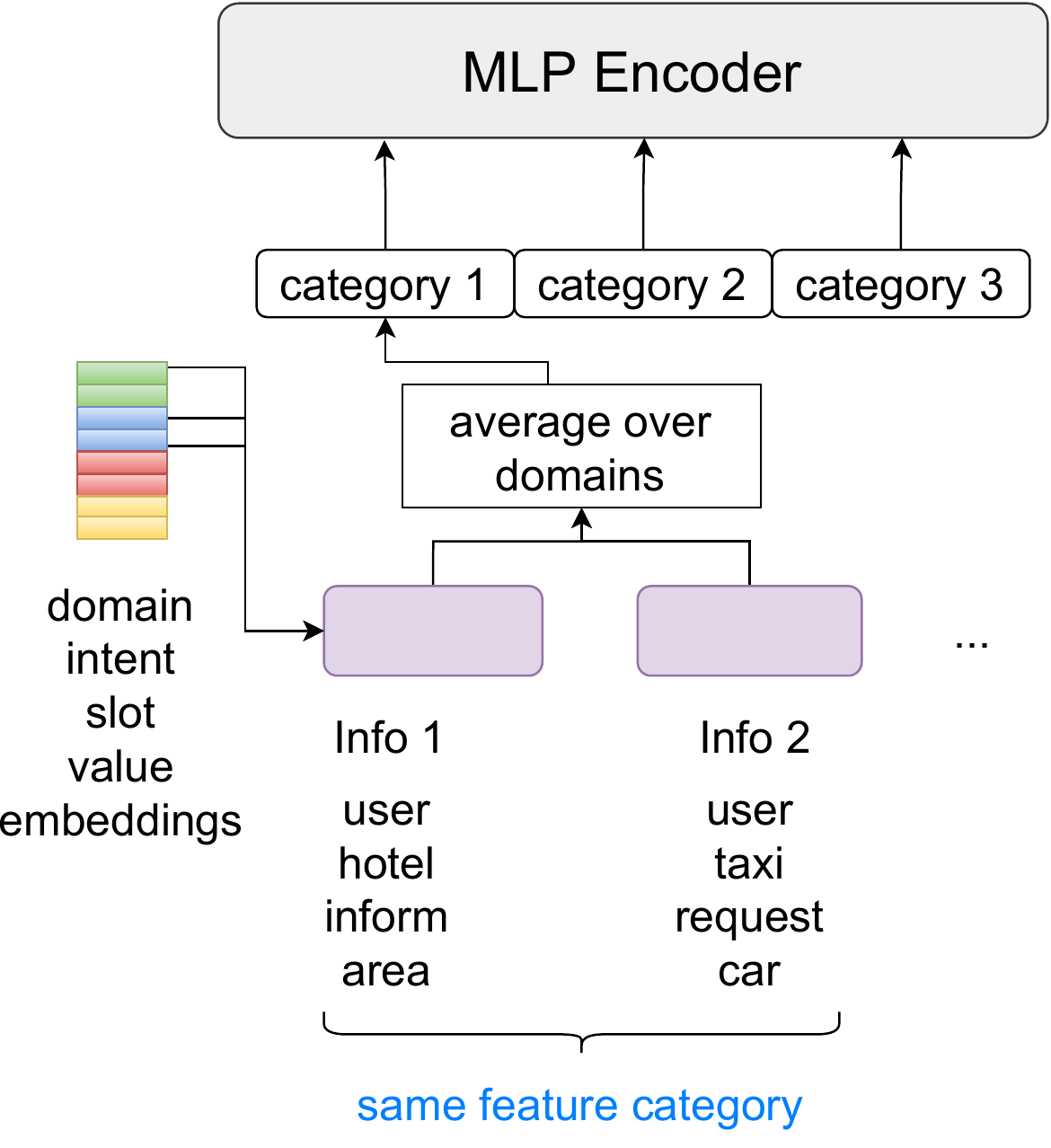} & \includegraphics[scale=0.41]{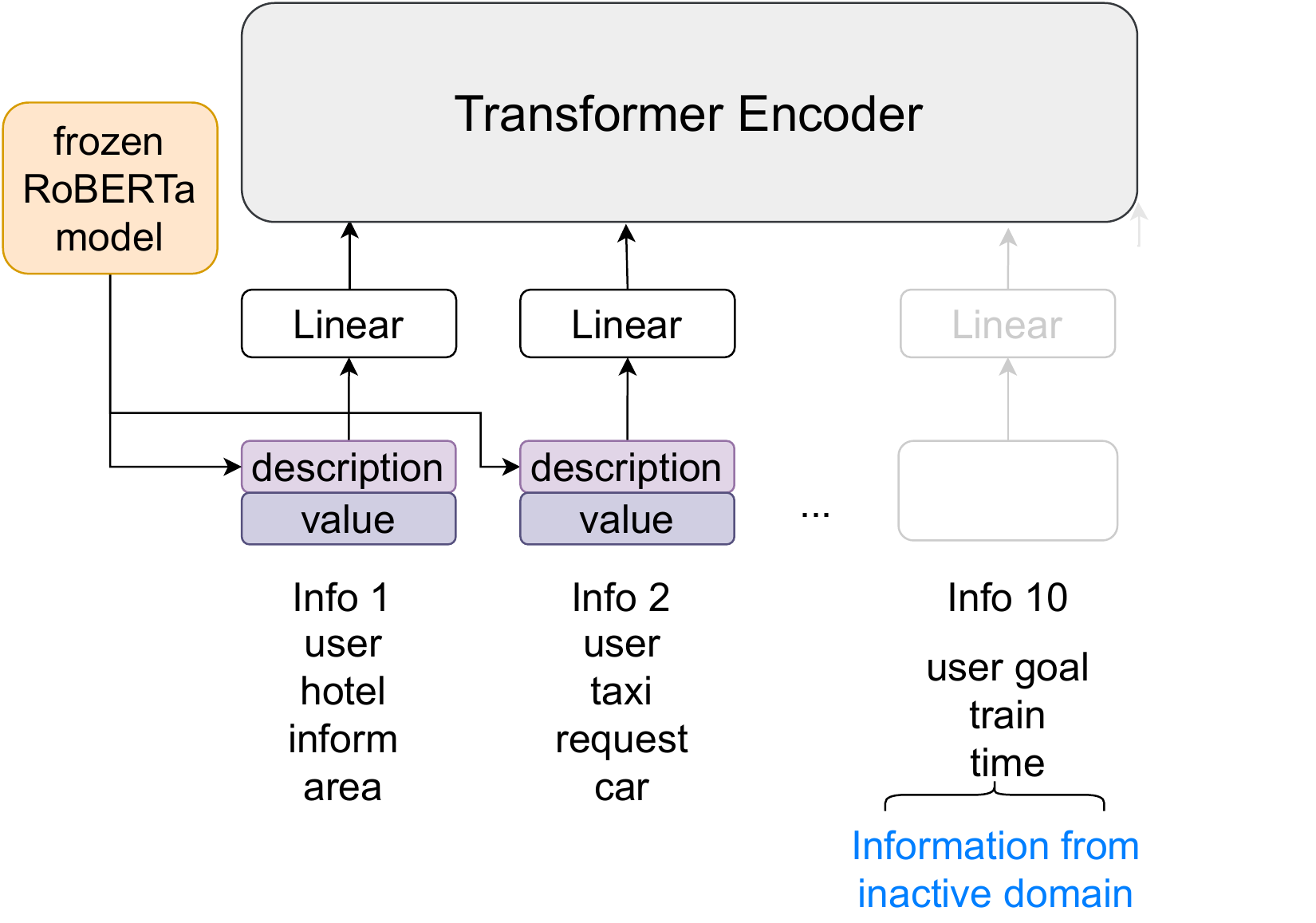} \\
        (a) Binary representation (Bin) & (b) Semantic features (Sem) & (c) Descriptions and values (DDPT)
    \end{tabular}
\caption{State representation for different architectures. (a) \emph{Bin} uses a flattened dialogue state with binary features, where the input size grows and new network weights need to be added when facing a new domain. (b) \emph{Sem} uses the idea from \citet{metapolicy}, using trainable embeddings for domain, intent, slot and value. The information corresponding to a specific feature category is then averaged over domains in order to be independent on the number of domains. (c) Our proposed DDPT model uses descriptions for every information which are embedded using a pretrained language model. The embedded description together with a value for the information is then fed into a linear layer and a transformer encoder.}
\label{state_representation}
\end{figure*}

\section{Background}

\subsection{Continual Reinforcement Learning}
In typical RL scenarios, an agent interacts with a stationary MDP $\mathcal{M}= \langle \mathcal{S}, \mathcal{A}, p, p_0, r \rangle$, where $\mathcal{S}$ and $\mathcal{A}$ constitute the state and action space of the agent, $p(s^\prime | s, a)$ models the probability of transitioning to state $s^\prime$ after executing action $a$ in state $s$, and $p_0(s)$ is the probability of starting in state $s$. The reward function $r$ defines the observed reward in every time-step. The goal is to maximise the cumulative sum of rewards in that MDP.

In contrast, continual reinforcement learning focuses on non-stationary or changing environments \cite{hadsell20}. Generally speaking, the agent faces a sequence of Markov decision processes $\{\mathcal{M}_z\}_{z=1}^\infty$ \cite{lecarpentier19, chandak20, towardsCRL} with possibly different transition dynamics, reward functions or even state or action spaces. The variable $z$ is often referred to as a task (or context) \cite{osaka20, sequoia21}. While the MDP can change from episode to episode, it is often assumed that the agent is exposed to a fixed MDP for a number of episodes and then switches to the next MDP. Once a new task (or MDP) is observed, the old task is either never observed again or only periodically \cite{clear, powers2021cora}. The goal is to retain performance on all seen tasks. This requires the model to prevent catastrophic forgetting of old tasks while at the same time adapting to new tasks.

A state-of-the art method for continual RL that uses a replay memory is CLEAR \cite{clear}. CLEAR manages the trade-off between preventing catastrophic forgetting and fast adaptation through an on-policy update step as well as an off-policy update step. The on-policy step is supposed to adapt the policy to the recent task by using the most recent dialogues while the off-policy step should lead to retaining performance on old tasks by updating on old experiences from the replay buffer. The off-policy update is further regularized such that policy and critic outputs are close to the historical prediction. More information on CLEAR is provided in the Appendix \ref{clear-background}.

In the context of dialogue, a task usually refers to a domain as defined in \citet{madotto20} and we will use these two terms interchangeably. As an example setting, a dialogue system is tasked with fulfilling user goals concerning hotel information and booking and after some amount of time with fulfilling goals related to train bookings. In terms of MDPs, the dialogue system first faces the MDP $\mathcal{M}_{z_1}, z_1 = \text{hotel}$, for some amount of dialogues and afterwards $\mathcal{M}_{z_2}, z_2 = \text{train}$. Once the train domain is introduced, the state and action space grows (as a result of the growing ontology) as depicted exemplarily in Figure \ref{state-action-growth}. As a consequence, the model needs to understand new topics such as the destination of the train and select new actions such as booking a train. In addition, the probability distributions $p$ and $p_0$
of $\mathcal{M}_{z_2}$ are different compared to $\mathcal{M}_{z_1}$ since the probability that the user talks about hotels should be close to $0$ while the probability that the agent's states contain information related to trains is close to $1.0$.

\subsection{Dialogue Policy in Modular Systems}

In modular task-oriented dialogue systems, the decision of a dialogue policy is commonly based on the hidden information state of the dialogue system. This hidden information state, according to \citet{his-model}, should consist of the following information: the predicted user action, the predicted user goal and a representation of the dialogue history. For reactive behaviour by the policy, the user action is important as it includes information related to requests made by the user. The predicted user goal summarises the current goal of the user, including specified constraints. Lastly, the dialogue history representation captures the relevant information mentioned in the dialogue history, such as the latest system action. The state can also include the likelihood of the predicted acts, goal and dialogue history in the form of confidence scores.
Moreover, the state often contains information about the database, for instance the number of entities that are available given the current predicted user goal. 

Each domain that the system can talk about is either active, meaning that it has already been mentioned by the user, or inactive. The active domains can be derived from the user acts, from the user goal or tracked directly \cite{carel21}.

Finally, the policy is supposed to take actions. As in \cite{shu19, zhang20}, each action can be represented as a sequence of tuples $(\mathit{domain}, \mathit{intent}, \mathit{slot})$. For instance, an action could be that the system requests the desired arrival time of a train or asks for executing a payment.



\label{background-dialogue-policy-section}

\section{Dynamic Dialogue Policy Transformer}

Our goal is to build a model that can talk about a potentially very large number of domains and is able to deal with new domains and domain extensions seamlessly without requiring any architectural changes. In particular, the number of model parameters should remain fixed.
This is challenging since new domains require understanding of previously unseen information and the ability to talk about new topics. 

\begin{center}
\begin{figure}[t]
\includegraphics[trim=0cm 0cm 0.0cm 0.0cm, width=0.48\textwidth]{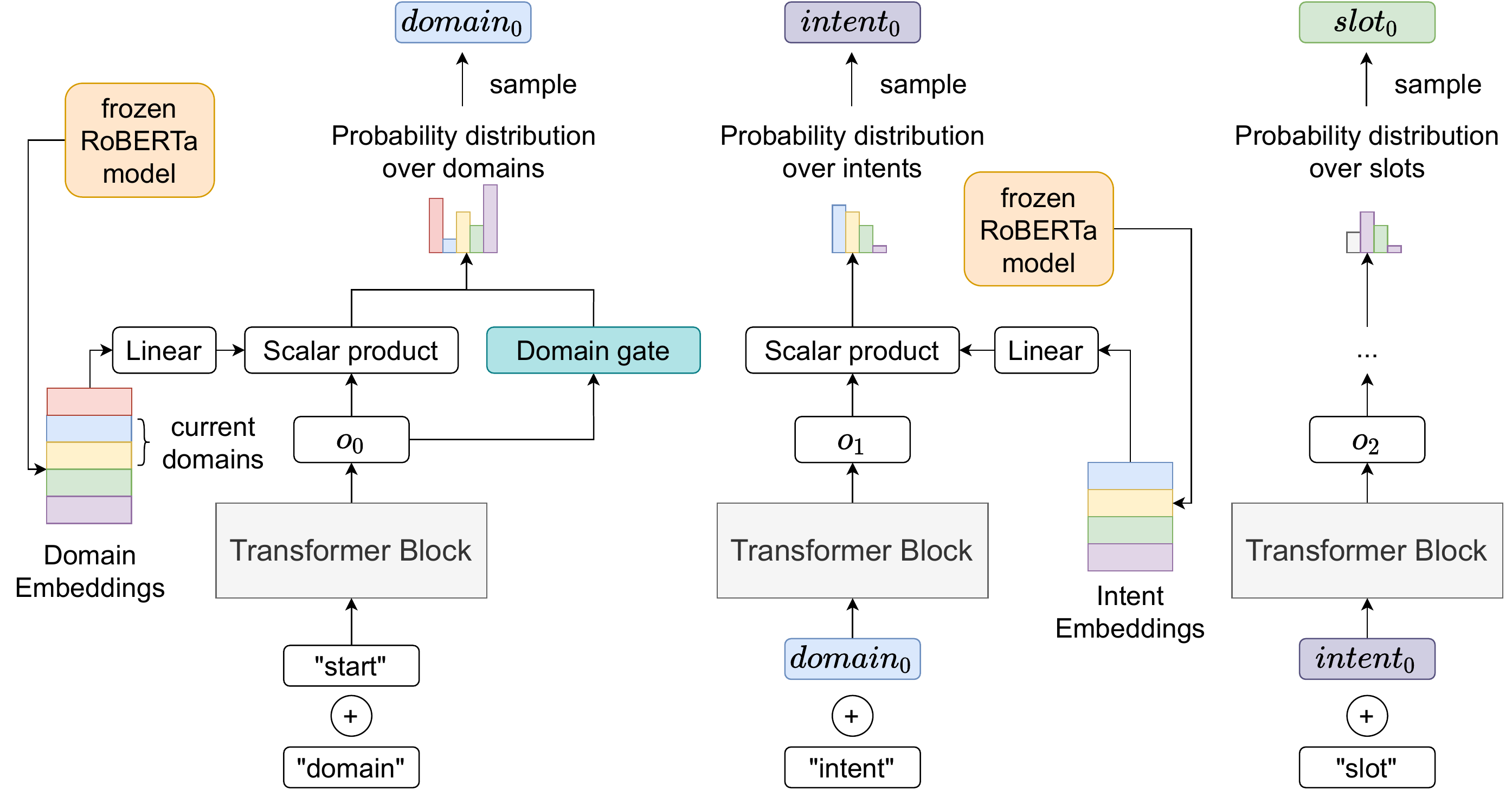}
\caption{Proposed action prediction in DDPT using a transformer decoder. In every decoding step, a token embedding for domain, intent or slot informs the model what needs to be predicted and the previous output is fed into the decoder. In case of domain prediction, we propose a domain gate that decides whether to choose a domain that the user currently talks about.}
\label{action-decoding}
\end{figure}
\end{center}

Our approach is inspired by the way an employee would explain and act upon a novel task: 1) describe the information that can be used and the actions that can be taken in natural language, 2) restrict the focus to the information that is important for solving the task at hand, 3) when an action needs to be taken, this action is based on the information that was attended to (e.g. for the action to request the area, one would put attention on the information whether the area is already given). We propose an architecture that uses the transformer encoder with information embeddings (Section \ref{state-representation-section} and Figure \ref{state_representation}(c)) to fulfill 1) and 2) and the transformer decoder that leverages the domain gate (Section \ref{action-prediction}, \ref{domain-gate} and Figure \ref{action-decoding}) to fulfill 3), which we call \emph{dynamic dialogue policy transformer} (DDPT).


\subsection{State Representation} \label{state-representation-section}

Recall from Section~\ref{background-dialogue-policy-section} that the agent is provided with information on various concepts $f$ for domain $d_f$:  the user goal (domain-slot pairs), the user action (intents) and the dialogue history (system intents and database results). We assume that the agent has access to an external dictionary providing a natural language description $\mathit{descr}_f$ of each of these, e.g.\ ``area of the hotel'' or ``number of hotel database results'', which is common in dialogue state tracking \cite{rastogi20,  carel21, leeDST21}. 
See Appendix \ref{appendix-descriptions} for the full list of descriptions. 
During a dialogue, the dialogue state or belief tracker assigns numerical values $v_f$, e.g.\ confidence scores for user goals or the number of data base results, etc. 
For every concept $f$ we define the information embedding
%
\begin{equation*}
    \vctr{e}_{\text{info}_f} = \mathrm{Lin} \left( \left[ \overline{\mathrm{LM}}(\mathit{descr}_f), \mathrm{Lin}(v_f) \right] \right) \in \mathbb{R}^h
\end{equation*}
where $\overline{\mathrm{LM}}$ denotes applying a language model such as RoBERTa \cite{roberta} and averaging of the token embeddings, and $\mathrm{Lin}$ denotes a linear layer. $\vctr{e}_{\text{info}_f}$ represents information in a high-dimensional vector space. Intuitively, every information can be thought of as a node in a graph.
The list of information embeddings are the input to a transformer encoder \cite{vaswani17}. The attention mechanism allows the agent to decide for every information embedding $\vctr{e}_{\text{info}_f}$ on which other embeddings $\vctr{e}_{\text{info}_g}$ it can put its attention. With a growing number of domains that the system can talk about, the number of information embeddings will increase, making it more difficult to handle the growing state space. However, we observe that only information that is related to active domains is important at the current point in time. Therefore, we prohibit the information embeddings from attending to information that is related to inactive domains in order to avoid the issue of growing state spaces. While the actual state space may be extremely large due to hundreds of domains, the effective state space remains small, making it possible to handle a very large number of domains. Our proposed state encoder is depicted in Figure \ref{state_representation}(c).

In this way, the state representation meets the following demands: 1) new concepts can be understood and incorporated seamlessly into the state without a growth in network parameters, as long as they are descriptive; 2) the description embeddings from a language model allow forward transfer by exploiting similarities and common structure among tasks; 3) the value $v_f$ allows numerical information such as confidence scores or other measures of model uncertainty to be included; 4) the state space will not be unreasonably large as information for inactive domains is masked.

\subsection{Action Prediction} \label{action-prediction}

Similar to existing work \cite{shu19, zhang20} we separately predict domains, intents and slots for action prediction. We define a domain set $\mathcal{D}$, intent set $\mathcal{I}$ and slot set $\mathcal{S}$ as follows. The domain set $\mathcal{D}$ consists of all domains the model has seen so far plus an additional \textit{stop} domain. The intent set $\mathcal{I}$ and slot set  $\mathcal{S}$ consist of all intents and slots we can use for actions, respectively. Every domain, intent and slot has an embedding vector, which we obtain by feeding the token of the domain, intent or slot into our pretrained language model. The embedding vectors are then fed into a linear layer that produces vectors of size $\mathbb{R}^h$. We thus obtain domain, intent and slot embeddings $\vctr{b}^d$~$\forall d \in \mathcal{D}$, $\vctr{b}^i$~$\forall i \in \mathcal{I}$, and $\vctr{b}^s$~$\forall s \in \mathcal{S}$.

The policy first chooses a domain. Then, based on the domain, it picks an intent from the list of intents that are possible for that domain. Lastly, it picks an adequate slot from the set of possible slots for that domain and intent. This process repeats until the policy selects the \textit{stop} domain. This will lead to a sequence $(\mathit{domain}_m, \mathit{intent}_m, \mathit{slot}_m)_{m=0}^n$. We leverage a transformer decoder \cite{vaswani17}, the aforementioned embeddings for domains, intents and slots and similarity matching to produce the sequence. In every decoding step $t$ the input to the transformer is $\vctr{b}_{t-1} + \vctr{l}_t$, where $\vctr{b}_{t-1}$ is the embedding of the previous prediction and $\vctr{l}_t$ is a token embedding for token \textit{domain}, \textit{intent} or \textit{slot} that indicates what needs to be predicted in turn $t$. $\vctr{b}_{-1}$ is an embedding of a \textit{start} token. 

If we need to predict a domain in step $t$, we calculate the scalar product between the decoder output vector $\vctr{o}_t$ and the different domain embeddings $\vctr{b}^d$ and apply the softmax function to obtain a probability distribution $\mathrm{softmax}[\vctr{o}_t \odot \vctr{b}^d, d \in \mathcal{D}]$ over domains from which we can sample. Intent and slot prediction is analogous. In order to guarantee exploration during training and variability during evaluation, we sample from the distributions. While it is important to explore domains during training, during evaluation the domain to choose should be clear. We hence take the domain with the highest probability during evaluation.

As in the state representation, the embeddings using a pretrained language model allow understanding of new concepts (such as a new intent) immediately, which facilitates zero-shot performance. 
We do not fine-tune any embedding that is produced by the language model.

\subsection{Domain Gate} \label{domain-gate}
If the policy is exposed to a new unseen domain, the most important point to obtain any zero-shot performance is that the policy predicts the correct domain to talk about. If we only use similarity matching of domain embeddings, the policy will likely predict domains it already knows. In dialogue state tracking we often observe that similarity matching approaches predict values they already know when faced with new unseen values, which leads to poor zero-shot generalisation~\cite{rastogi2018multi}. To circumvent that, we propose the domain gate. Let $\mathcal{D}_{\text{curr}}$ be the set of domains that the user talks about in the current turn. In every decoding step $t$ where a domain needs to be predicted, the domain gate obtains $\vctr{o}_t$ as input and predicts the probability $p_{\text{curr}}$ of using a domain from $\mathcal{D}_{\text{curr}}$. When the policy needs to predict a domain in step $t$, it now uses the probability distribution given by $p_{\text{curr}} \cdot \mathrm{softmax}[\vctr{o}_t \odot \vctr{b}_d, d \in \mathcal{D}_{\text{curr}}] + (1 - p_{\text{curr}}) \cdot \mathrm{softmax}[\vctr{o}_t \odot \vctr{b}_d, d \not\in \mathcal{D}_{\text{curr}}]$.

In this process, the policy does not have to predict the new domain immediately but can abstractly first decide whether it wants to use a domain that the user talks about at the moment. 
The decoding process is depicted in Figure \ref{action-decoding}.


\section{Experimental Setup}

\subsection{Metrics}\label{metrics-section}
We follow the setup recently proposed by \citet{powers2021cora}, which assumes that our $N$ tasks/domains $z_1, ..., z_N$ are represented sequentially and each task $z_i$ is assigned a budget $k_{z_i}$. We can cycle through the tasks $M$ times, leading to a sequence of tasks $x_1, ..., x_{N \cdot M}$.
The cycling over tasks defines a more realistic setting than only seeing a task once in the agent's lifetime, in particular in dialogue systems where new domains are introduced but rarely removed. 
\\
\textbf{Continual evaluation}: We evaluate performance on all tasks periodically during training. We show the performance for every domain separately to have an in-depth evaluation and the average performance over domains for an overall trend whether the approaches continually improve.
\\
\textbf{Forgetting}: We follow the definition proposed by \citet{chaudhry18} and \citet{powers2021cora}. Let $m_{i, k}$ be a metric achieved on task $z_i$ after training on task $x_k$, such as the average return or the average dialogue success. For seeds $s$, tasks $z_i$ and $x_j$, where $i < j$, we define
    \begin{equation}
        \mathcal{F}_{i,j} = \frac{1}{s} \sum_s \max_{k \in [0, j - 1]} \{ m_{i,k} - m_{i,j} \}.
    \end{equation}
$\mathcal{F}_{i,j}$ compares the maximum performance achieved on task $z_i$ before training on task $x_j$ to the performance for $z_i$ after training on task $x_j$. If $\mathcal{F}_{i,j}$ is positive, the agent has become worse at past task $z_i$ after training on task $x_j$, indicating forgetting. When $\mathcal{F}_{i,j}$ is negative, the agent has become better at task $z_i$, indicating backward transfer. 
We define $\mathcal{F}_{i}$ as the average over the $\mathcal{F}_{i,j}$ and $\mathcal{F}$ as the average over $\mathcal{F}_{i}$.
\\
\textbf{(Zero-Shot) Forward transfer}: For seeds $s$, tasks $z_i$ and $z_j$, where $j < i$, we define 
    \begin{equation}
        \mathcal{Z}_{i,j} = \frac{1}{s} \sum_s m_{i,j}.
    \end{equation}
We do not substract initial performance as in \citet{powers2021cora} as we are interested in the absolute performance telling us how well we do on task $z_i$ after training on a task $z_j$. We define $\mathcal{Z}_{i}$ as the average over the $\mathcal{Z}_{i,j}$ and $\mathcal{Z}$ as the average over $\mathcal{Z}_{i}$.


\subsection{Baselines} \label{baselines}
We implemented two baselines in order to compare against our proposed DDPT architecture. We do not include a baseline based on expert-defined domain-independent features \cite{dip} as this requires a significant amount of hand-coding and suffers from scalabilility issues. 

\begin{figure*}[t!]
  \begin{subfigure}[]{0.33\textwidth}
    \includegraphics[scale=0.35]{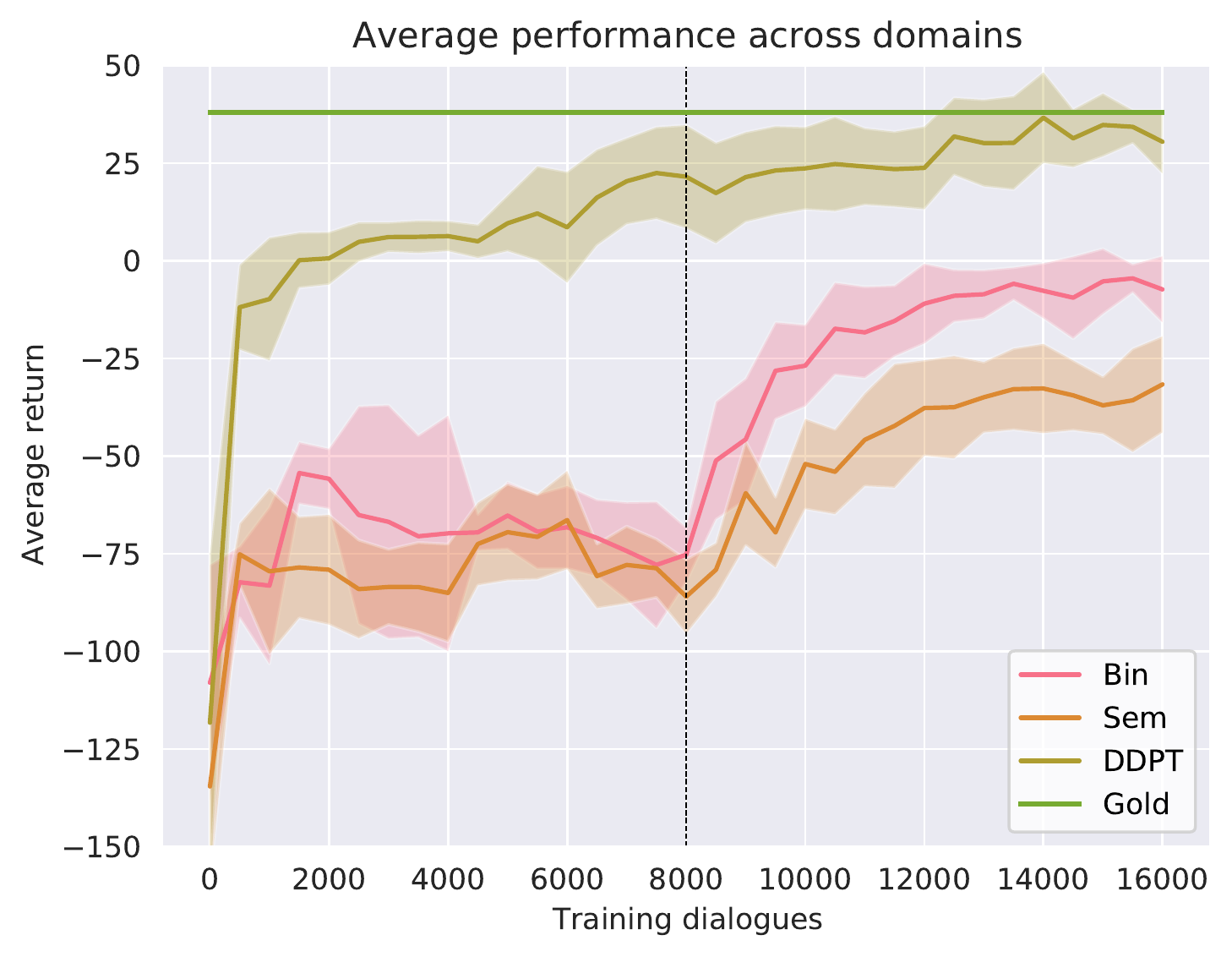}
    \caption{easy-to-hard}
  \end{subfigure}
  \begin{subfigure}[]{0.33\textwidth}
    \includegraphics[scale=0.35]{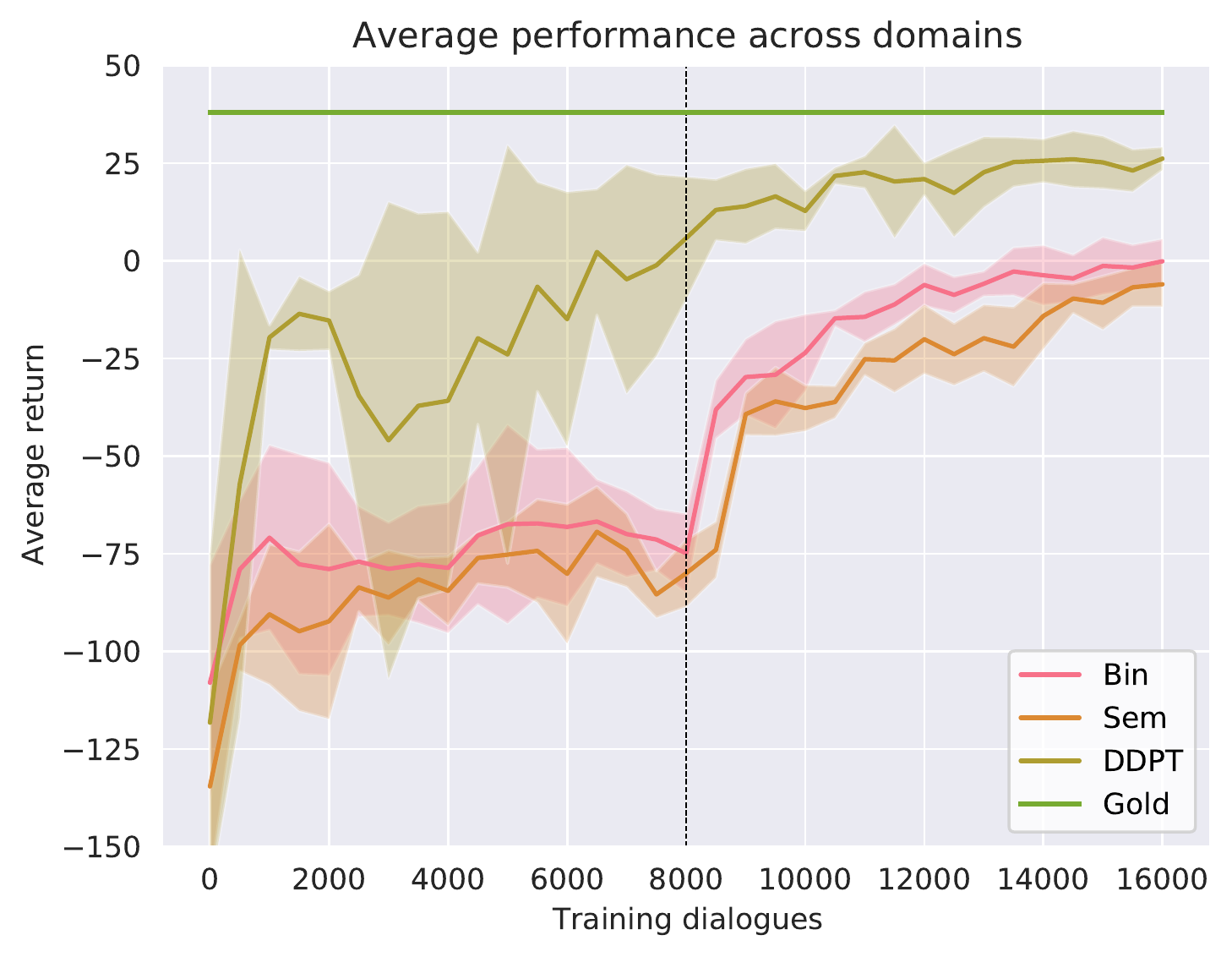}
    \caption{hard-to-easy}
  \end{subfigure}
  \begin{subfigure}[]{0.30\textwidth}
    \includegraphics[scale=0.35]{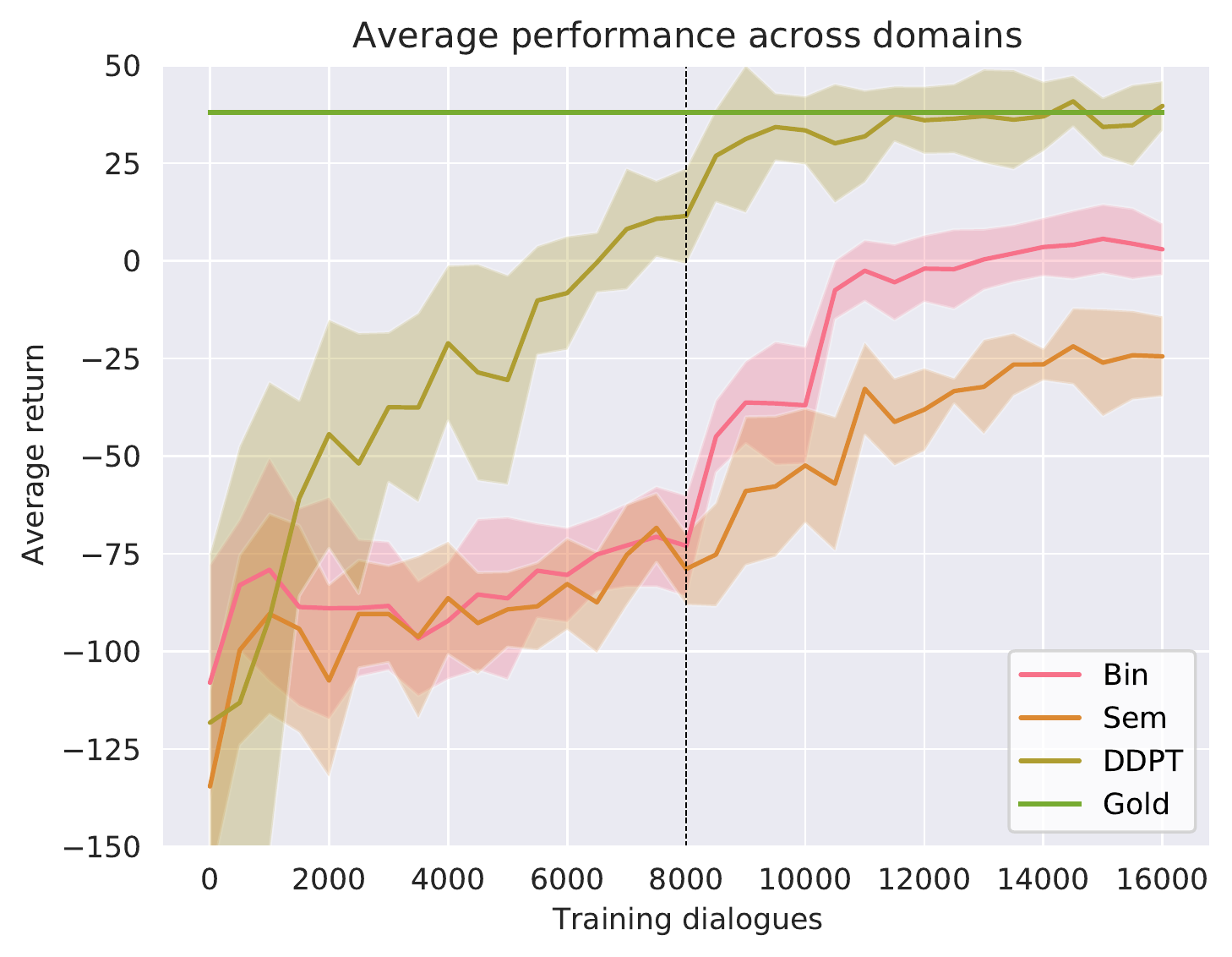}
    \caption{mixed}
  \end{subfigure}

\caption{Training Bin, Sem and DDPT (ours) using CLEAR on three different domain orders, each with 5 different seeds, by interacting with the rule-based user simulator. Each model is evaluated every 500 training dialogues on 100 dialogues per domain. The plots show the average return, where performance is averaged over domains. The vertical line at 8000 dialogues indicates the start of cycle 2. The shaded area represents standard deviation. Gold serves as an upper bound.}
\label{domains-averaged-results}
\end{figure*}

\subsubsection{Baseline State Representations}
We will abbreviate the following baselines with \textbf{Bin} and \textbf{Sem} that indicate their characteristic way of state representation.

\textbf{Bin:} The first baseline uses a flattened dialogue state for the state representation with \emph{binary} values for every information which is the most common way \cite{gdpl, convlab2, acer2}. If a new domain $d$ appears, the input vector must be enlarged in order to incorporate the information from $d$ and new network parameters need to be initialised. The state encoding can be seen in Figure \ref{state_representation}(a). This baseline serves as a representative of methods where new domains necessitate additional parameters.

\textbf{Sem:} The second baseline implements the idea from \citet{metapolicy}, which uses trainable embeddings for domains, intents, slots and values that can capture \emph{semantic} meaning and allow cross-domain transfer.
Using trainable embeddings, one representation is calculated for every feature in every feature category (such as user-act, user goal, etc.) in every domain. The feature representations in a category are then averaged over domains to obtain a final representation. More information can be found in Appendix \ref{appendix-baselines}. This baseline serves as a representative of methods where feature representations remain fixed.

\subsubsection{Action Prediction for Baselines}
Unlike DDPT, which uses a transformer for action prediction, the baselines Bin and Sem use an RNN model for action prediction \cite{shu19, zhang20}. This model uses the decoding process explained in Section \ref{action-prediction} with the exception that the baselines use trainable embeddings for domain, intent and slot (randomly initialised) instead of using embeddings from a pretrained language model as DDPT does. Moreover, they do not use the proposed domain gate.



\subsection{Setup}
We use ConvLab-2 \cite{convlab2} as the backbone of our implementation. We take five different tasks from the MultiWOZ dataset \cite{multiwoz2} which are hotel, restaurant, train, taxi and attraction. Hotel, restaurant and train are more difficult compared to attraction and taxi as they require the agent to do bookings in addition to providing information about requested slots.
We exclude police and hospital from the task list as they are trivial. We use the rule-based dialogue state tracker and the rule-based user simulator provided in ConvLab-2 \cite{convlab2} to conduct our experiments. Typically, the reward provided is $-1$ in every turn to encourage efficiency, and a reward of $80$ or $-40$ for dialogue success or failure. A dialogue is successful if the system provided the requested information to the user and booked the correct entities (if possible). We stick to the above reward formulation with one exception: Instead of the turn level reward of $-1$, we propose to use information overload \cite{information-overload1}. The reason is that dialogue policies tend to over-generate actions, especially if they are trained from scratch. While the user simulator ignores the unnecessary actions, real humans do not. We define information overload for an action $(\mathit{domain}_m, \mathit{intent}_m, \mathit{slot}_m)_{m=1}^n$ as $r_{io} = -\rho \cdot n$, where $\rho \in \mathbb{N}$ defines the degree of the penalty. Information overload generalizes the reward of $-1$ in single action scenarios. We use $\rho=3$ in the experiments.

We train each of the three architectures using CLEAR \cite{clear}. We set the replay buffer capacity to 5000 dialogues and use reservoir sampling \cite{isele18} when the buffer is full. We assign a budget of 2000 dialogues to restaurant, hotel and train and 1000 to attraction and taxi and cycle through these tasks two times, resulting in 16000 training dialogues in total. Since task ordering is still an open area of research \cite{minqi20}, we test three different permutations so that our results do not depend on a specific order. The domain orders we use are 1) \emph{easy-to-hard}: attraction, taxi, train, restaurant, hotel 2) \emph{hard-to-easy}: hotel, restaurant, train, taxi, attraction and 3) \emph{mixed}: restaurant, attraction, hotel, taxi, train.

\section{Results}

\subsection{Continual Evaluation}\label{continual-eval-section}

We show performance in terms of average return for all three task orders in Figure \ref{domains-averaged-results}(a)-(c). The plots show the performance averaged over domains. We refer to Appendix \ref{appendix-ce} for in-depth evaluations for each individual domain. The horizontal line \textbf{Gold} denotes an upper limit for the models that was obtained by training a Bin model separately on each domain until convergence. We can observe that DDPT outperforms the baselines regardless of task order, almost reaching the upper bound. We will see in Section \ref{ft-fg} that the baselines suffer more from forgetting compared to DDPT, such that training on a new domain reduces performance on previous domains. We suspect that this contributes to the lower final performance of the baselines. Moreover, we can observe that the final performance of DDPT barely depends on a specific task order. Nevertheless, we can see that training starts off faster in easy-to-hard order, which shows that behaviour learned for attraction transfers well to other domains. Lastly, the second training cycle is necessary for increasing performance of the models. We note that even though it looks like the baselines don't learn at all in the first round, they do learn but tend to forget previous knowledge. This can be observed in detail in Appendix \ref{appendix-ce}.

\subsection{Forward Transfer and Forgetting} \label{ft-fg}

We calculated forward and forgetting metrics as explained in Section \ref{metrics-section}. Table \ref{tab:summary_statistics} shows success rates instead of average return because success is easier to interpret. We can see for every model the summary statistics $\mathcal{F}$ and $\mathcal{Z}$ measuring average forgetting and forward transfer, respectively. To obtain lower bounds we added forward and forgetting of a random model that is initialised randomly again every time it observes a domain.

Table \ref{tab:summary_statistics} reveals that DDPT outperforms the baselines significantly in terms of absolute numbers and also relative numbers compared to the random performance. As expected, Bin shows almost no zero-shot performance improvement compared to the random model, whereas Sem obtains slight improvement. DDPT shows large forward transfer capabilities and strong robustness against forgetting. We attribute this to the frozen description and action embeddings stemming from the language model and the domain gate. The language model allows us to interpret new information and actions immediately, enabling the model to draw connections between learned tasks and new ones. At the same time, frozen embeddings are robust to forgetting. The domain gate allows the model to choose the domain more abstractly without initial exploration due to the decision between current or non-current domains, which facilitates zero-shot performance. Moreover, the baselines need to make a hard decision between domains (balancing between choosing a domain we learn about at the moment and old domains), whereas the domain decision for DDPT is abstracted through the domain gate, leading to robustness against forgetting. Both baselines perform substantially better than the lower bound, suggesting that these are non-trivial baselines.

\begin{table}[t!]
\begin{center}
\resizebox{\columnwidth}{!}{%
\begin{tabular}{c|cc|cc|cc|cc}
    \hline
    \multicolumn{1}{c}{} &\multicolumn{2}{c}{Easy-to-hard} & \multicolumn{2}{c}{Hard-to-easy} &\multicolumn{2}{c}{Mixed order} &\multicolumn{2}{c}{Random} \\
    Model &  $\mathcal{F} \downarrow$ & $\mathcal{Z} \uparrow$ & $\mathcal{F} \downarrow$ & $\mathcal{Z} \uparrow$ &$\mathcal{F} \downarrow$ & $\mathcal{Z} \uparrow$ & $\mathcal{F} \downarrow$ & $\mathcal{Z} \uparrow$ \\
    \hline
    Bin & 0.14 & 0.39 & 0.14 & 0.45 & 0.14 & 0.38 & 0.43 & 0.39 \\
    \hline
    Sem & 0.20 & 0.39 & 0.17 & 0.37 & 0.18 & 0.29 & 0.43 & 0.26 \\
        \hline
    DDPT & \textbf{0.01} & \textbf{0.73} & \textbf{0.02} & \textbf{0.68} & \textbf{0.03} & \textbf{0.57} & 0.43 & 0.34 \\

    \hline
\end{tabular}
}
\end{center}
    \caption{Showing summary statistics in terms of success for forgetting $\mathcal{F}$ (ranging between -1 and 1, the lower the better) and forward transfer $\mathcal{Z}$ (ranging between 0 and 1, the higher the better).}
\label{tab:summary_statistics}
\end{table}

\subsection{Benefits of Domain Gate}
In order to analyse the contribution of the domain gate to the forward capabilities of DDPT, we train a DDPT model without domain gate on the easy-to-hard order, where DDPT showed the highest forward transfer. From Table \ref{tab:ablationtable} we can observe that performance drops significantly for all domains if the domain gate is not employed, which shows the importance of this mechanism. 

\begin{table}[t!]
\begin{center}
\resizebox{\columnwidth}{!}{%
\begin{tabular}{c|cccc|c}
    & Taxi & Train & Restaurant & Hotel & $\mathcal{Z} \uparrow$ \\
    \hline
    DDPT & 0.90 & 0.76 & 0.73 & 0.53 & 0.73 \\
        \hline
    DDPT w/o domain gate & 0.68 & 0.19 & 0.57 & 0.28 & 0.43 \\
        \hline
\end{tabular}
}
\end{center}
    \caption{Forward transfer metrics $\mathcal{Z}_i$ in terms of success for different domains $i$ trained on easy-to-hard order with and without domain gate.}
\label{tab:ablationtable}
\end{table}

\begin{center}
\begin{figure}[t]
\includegraphics[trim=0cm 0cm 0.0cm 0.0cm, width=0.48\textwidth]{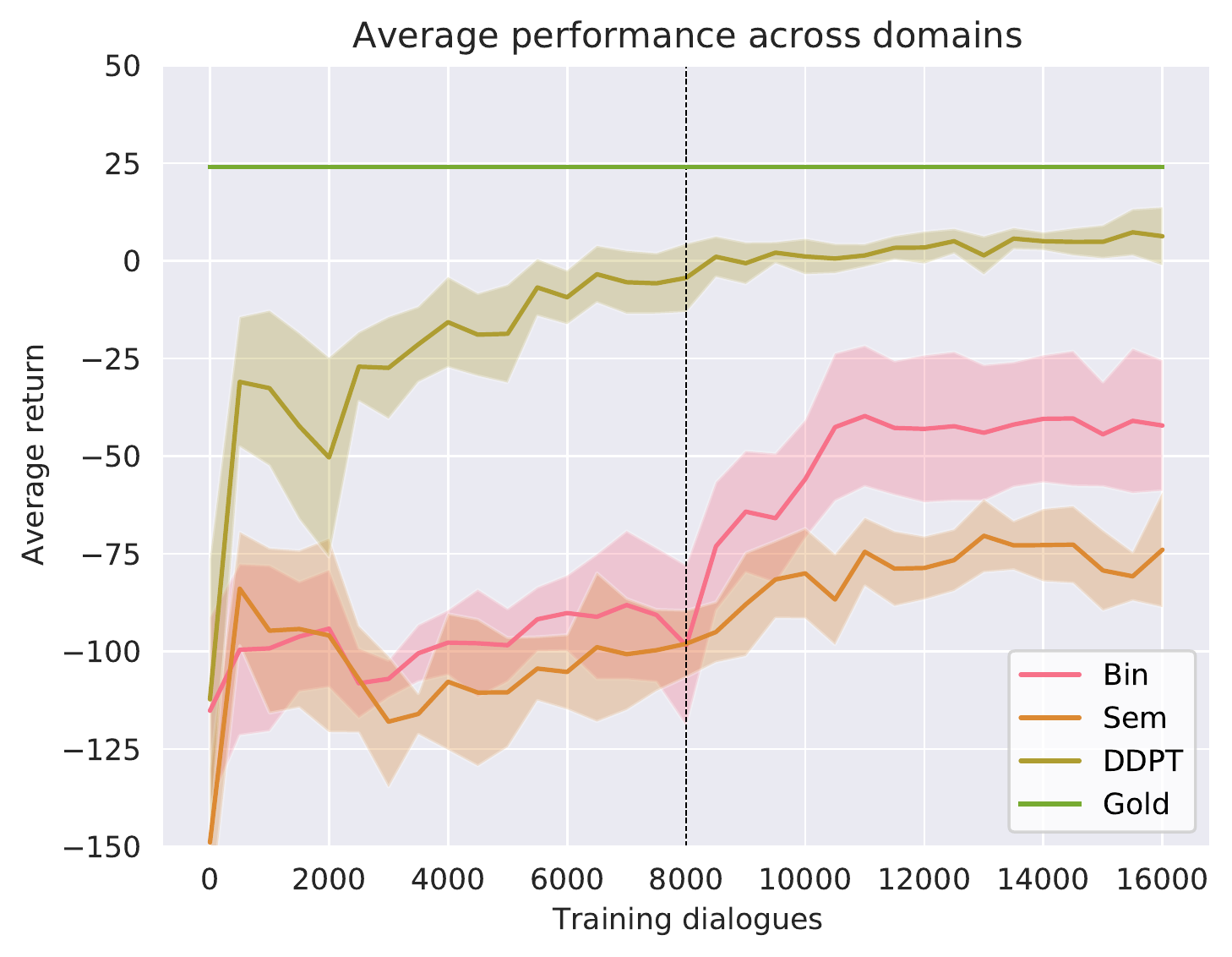}
\caption{Training Bin, Sem and DDPT (ours) on the mixed domain order with the transformer based user simulator TUS.}
\label{tus-results}
\end{figure}
\end{center}

\subsection{Results on Transformer-based Simulator}

In order to strengthen our results and show that they do not depend on the simulator used, we conducted an additional experiment using the transformer-based user simulator TUS \cite{lin-etal-2021-domain}. We only show results for the mixed order, having in mind that results have not been dependent on the domain order used. Figure \ref{tus-results} shows that DDPT again outperforms the baseline.

\subsection{Results on Human Trial}
We further validate the results by conducting a human trial. We compare Bin, Gold and DDPT, where Bin and DDPT were trained on the mixed domain order. We hire humans through Amazon Mechanical Turk and let them directly interact with our systems, thereby collecting 258, 278 and 296 dialogues for Bin, Gold and DDPT, respectively. After a user finished the dialogue we asked 1) whether the dialogue was successful (Success), 2) whether the system often mentioned something the user did not ask for such as a wrong domain (UnnecInfo) 3), whether the system gave too much information (TooMuchInfo) and 4) about the general performance (Performance). Table \ref{tab:human_trial} shows that the upper bound Gold and DDPT perform equally well ($p > 0.05$) in every metric whereas Bin performs statistically significant worse. The low performance of Bin can be partially attributed to frequently choosing a wrong domain that humans are more sensitive to than a user simulator. Example dialogues are given in Appendix \ref{trial-examples}.

\begin{table}[t!]
\begin{center}
\resizebox{\columnwidth}{!}{%
\begin{tabular}{c|cccc}
    & Success $\uparrow$ & UnnecInfo $\downarrow$ & TooMuchInfo $\downarrow$ & Performance $\uparrow$\\
    \hline
    Bin & 0.45 & 3.98 & 3.15 & 2.45 \\
        \hline
    Gold & 0.81 & 2.79 & 2.71 & 3.65\\
        \hline
    DDPT & 0.77 & 2.75 & 2.56 & 3.67\\
        \hline
\end{tabular}
}
\end{center}
    \caption{Human trial results where Bin, Gold and DDPT interacted with real users. There is no statistically significant difference (p > 0.05) between DDPT  and Gold, while Bin is statistically significantly worse (p < 0.05) than Gold and DDPT.}
\label{tab:human_trial}
\end{table}


\section{Conclusion}
In this work we provided an algorithm, baseline models and evaluation metrics to enable continual RL for dialogue policy optimisation. Moreover, we proposed a dynamic dialogue policy model called DDPT that builds on information descriptions, a pretrained language model and the transformer encoder-decoder architecture. It integrates new information seamlessly as long as it is descriptive, and obtains significant zero-shot performance on unseen domains while being robust to forgetting. The strengths of DDPT were validated in simulation with two simulators as well as humans. This opens the door for building evolving dialogue systems, that continually expand their knowledge and improve their behaviour throughout their lifetime.

\section*{Acknowledgements}
This work is a part of DYMO project which has received funding from the European Research Council (ERC) provided under the Horizon 2020 research and innovation programme (Grant agreement No. STG2018 804636). N. Lubis, C. van Niekerk, M. Heck and S. Feng are funded by an Alexander von Humboldt Sofja Kovalevskaja Award endowed by the German Federal Ministry of Education and Research. Computing resources were provided by Google Cloud and HHU ZIM.

\bibliography{anthology,custom}
\bibliographystyle{acl_natbib}

\appendix

\section{Appendix}\label{appendix}

\subsection{Background on CLEAR} \label{clear-background}

\subsubsection{VTRACE Algorithm}

VTRACE \cite{espeholt18} is an off-policy actor critic algorithm. As such, it optimizes both a policy $\pi_\theta$ and a corresponding critic $V_\psi$ that estimates the state-value function $V$ of $\pi_\theta$. Actor and critic are both updated using experience from a replay buffer $\mathcal{B}$. 

Given a trajectory $\tau = (s_t, a_t, r_t)^{t=k+n}_{t=k}$ generated by a behaviour policy $\mu$, the $n$-steps vtrace-target for $V(s_k)$ is defined as
\begin{align*}
v_k = V(s_k) + \sum_{t=k}^{k+n-1} \gamma^{t-k} (\prod_{i=k}^{t-1} c_i) \delta_t V, 
\end{align*}
where $\delta_t V = \rho_t(r_t + \gamma V(s_{t+1}) - V(s_t))$ is a temporal difference for $V$, and $\rho_t = \min (\overline{\rho}, \frac{\pi(a_t|s_t)}{\mu(a_t|s_t)})$ and $c_i = \min (\overline{c}, \frac{\pi(a_i|s_i)}{\mu(a_i|s_i)})$ are truncated importance sampling weights. The scalars $\overline{\rho}$ and $\overline{c}$ are hyperparameters where it is assumed that $\overline{\rho} \geq \overline{c}$.

The critic function is then optimized to minimize the gap between its prediction and the vtrace-target:

\begin{equation} \label{critic-loss}
    \mathcal{L}_{critic}(\psi) = \mathbb{E}_{\tau \sim \mathcal{B}}[(v_k - V_\psi(s_k))^2]
\end{equation}

The actor is optimized using the following off-policy policy gradient:

\begin{equation} \label{actor-loss}
    \mathbb{E}_{\tau \sim \mathcal{B}}[\frac{\pi(a_k|s_k)}{\mu(a_k|s_k)} A_k \nabla_\theta \log \pi_\theta(a_k|s_k)]
\end{equation}
where $A_k = (r_k + \gamma v_{k+1} - V_\psi(s_k))$ is an estimate of the advantage function. To prevent premature convergence, they add an entropy loss $\mathcal{L}_{entropy}(\theta)$ during optimization.

\subsubsection{CLEAR}

CLEAR is a continual learning algorithm that adapts VTRACE to fulfill the continual learning requirements. The goal is to obtain fast adaptation capabilities as well as preventing catastrophic forgetting. Fast adaptation is tackled by using the most recent trajectories instead of randomly sampling from the buffer $\mathcal{B}$ in Equations \ref{critic-loss} and \ref{actor-loss}.

In order to prevent catastrophic forgetting, they sample non-recent experience from the replay buffer and update policy and critic using Equations \ref{critic-loss} and \ref{actor-loss}. To further regularize these non-recent updates, they introduce regularization losses $\mathcal{L}_{\pi-reg}$ and $\mathcal{L}_{v-reg}$. $\mathcal{L}_{v-reg}$ forces the critic prediction to be close to the historic prediction through a mean-squared error loss. $\mathcal{L}_{\pi-reg}$ regularizes the actor to minimize the KL-divergence between the behaviour policy $\mu$ and current policy $\pi_\theta$:

\begin{align*}
    \mathcal{L}_{v-reg}(\psi) &= \mathbb{E}_{\tau \sim \mathcal{B}}[(V_\psi(s_k) - V_{replay}(s_k))^2] \\
    \mathcal{L}_{\pi-reg}(\theta) &= \mathbb{E}_{\tau \sim \mathcal{B}}[\sum_a \mu(a_|s_k) \log \frac{\mu(a_|s_k)}{\pi_\theta(a_|s_k)}]
\end{align*}

An online-offline ratio determines how many recent and non-recent experience is used in an update, thereby trading-off fast adaptation and catastrophic forgetting prevention.

\subsection{Training details}
For the baselines, the MLP encoder uses a 3-layer MLP with hidden dimension of 128 and RELU as activation function. We use a GRU with 2 layers and input size as well as hidden size of 128 for action decoding. The domain, intent and slot embeddings for action prediction have a size of $64$. They are fed through a linear layer that projects it to a vector of size $128$ (same size as GRU output) in order to allow computation of the scalar product with the GRU output. The semantic encoding in Sem uses an embedding size of 32 for domain, intent, slot and values. The critic for Bin and Sem has the same architecture as the MLP encoder, with an additional linear layer to project the output to a real valued number.

For the DDPT model, we use an input size and hidden size of 128 in both transformer encoder and decoder. We use two heads for the encoder and decoder, 4 transformer layers for the encoder and 2 for the decoder. The critic for DDPT has the same architecture as the transformer encoder, obtaining the same input as the policy module plus an additional CLS vector (as in RoBERTa). The output of the CLS vector is fed into a linear layer to obtain the critic prediction.

For every model, we use the same training configurations. We use the ADAM optimiser \cite{adam15} with a learning rate of 5e-5 and 1e-4 for policy and critic module, respectively. We sample a batch of 64 episodes for updating the model after every 2 new dialogues. The replay buffer size is set to 5000. For the VTRACE algorithm, the parameters $\bar{\rho}$ and $\bar{c}$ are set to $1.0$. For CLEAR we use an online-offline ratio of $0.2$, i.e. $20\%$ of the dialogues in a batch are from the most recent dialogues and the remaining $80\%$ from historical dialogues. The regularization losses are weighted by $0.1$ and the entropy loss by $0.01$.

We used a NVIDIA Tesla T4 provided by the Google Cloud Platform for training the models. The training of one model took 10 to 16 hours depending on the architecture used.

\subsection{Masking of illegal actions}
To aid the policy in the difficult RL environment, we add a simple masking mechanism that prohibits illegal actions. The action masking includes the following

\begin{itemize}
    \item If the data base query tells us that entities for a domain are available, the policy is not allowed to say that there are no entities available.
    \item If there is no entity found with the current constraints, the policy is not allowed to inform on information about entities.
    \item The \emph{Booking} domain is only usable for hotel and restaurant.
\end{itemize}

\subsection{Baselines} \label{appendix-baselines}

As mentioned in Section \ref{baselines}, the second baseline incorporates the idea from \citet{metapolicy}, which uses trainable embeddings for domains, intents and slots to allow cross-domain transfer. For every feature category (such as user-act, user goal, etc.) and every domain, it calculates for every feature in that category a representation using trainable domain, intent and slot embeddings. The features in a category are then averaged over domains to obtain a final representation.

For instance, considering the user-act category for a domain $d$, the user act $(d, i_k, s_k)_{k=0}^n$ is first embedded as $\hat{\vctr{s}}_{\text{u-act}, d} = \frac{1}{n} \sum_{k=0}^n [\vctr{v}_d, \vctr{v}_{i_k}, \vctr{v}_{s_k}]$, where $\vctr{v}_d, \vctr{v}_{i_k}$ and $\vctr{v}_{s_k}$ are trainable embeddings for domain $d$, intents $i_k$ and slots $s_k$ and afterwards fed through a residual block, leading to $\vctr{s}_{\text{u-act}, d} = \hat{\vctr{s}}_{\text{u-act}, d} + \mathrm{ReLU}(\mtrx{W}_{\text{u-act}} \hat{\vctr{s}}_{\text{u-act}, d} + \vctr{b}_{\text{u-act}})$. If there is no user-act for domain $d$, we use an embedding for \textit{no-user-act} to indicate that. The overall feature representation for the user-act is then given by $\vctr{s}_{\text{u-act}} = \frac{1}{|\mathcal{D}|} \sum_{d \in \mathcal{D}} \vctr{s}_{\text{u-act}, d}$.

The representations for different feature categories are then concatenated and fed into a multi-layer perceptron encoder. The state encoding can be seen in Figure \ref{state_representation}(b). We abbreviate this baselines as \emph{Sem} as it uses semantic features. 

\subsection{Descriptions} \label{appendix-descriptions}
Our DDPT model uses descriptions for every possible information. This allows us to seamlessly deal with new information we have not seen before yet by leveraging a pretrained language model. The language model provides us token embeddings for the description, which are averaged in order to obtain the description embedding. The descriptions are built as follows. 

\begin{itemize}
    \item For every domain \texttt{d} and every slot \texttt{s} the user can inform on, the description is given by \texttt{user goal <d> <s>}. The corresponding value is $1$, if that slot has been mentioned and $0$ else.
    \item For every atomic user act \texttt{d i s} that was used in the current turn, the description is given by \texttt{user act <d> <i> <s>}. We consider each atomic user act as one information and only provide user acts that were used in the current turn to the model with a corresponding value of $1$.
    \item For every atomic system act \texttt{d i s} that was used in the previous turn, the description is given by \texttt{last system act <d> <i> <s>} with a corresponding value of $1$.
    \item For every domain \texttt{d} where a data base query is possible to obtain the number of entities that fulfill the user constraints, the description is given by \texttt{data base <d> <number of entities>} with a corresponding value indicating the number of search results.
    \item For every domain \texttt{d} where an entity can be booked, the description is given by \texttt{general <d> <booked>} with a binary indicating whether an entity has already been booked.
\end{itemize}

\subsection{Human trial} \label{trial-examples}

\begin{center}
\begin{figure}[t]
\includegraphics[trim=0cm 0cm 0.0cm 0.0cm, width=0.48\textwidth]{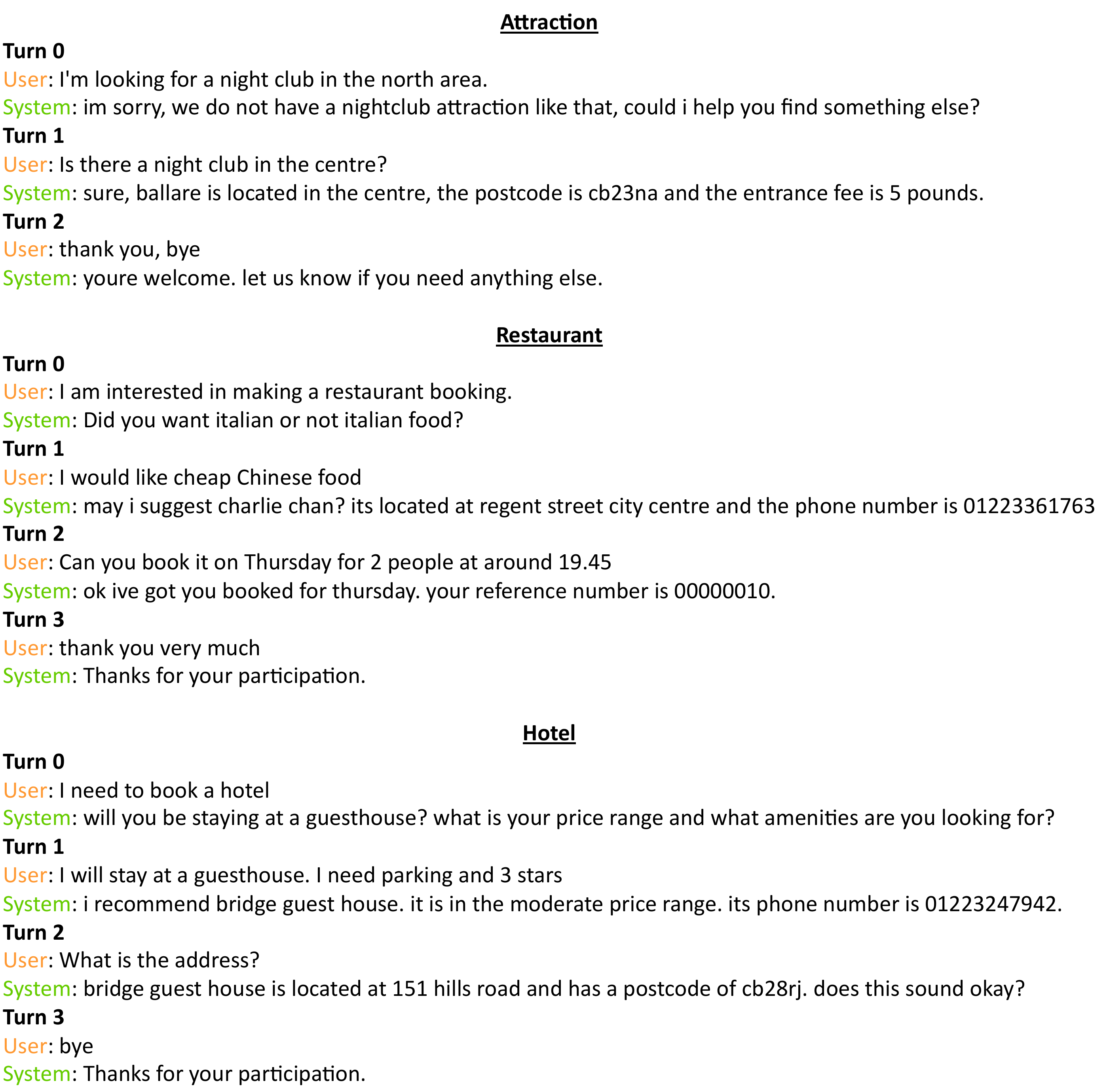}
\caption{Example dialogues that were collected during the human trial. Users hired through Amazon Mechanical Turk interact with our DDPT model.}
\label{example_dialogues_ddpt}
\end{figure}
\end{center}

We conducted a human trial to validate our results in simulation. The website was build using DialCrowd \cite{dialcrowd} and users were hired using Amazon Mechanical Turk. We used SetSUMBT \cite{carel21} as belief tracker and SC-GPT \cite{scgpt} as NLG module to accompany the dialogue policies Bin, Gold and DDPT in the dialogue system pipelines. Example dialogues, where DDPT interacted with users hired through Amazon Mechanical Turk, are depicted in Figure \ref{example_dialogues_ddpt}.

\subsection{Forward Transfer and Forgetting}
We provide the forward and forgetting tables in terms of success rate and average return in Tables \ref{tab:forwardtable}, \ref{tab:forgettingtable}, \ref{forwardtable-return}, \ref{forgettingtable-return}.

\begin{table}[t!]
\begin{center}
\resizebox{\columnwidth}{!}{%
\begin{tabular}{c|ccc|ccc|ccc}
    \hline
    \multicolumn{1}{c}{} &\multicolumn{3}{c}{Easy-to-hard} & \multicolumn{3}{c}{Hard-to-easy} &\multicolumn{3}{c}{Mixed order}  \\
    Task &  Bin & Sem & DDPT & Bin & Sem & DDPT & Bin & Sem & DDPT \\
    \hline
    Attraction & / & / & / & 0.43 & 0.35 & 0.83 & 0.60 & 0.33 & 0.79  \\
    \hline
    Taxi & 0.51 & 0.75 & 0.90 & 0.51 & 0.47 & 0.85 & 0.35 & 0.43 & 0.77 \\
        \hline
    Train & 0.21 & 0.18 & 0.76 & 0.23 & 0.15 & 0.28 & 0.17 & 0.09 & 0.34 \\
        \hline
    Restaurant & 0.47 & 0.36 & 0.73 & 0.62 & 0.52 & 0.74 & / & / & /  \\
        \hline
    Hotel & 0.36 & 0.26 & 0.53 & / & / & / & 0.39 & 0.28 & 0.39 \\
    \hline 
    \textbf{Average} & 0.39 & 0.39 & 0.73 & 0.45 & 0.37 & 0.68 & 0.38 & 0.29 & 0.57 \\
    
    \textbf{Random} & 0.39 & 0.26 & 0.34 & 0.39 & 0.26 & 0.34 & 0.39 & 0.26 & 0.34 \\
    \hline
\end{tabular}
}
\end{center}
    \caption{Forward transfer table showing for every domain $i$ the metric $\mathcal{Z}_i$ in terms of success rate, where numbers range between 0 and 1. The higher the number, the more forward transfer is achieved.}
\label{tab:forwardtable}
\end{table}

\begin{table}[t!]
\begin{center}
\resizebox{\columnwidth}{!}{%
\begin{tabular}{c|ccc|ccc|ccc|c}
    \hline
    \multicolumn{1}{c}{} &\multicolumn{3}{c}{Easy-to-hard} & \multicolumn{3}{c}{Hard-to-easy} &\multicolumn{3}{c}{Mixed order} & Random \\
    Task &  Bin & Sem & DDPT & Bin & Sem & DDPT & Bin & Sem & DDPT & \\
    \hline
    Attraction & 0.28 & 0.49 & 0.03 & 0.08 & 0.09 & 0.02 & 0.29 & 0.40 & 0.0 & \\
        \hline
    Taxi & 0.13 & 0.15 & 0.01 & 0.01 & 0.01 & 0.02 & 0.01 & 0.02 & 0.0 & \\
        \hline
    Train & 0.18 & 0.20 & 0.02 & 0.13 & 0.14 & -0.01 & 0.03 & 0.03 & 0.0 & \\
        \hline
    Restaurant & 0.06 & 0.11 & -0.01 & 0.16 & 0.19 & 0.0 & 0.22 & 0.26 & 0.09 & \\
        \hline
    Hotel & 0.04 & 0.07 & 0.0 & 0.32 & 0.41 & 0.07 & 0.14 & 0.19 & 0.03 &\\
    \hline
    \textbf{Average} & 0.14 & 0.20 & 0.01 & 0.14 & 0.17 & 0.02 & 0.14 & 0.18 & 0.03 & 0.43\\
    \hline
\end{tabular}
}
\end{center}
    \caption{Forgetting table showing for every domain $i$ the metric $\mathcal{F}_i$ in terms of success rate, where numbers range between -1 and 1. Negative numbers indicate backward transfer whereas positive numbers indicate forgetting.}
\label{tab:forgettingtable}
\end{table}

\begin{table}[t!]
\begin{center}
\resizebox{\columnwidth}{!}{%
\begin{tabular}{c|ccc|ccc|ccc}
    \hline
    \multicolumn{1}{c}{} &\multicolumn{3}{c}{Easy-to-hard} & \multicolumn{3}{c}{Hard-to-easy} &\multicolumn{3}{c}{Mixed order} \\
    Task &  Bin & Sem & DDPT & Bin & Sem & DDPT & Bin & Sem & DDPT \\
    \hline
    Attraction & / & / & / & -88 & -124 & 12 & -16 & -125 & -3 \\
    \hline
    Taxi & -91 & -32 & 23 & -65 & -117 & 13 & -85 & -127 & -12 \\
        \hline
    Train & -149 & -156 & -17 & -66 & -180 & -108 & -140 & -189 & -112\\
        \hline
    Restaurant & -94 & -119 & -15 & -15 & -97 & -19 & / & / & /\\
        \hline
    Hotel & -121 & -143 & -81 & / & / & / & -45 & -139 & -107\\
    \hline
    \textbf{Average} & -114 & -113 & -23 & -58 & -129 & -25 & -71 & -145 & -58 \\
    \hline
\end{tabular}
}
\end{center}
    \caption{Forward transfer table showing for every domain $i$ the metric $\mathcal{Z}_i$ in terms of average return. The higher the number, the more forward transfer is achieved.}
\label{forwardtable-return}
\end{table}

\begin{table}[t!]
\begin{center}
\resizebox{\columnwidth}{!}{%
\begin{tabular}{c|ccc|ccc|ccc}
    \hline
    \multicolumn{1}{c}{} &\multicolumn{3}{c}{Easy-to-hard} & \multicolumn{3}{c}{Hard-to-easy} &\multicolumn{3}{c}{Mixed order} \\
    Task &  Bin & Sem & DDPT & Bin & Sem & DDPT & Bin & Sem & DDPT \\
    \hline
    Attraction & 99 & 151 & 6 & 34 & 36 & 2 & 93 & 126 & 1 \\
        \hline
    Taxi & 73 & 89 & 4 & 16 & 23 & 4 & 18 & 29 & 1 \\
        \hline
    Train & 68 & 68 & 1 & 43 & 49 & -2 & 10 & 10 & -1 \\
        \hline
    Restaurant & 35 & 38 & -1 & 59 & 71 & 2 & 78 & 91 & 26 \\
        \hline
    Hotel & 12 & 21 & -1 & 89 & 112 & 18 & 51 & 59 & 7\\
    \hline
    \textbf{Average} & 58 & 73 & 2 & 48 & 58 & 5 & 50 & 63 & 7 \\
    \hline
\end{tabular}
}
\end{center}
    \caption{Forgetting table showing for every domain $i$ the metric $\mathcal{F}_i$ in terms of average return. Negative numbers indicate backward transfer whereas positive numbers indicate forgetting.}
\label{forgettingtable-return}
\end{table}

\subsection{Continual Evaluation} \label{appendix-ce}

\begin{figure*}[t!]
  \begin{subfigure}[]{0.3\textwidth}
    \includegraphics[scale=0.37]{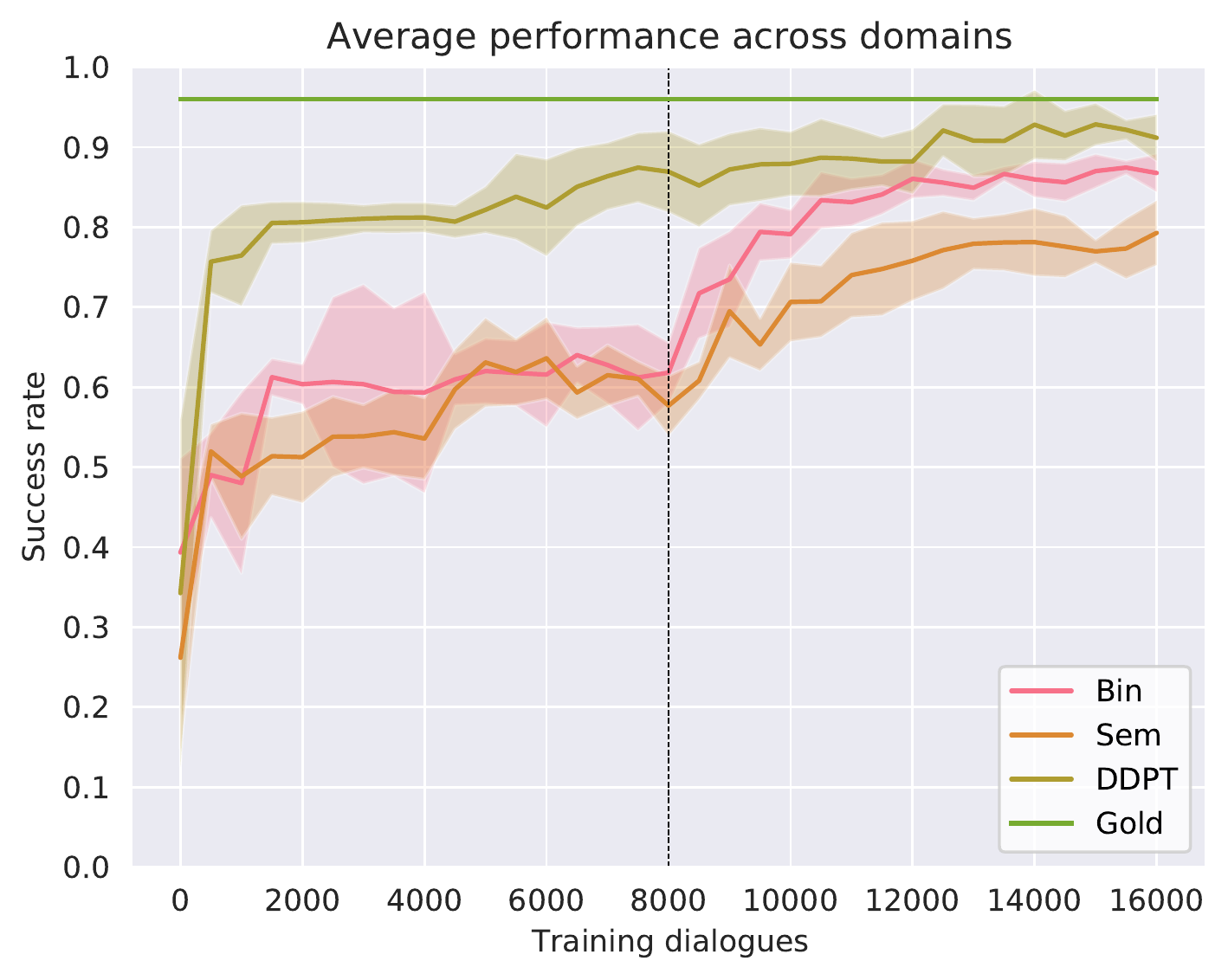}
    \caption{easy-to-hard order}
  \end{subfigure}
  \quad
  \begin{subfigure}[]{0.3\textwidth}
    \includegraphics[scale=0.37]{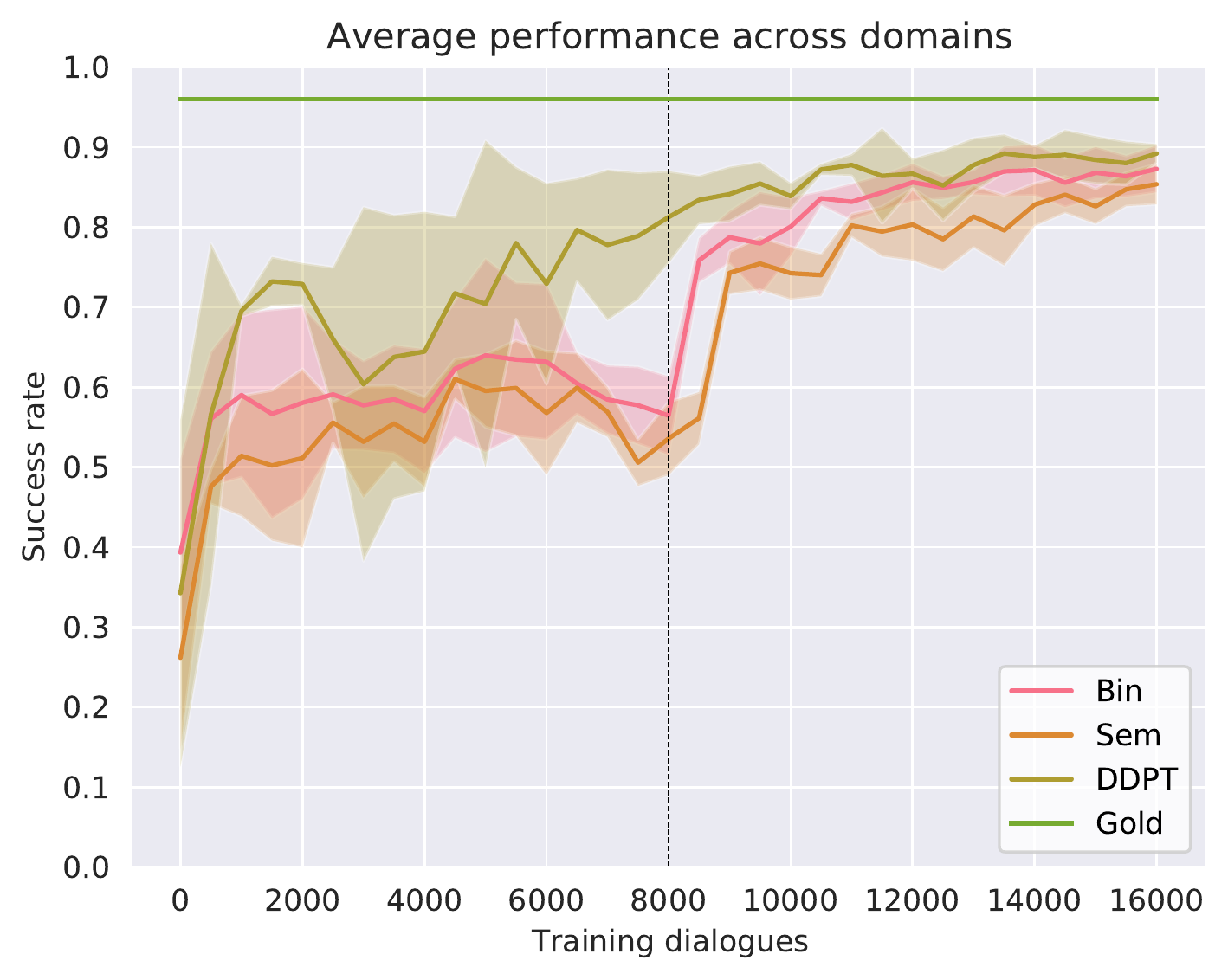}
    \caption{hard-to-easy order}
  \end{subfigure}
  \quad
  \begin{subfigure}[]{0.3\textwidth}
    \includegraphics[scale=0.37]{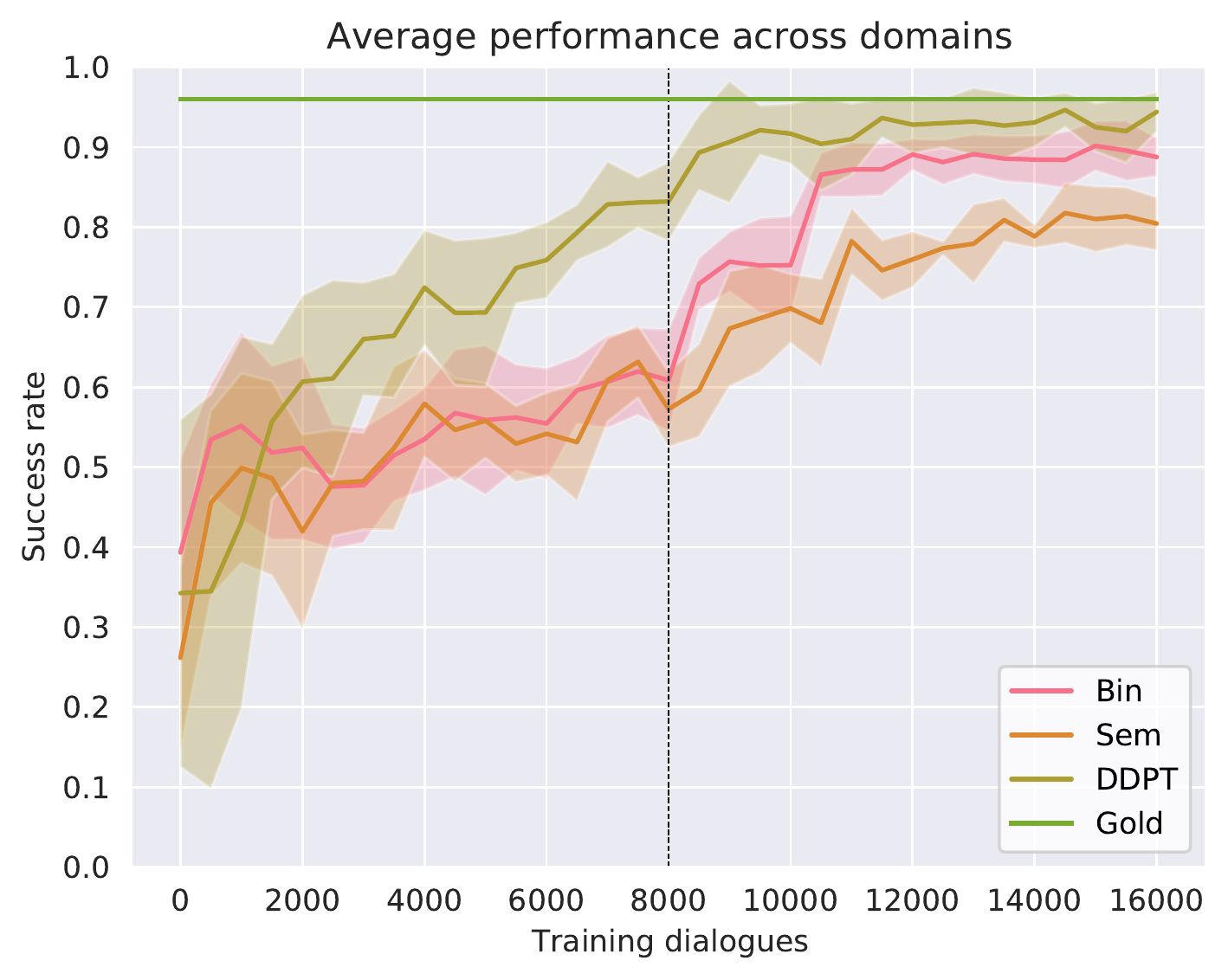}
    \caption{mixed domain order}
  \end{subfigure}

\caption{Training the three architectures Bin, Sem and DDPT using CLEAR on three different domain orders, each with 5 different seeds. Each model is evaluated every 500 training dialogues on 100 dialogues per domain. The plots show the success rate, where performance is averaged over domains. The vertical line at 8000 dialogues indicates the start of cycle 2.}
\label{domains-averaged-results-success}
\end{figure*}

Here, we provide in-depth results for all experiments. Each graph shows the performance of a single domain during training. Moreover, we provide the average performance over domains in terms of success rate in Figure \ref{domains-averaged-results-success} to complement Figure \ref{domains-averaged-results}.

\begin{figure*}[t!]
  \begin{subfigure}[]{0.3\textwidth}
    \includegraphics[scale=0.37]{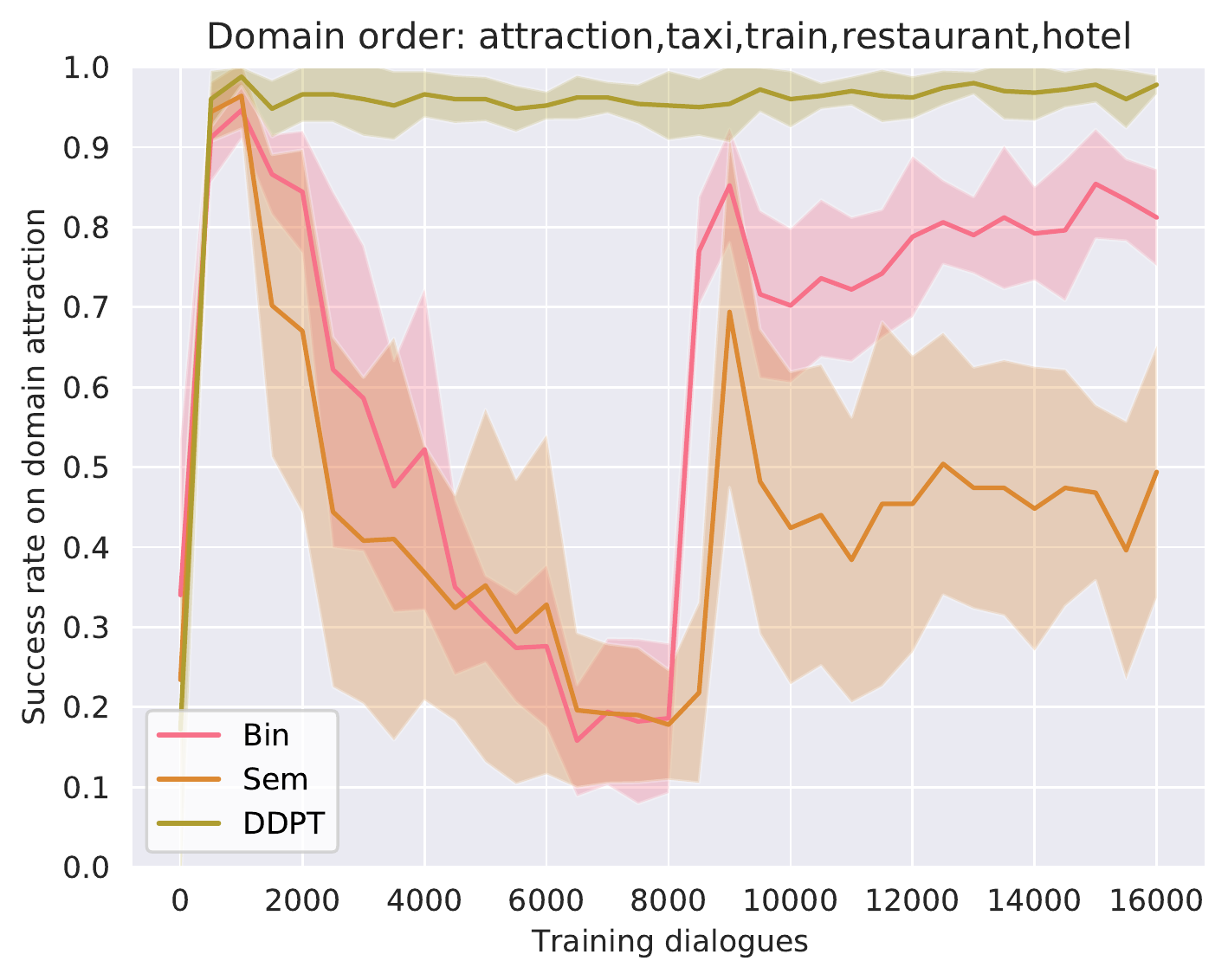}
    \caption{Success rate on attraction domain}
  \end{subfigure}
  \quad
  \begin{subfigure}[]{0.3\textwidth}
    \includegraphics[scale=0.37]{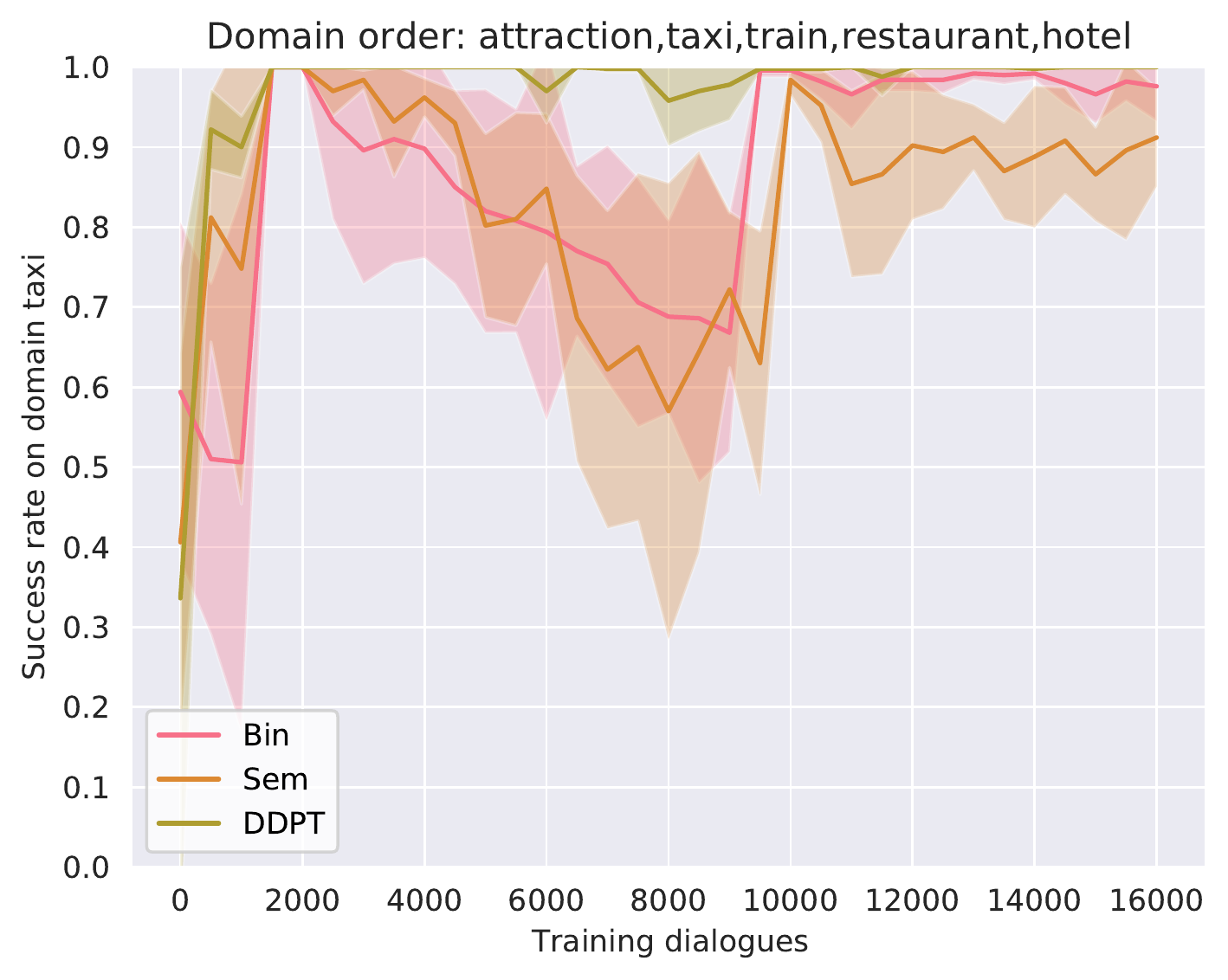}
    \caption{Success rate on taxi domain}
  \end{subfigure}
  \quad
  \begin{subfigure}[]{0.3\textwidth}
    \includegraphics[scale=0.37]{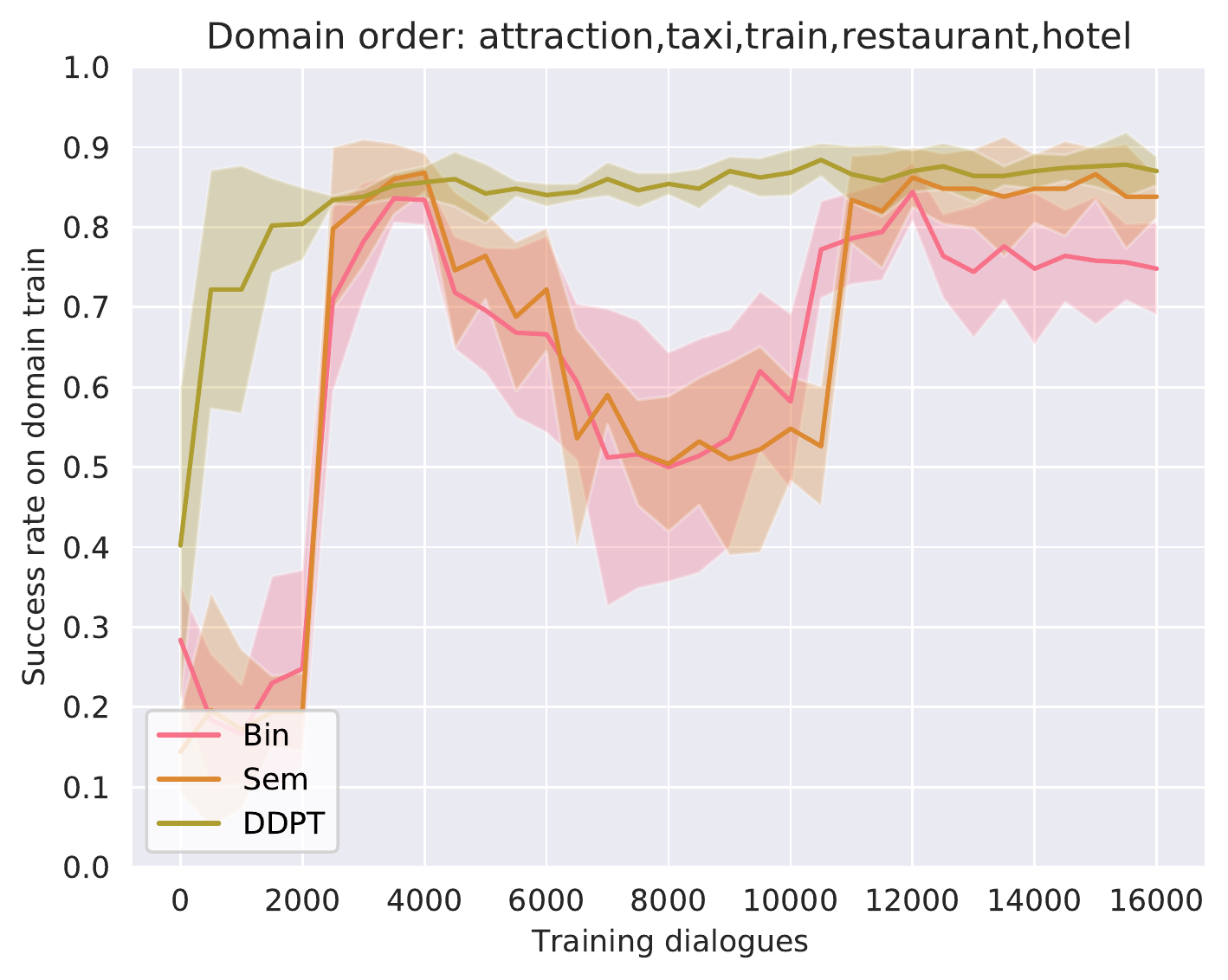}
    \caption{Success rate on train domain}
  \end{subfigure}
\\
  \begin{subfigure}[]{0.3\textwidth}
    \includegraphics[scale=0.37]{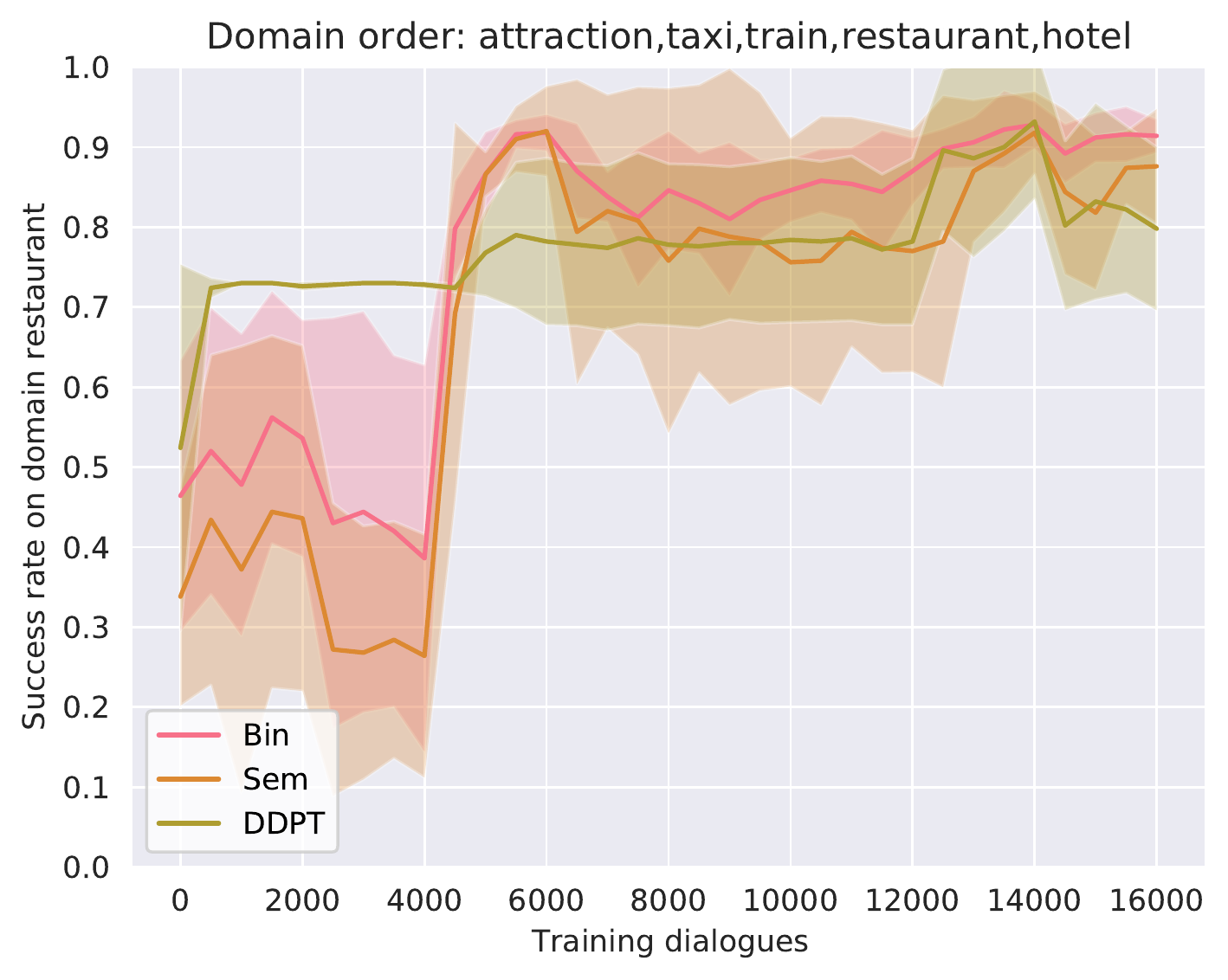}
    \caption{Success rate on restaurant domain}
  \end{subfigure}
  \quad
  \begin{subfigure}[]{0.3\textwidth}
    \includegraphics[scale=0.37]{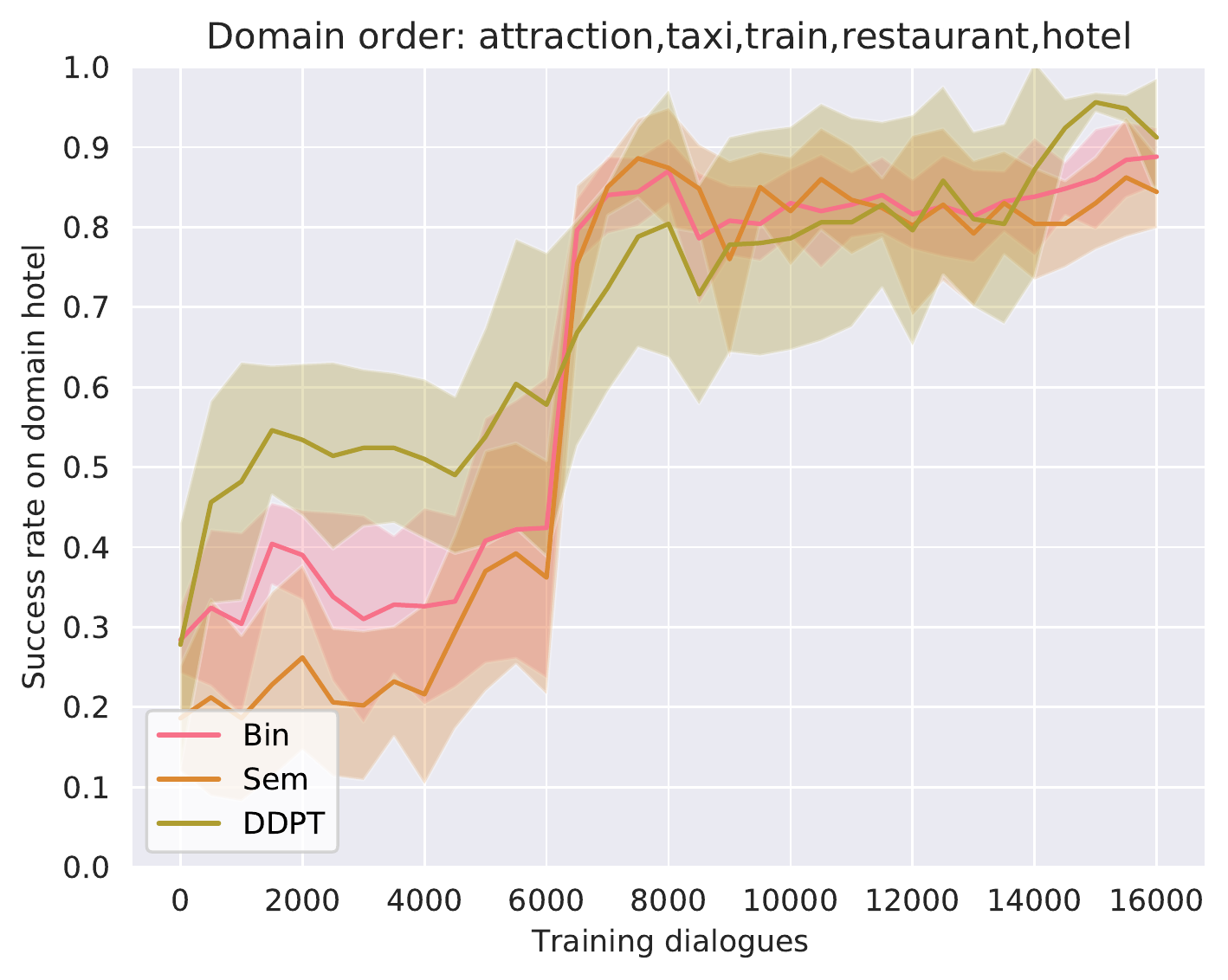}
    \caption{Success rate on hotel domain}
  \end{subfigure}

\caption{Success rate for each individual domain, where algorithms are trained in the order easy-to-hard.}
\label{}
\end{figure*}

\begin{figure*}[t!]
  \begin{subfigure}[]{0.3\textwidth}
    \includegraphics[scale=0.37]{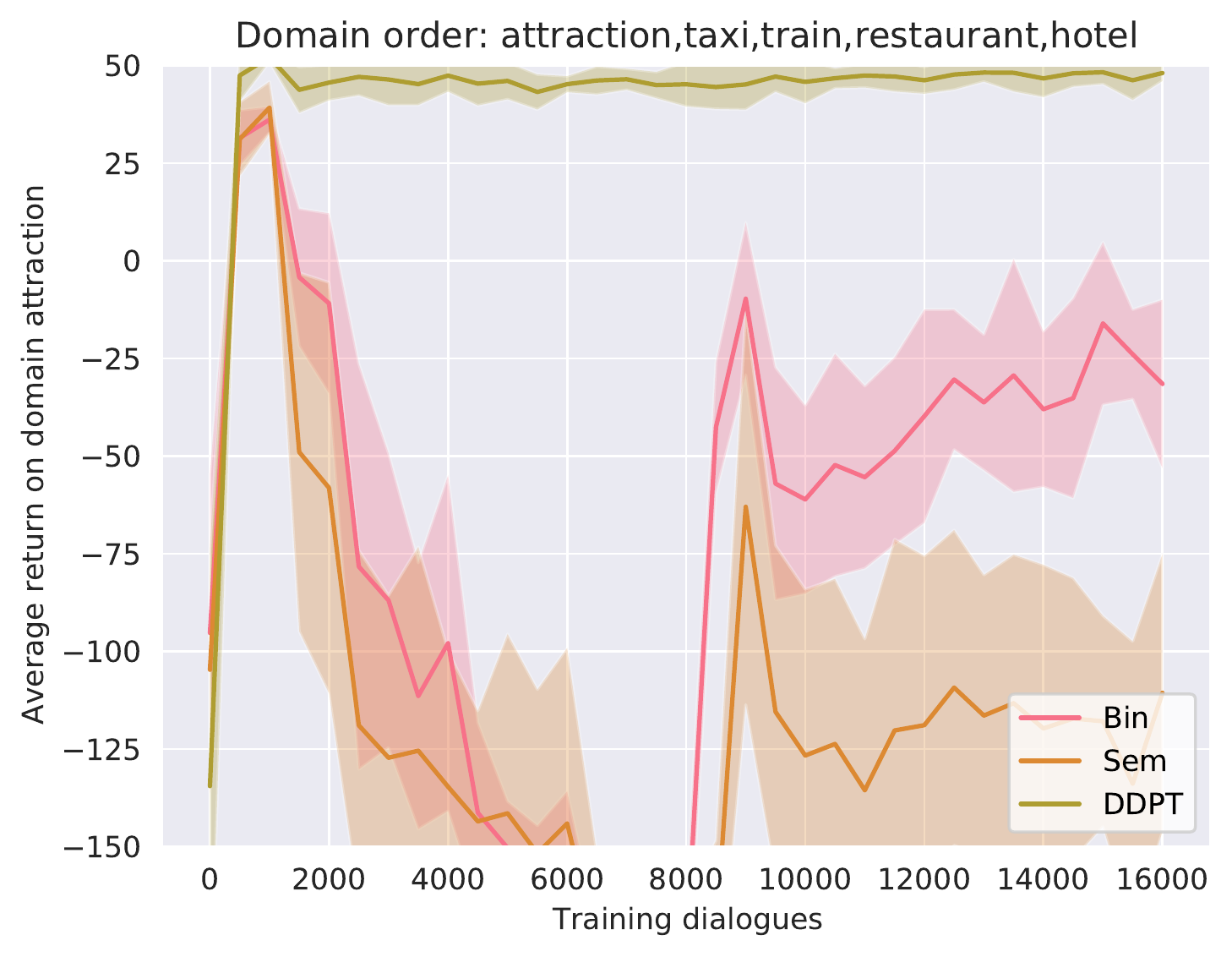}
    \caption{Average return on attraction domain}
  \end{subfigure}
  \quad
  \begin{subfigure}[]{0.3\textwidth}
    \includegraphics[scale=0.37]{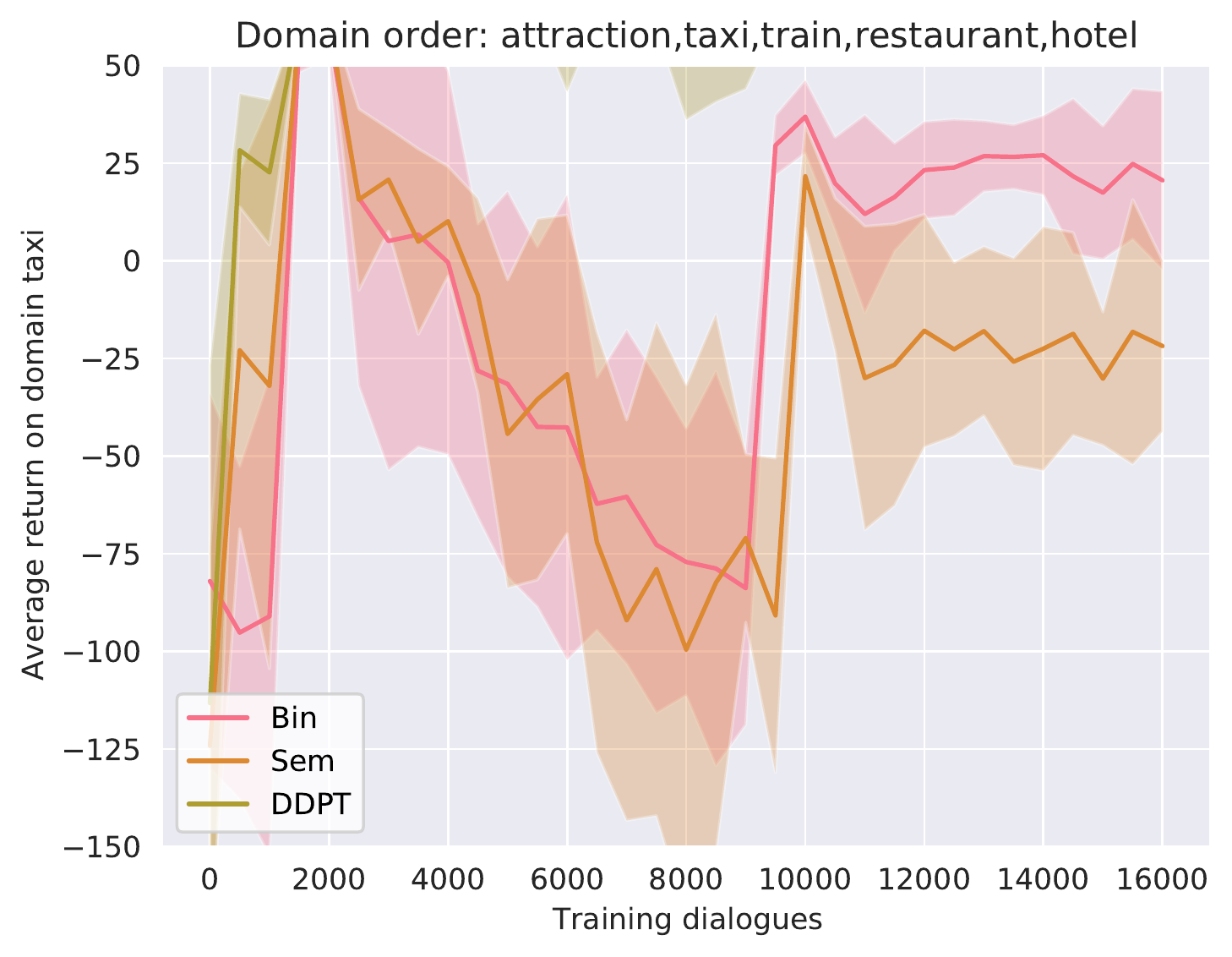}
    \caption{Average return on taxi domain}
  \end{subfigure}
  \quad
  \begin{subfigure}[]{0.3\textwidth}
    \includegraphics[scale=0.37]{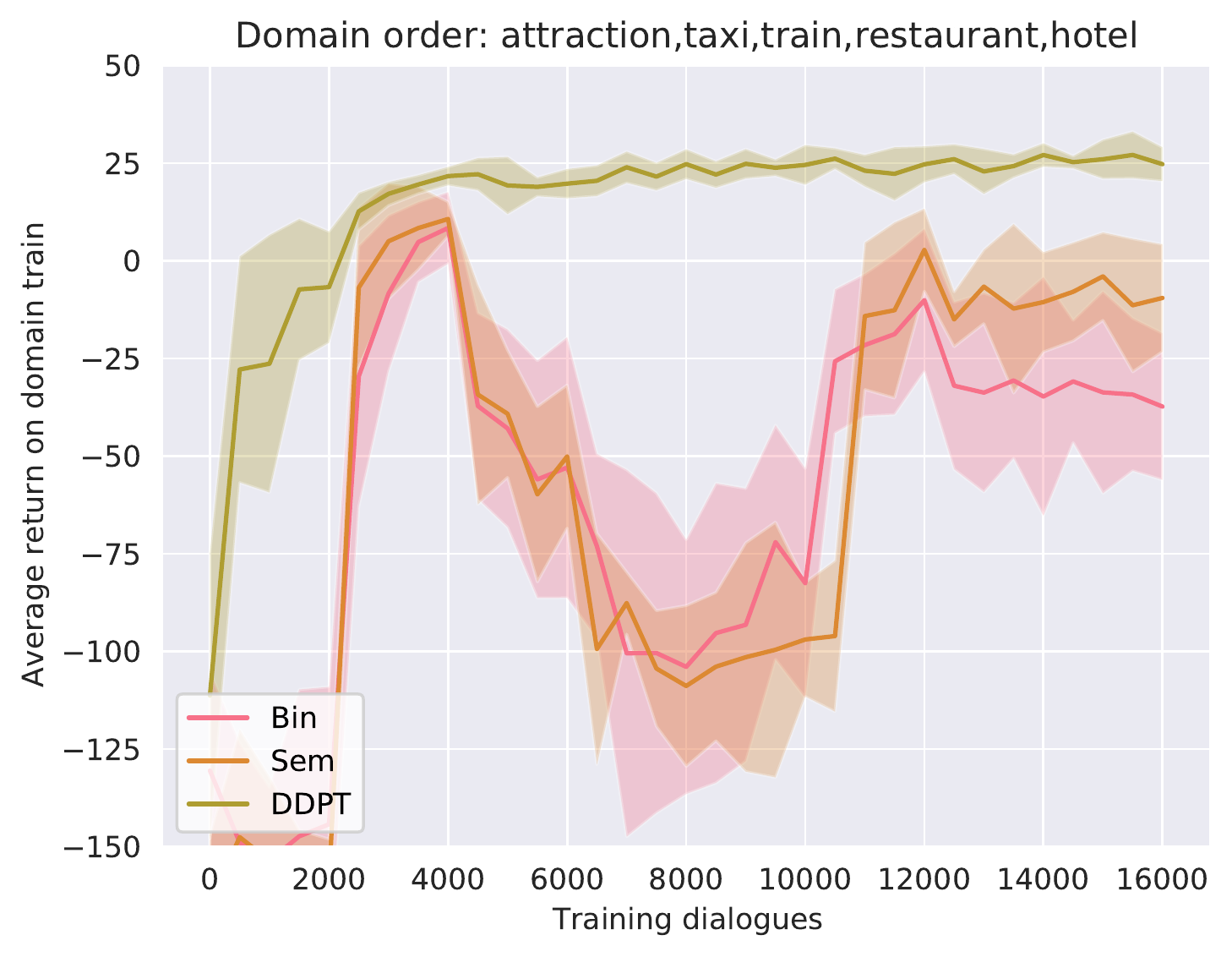}
    \caption{Average return on train domain}
  \end{subfigure}
\\
  \begin{subfigure}[]{0.3\textwidth}
    \includegraphics[scale=0.37]{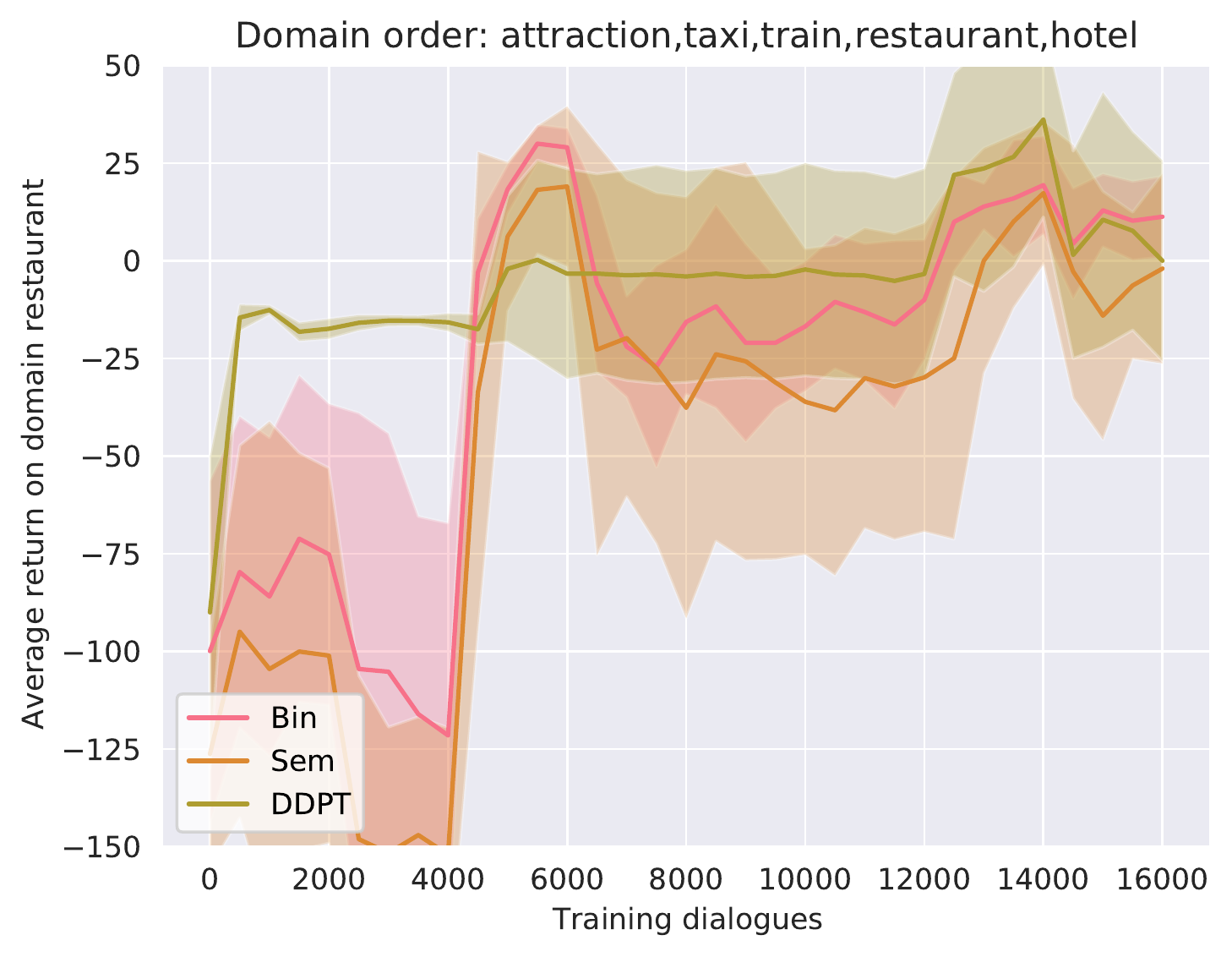}
    \caption{Average return on restaurant domain}
  \end{subfigure}
  \quad
  \begin{subfigure}[]{0.3\textwidth}
    \includegraphics[scale=0.37]{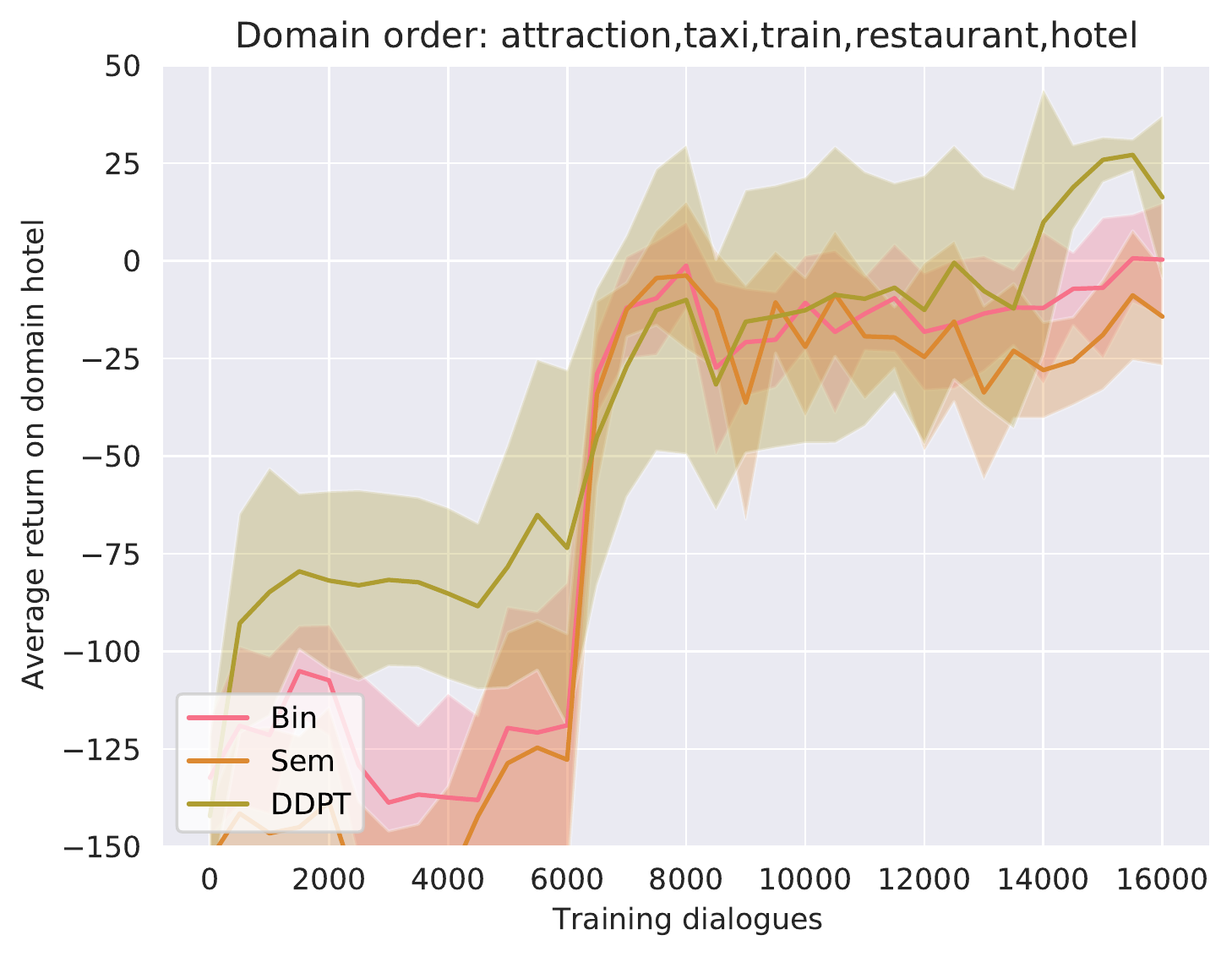}
    \caption{Average return on hotel domain}
  \end{subfigure}

\caption{Average return for each individual domain, where algorithms are trained in the order easy-to-hard.}
\label{}
\end{figure*}

\begin{figure*}[t!]
  \begin{subfigure}[]{0.3\textwidth}
    \includegraphics[scale=0.37]{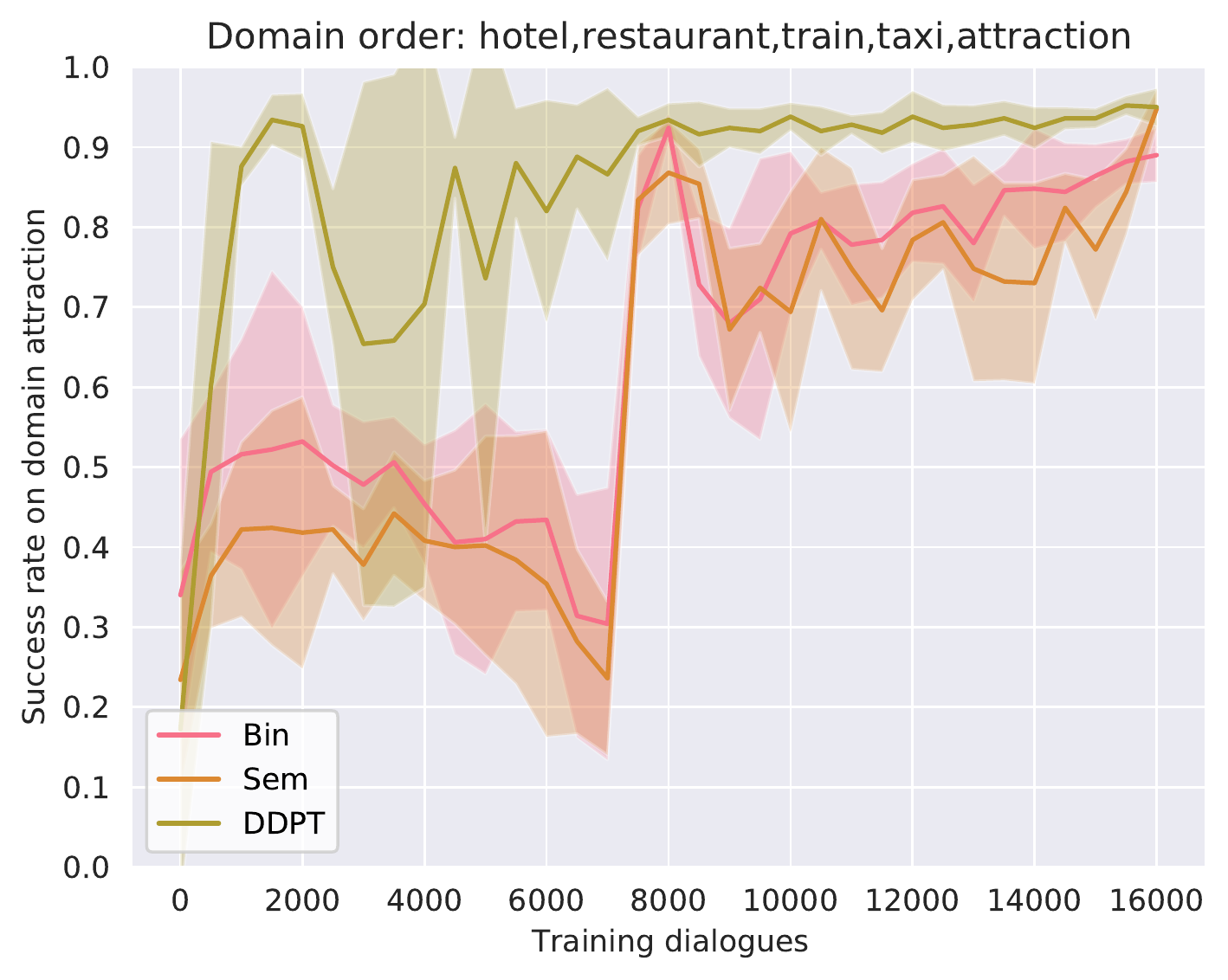}
    \caption{Success rate on attraction domain}
  \end{subfigure}
  \quad
  \begin{subfigure}[]{0.3\textwidth}
    \includegraphics[scale=0.37]{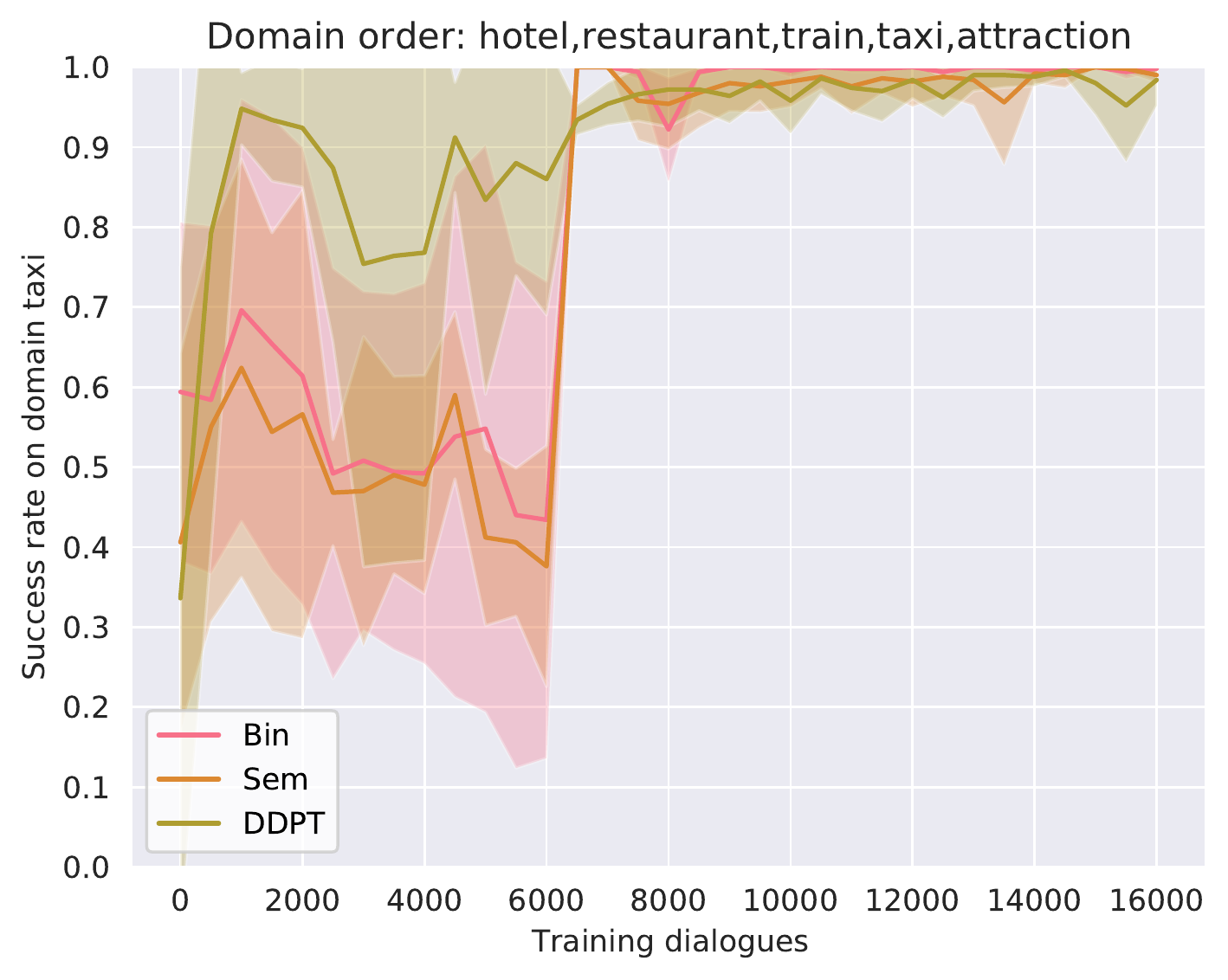}
    \caption{Success rate on taxi domain}
  \end{subfigure}
  \quad
  \begin{subfigure}[]{0.3\textwidth}
    \includegraphics[scale=0.37]{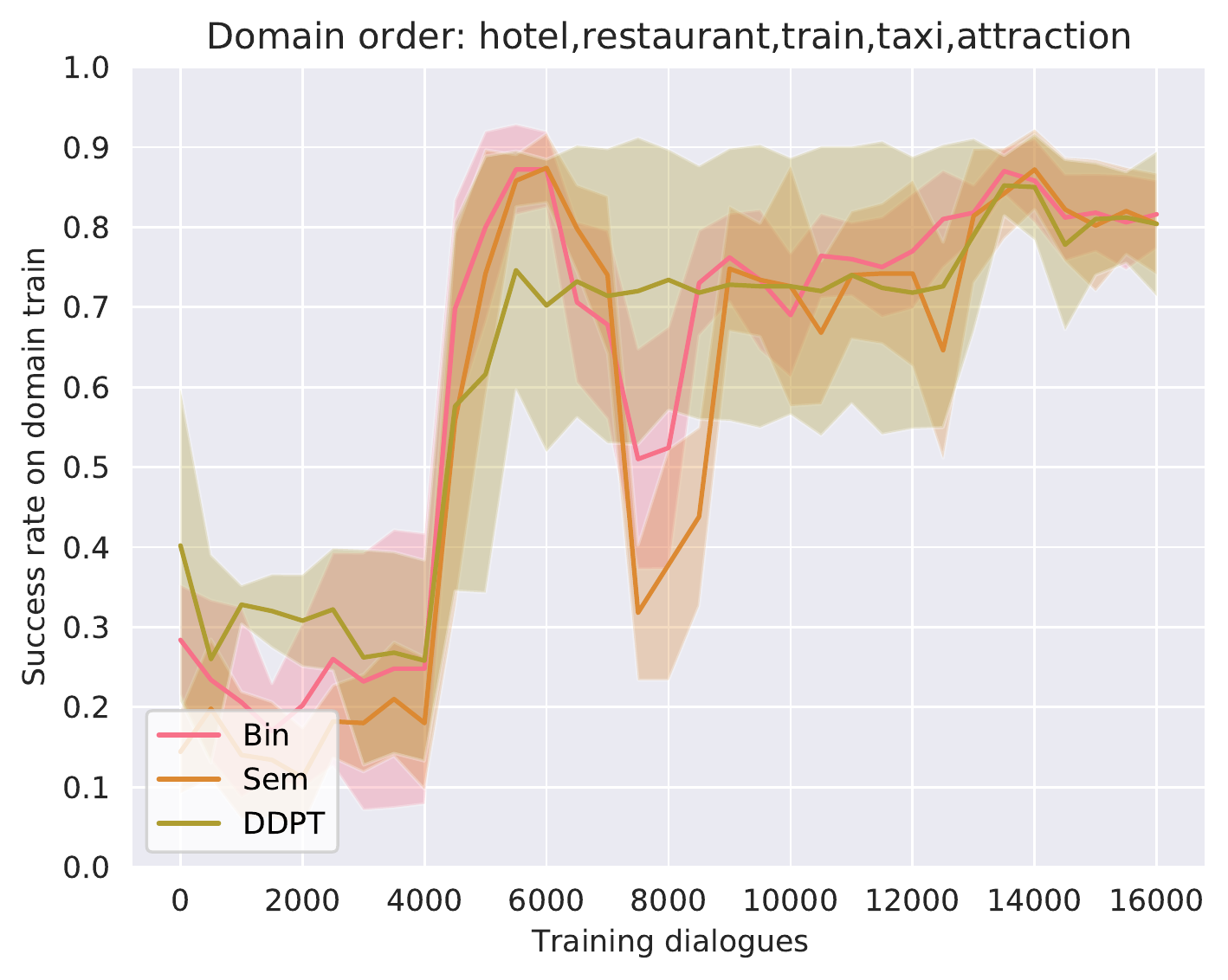}
    \caption{Success rate on train domain}
  \end{subfigure}
\\
  \begin{subfigure}[]{0.3\textwidth}
    \includegraphics[scale=0.37]{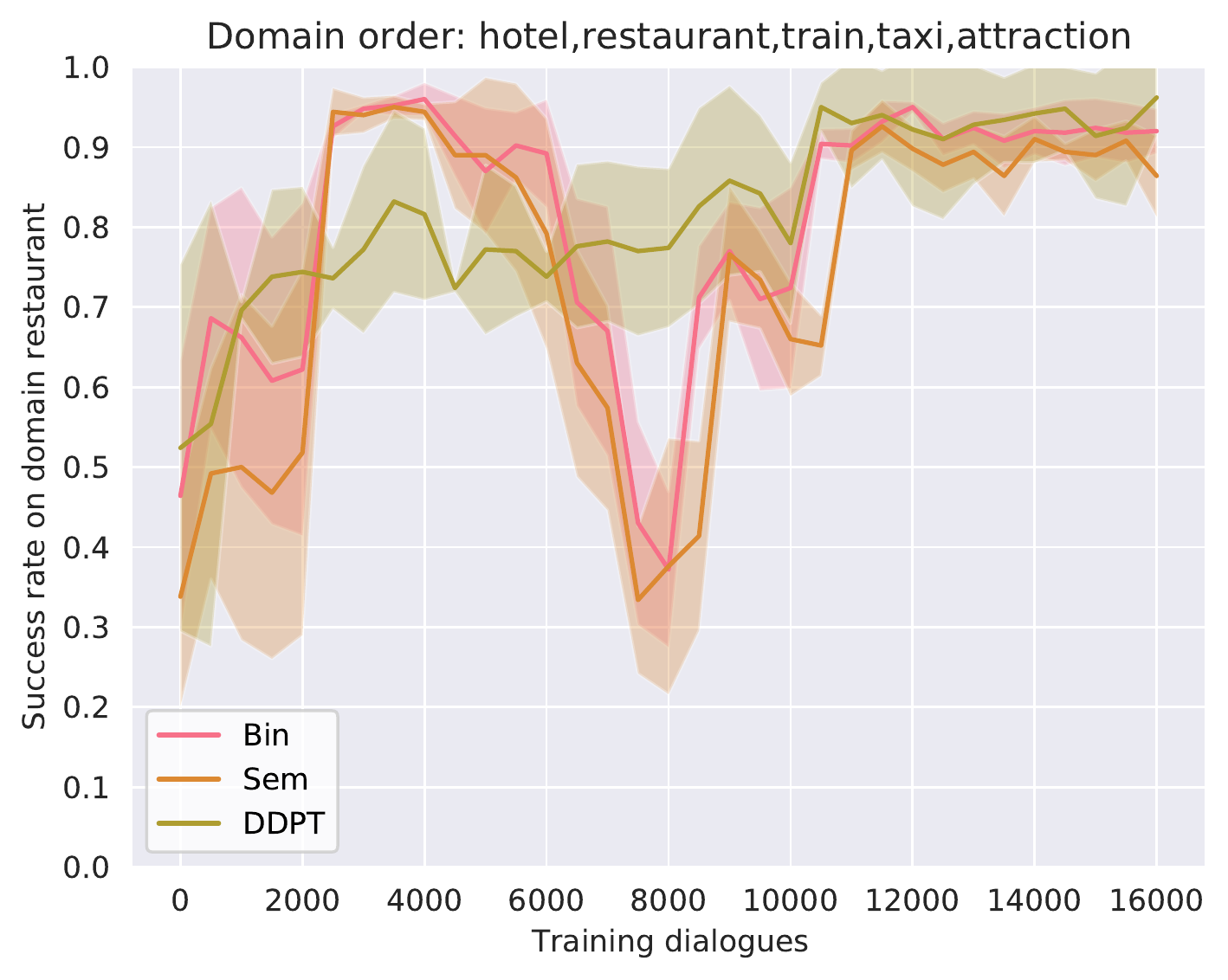}
    \caption{Success rate on restaurant domain}
  \end{subfigure}
  \quad
  \begin{subfigure}[]{0.3\textwidth}
    \includegraphics[scale=0.37]{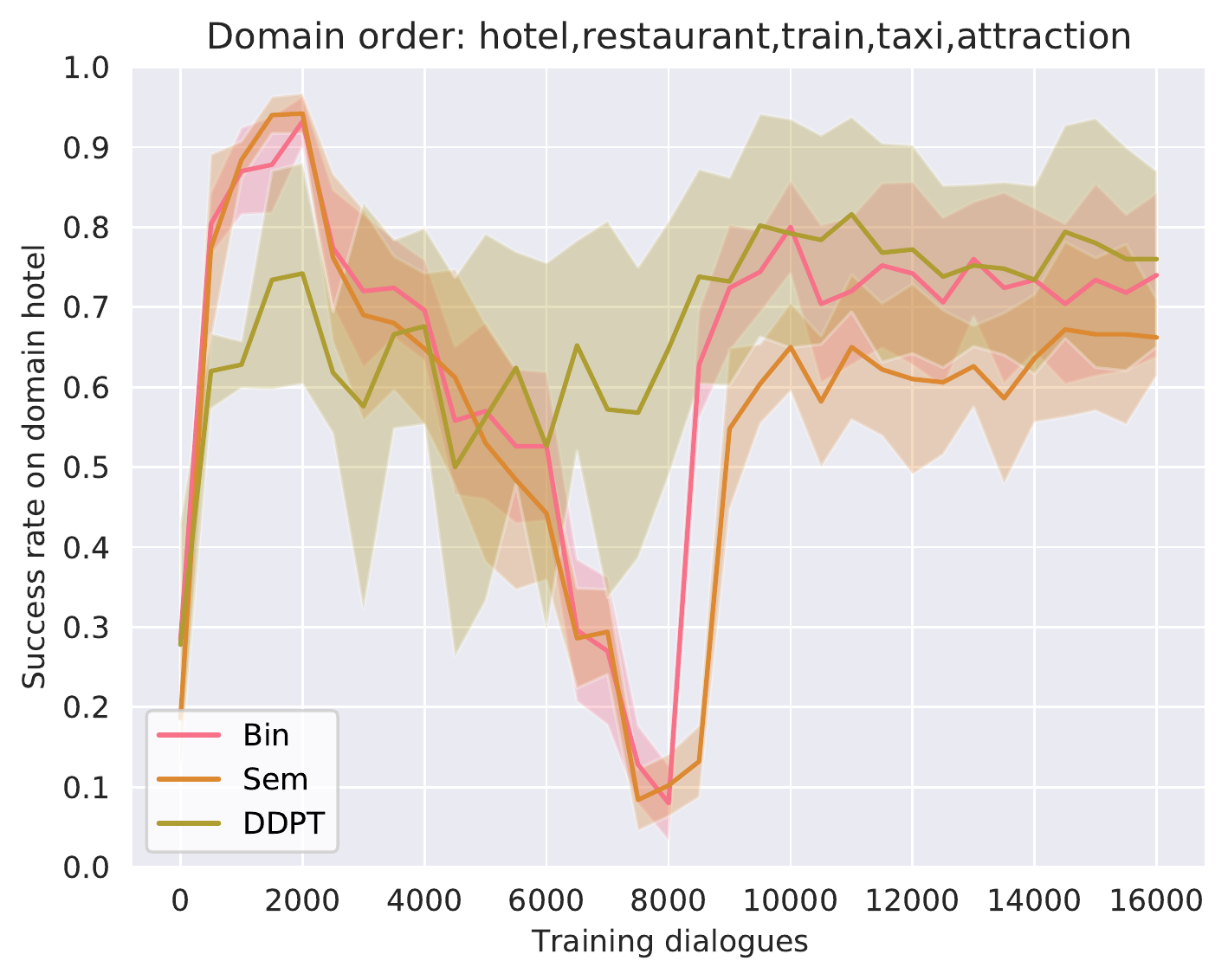}
    \caption{Success rate on hotel domain}
  \end{subfigure}

\caption{Success rate for each individual domain, where algorithms are trained in the order hard-to-easy.}
\label{}
\end{figure*}

\begin{figure*}[t!]
  \begin{subfigure}[]{0.3\textwidth}
    \includegraphics[scale=0.37]{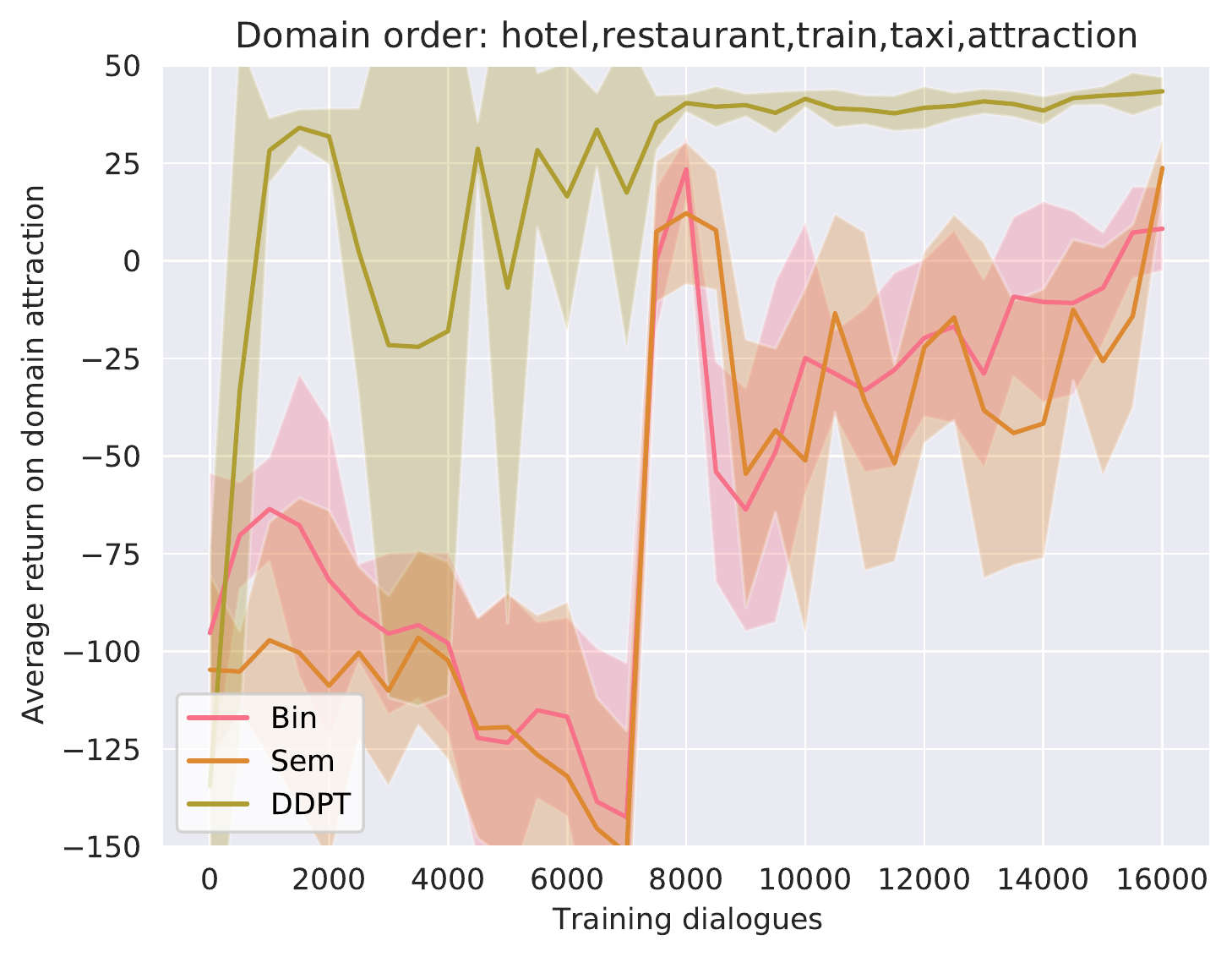}
    \caption{Average return on attraction domain}
  \end{subfigure}
  \quad
  \begin{subfigure}[]{0.3\textwidth}
    \includegraphics[scale=0.37]{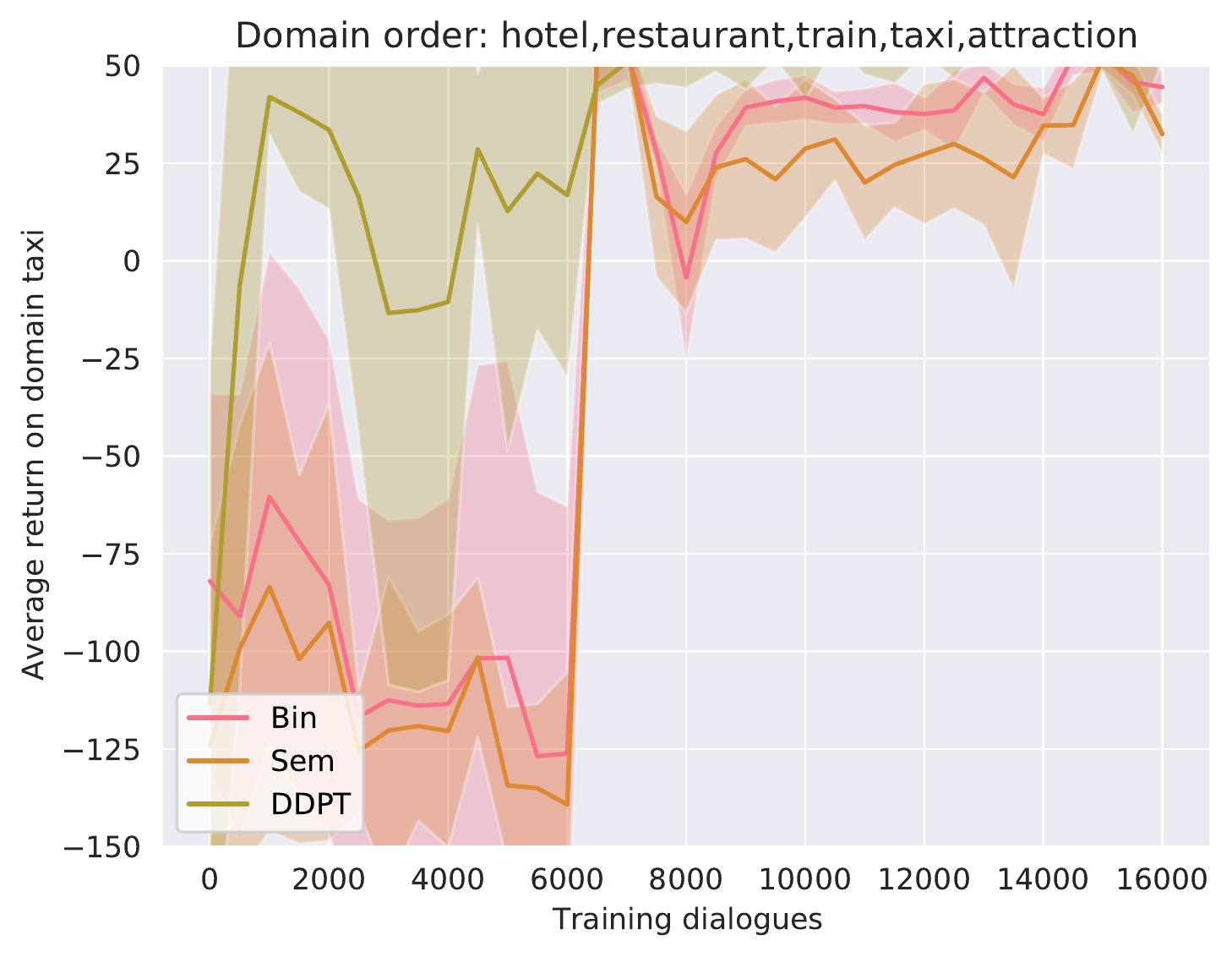}
    \caption{Average return on taxi domain}
  \end{subfigure}
  \quad
  \begin{subfigure}[]{0.3\textwidth}
    \includegraphics[scale=0.37]{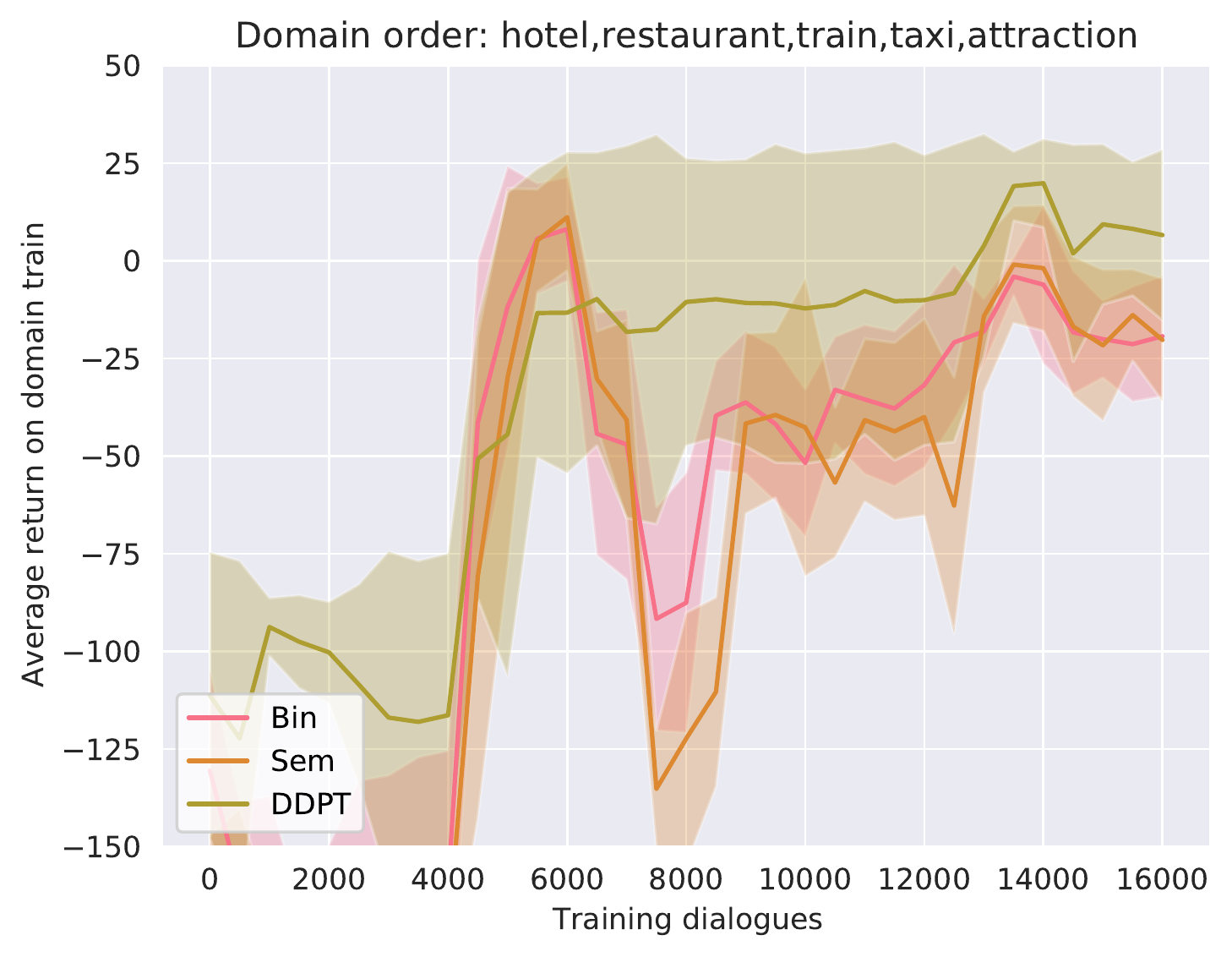}
    \caption{Average return on train domain}
  \end{subfigure}
\\
  \begin{subfigure}[]{0.3\textwidth}
    \includegraphics[scale=0.37]{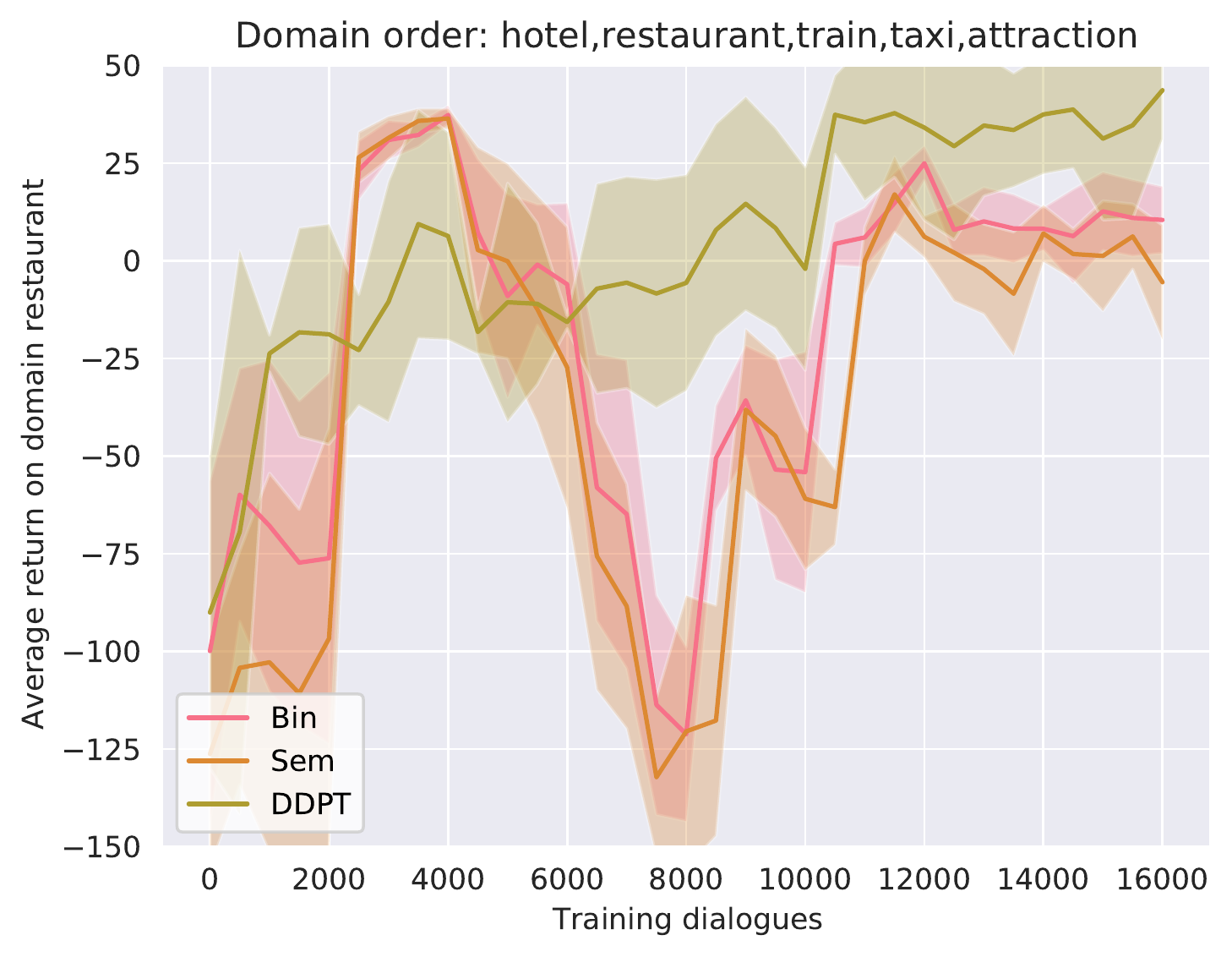}
    \caption{Average return on restaurant domain}
  \end{subfigure}
  \quad
  \begin{subfigure}[]{0.3\textwidth}
    \includegraphics[scale=0.37]{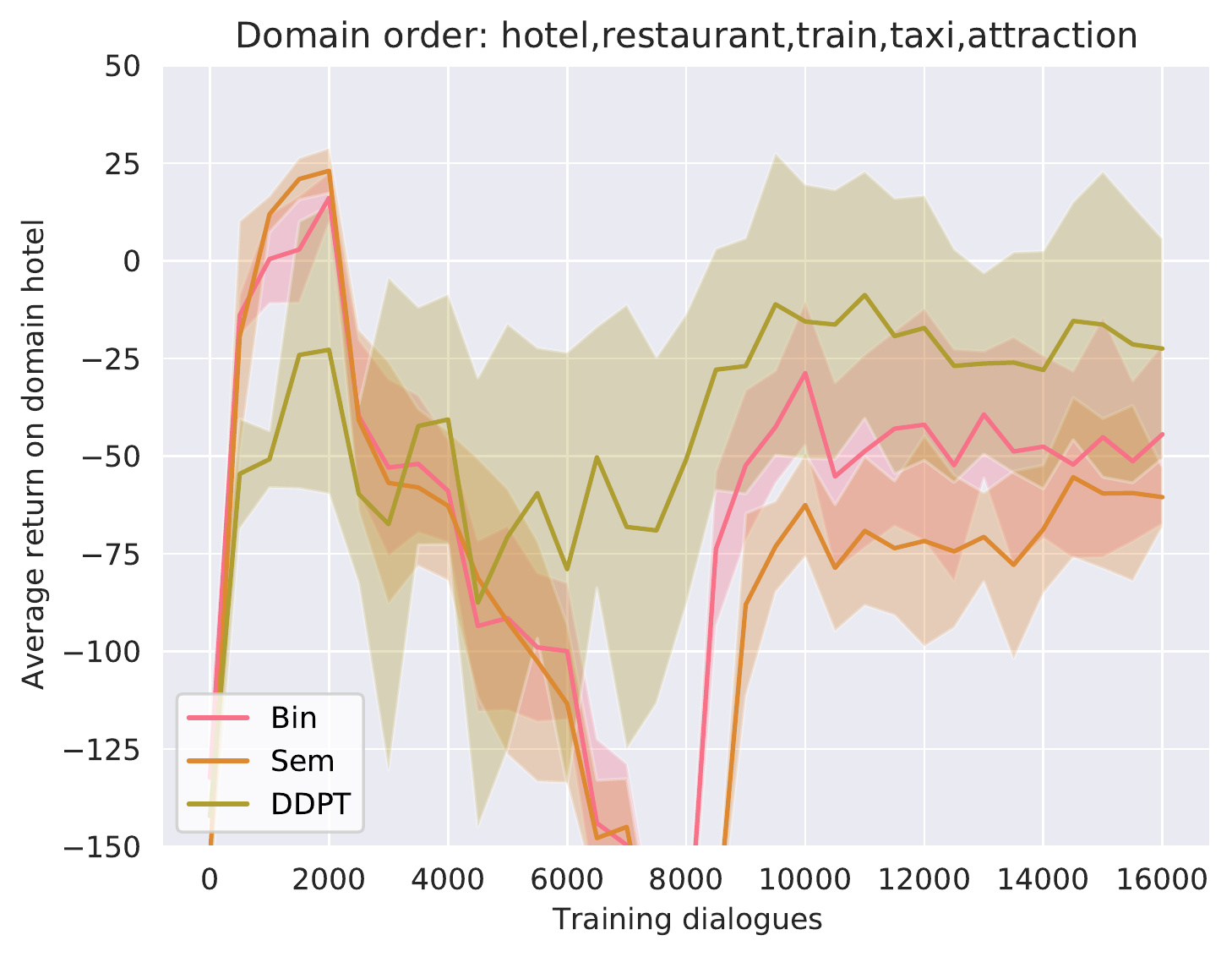}
    \caption{Average return on hotel domain}
  \end{subfigure}

\caption{Average return for each individual domain, where algorithms are trained in the order hard-to-easy.}
\label{}
\end{figure*}

\begin{figure*}[t!]
  \begin{subfigure}[]{0.3\textwidth}
    \includegraphics[scale=0.37]{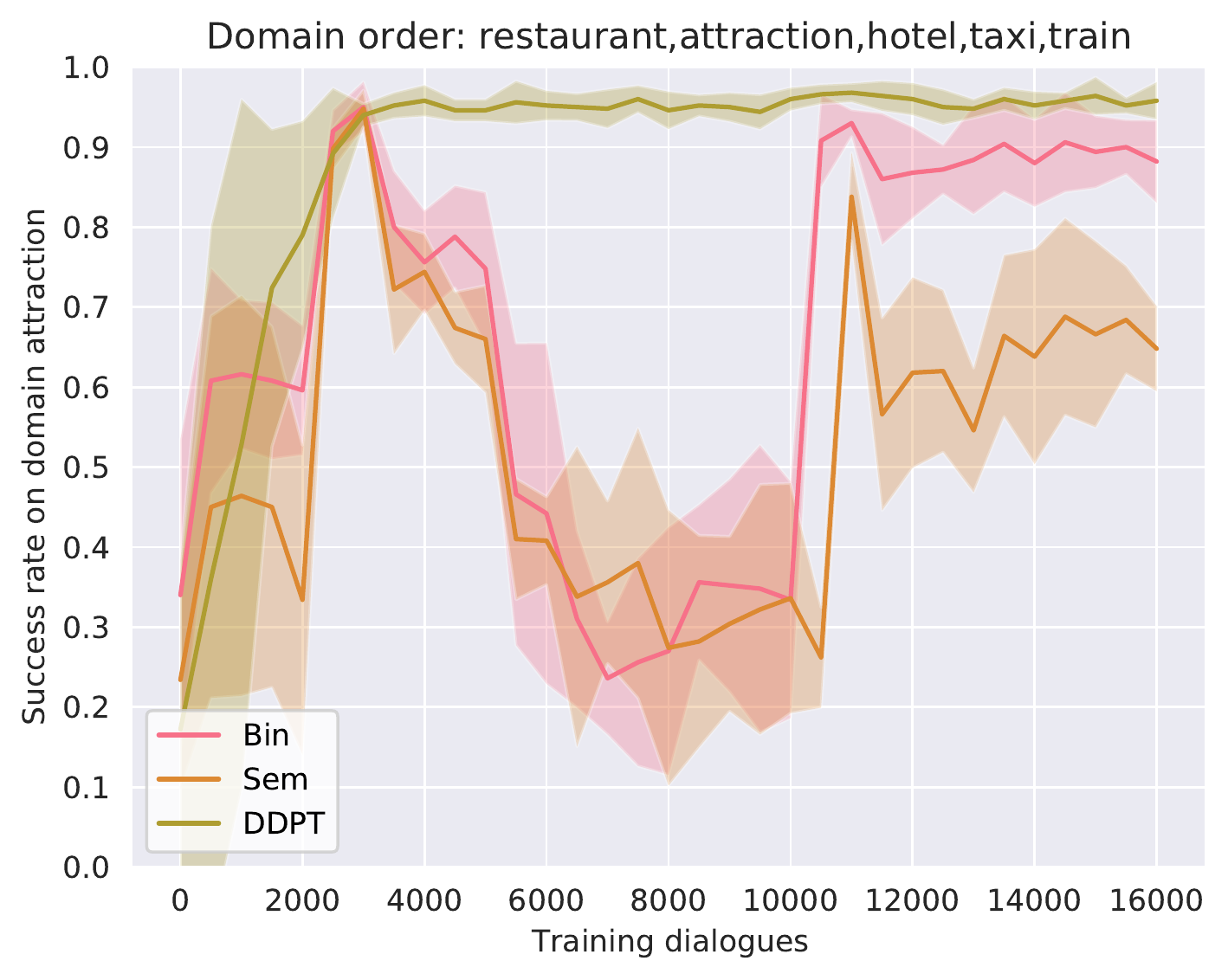}
    \caption{Success rate on attraction domain}
  \end{subfigure}
  \quad
  \begin{subfigure}[]{0.3\textwidth}
    \includegraphics[scale=0.37]{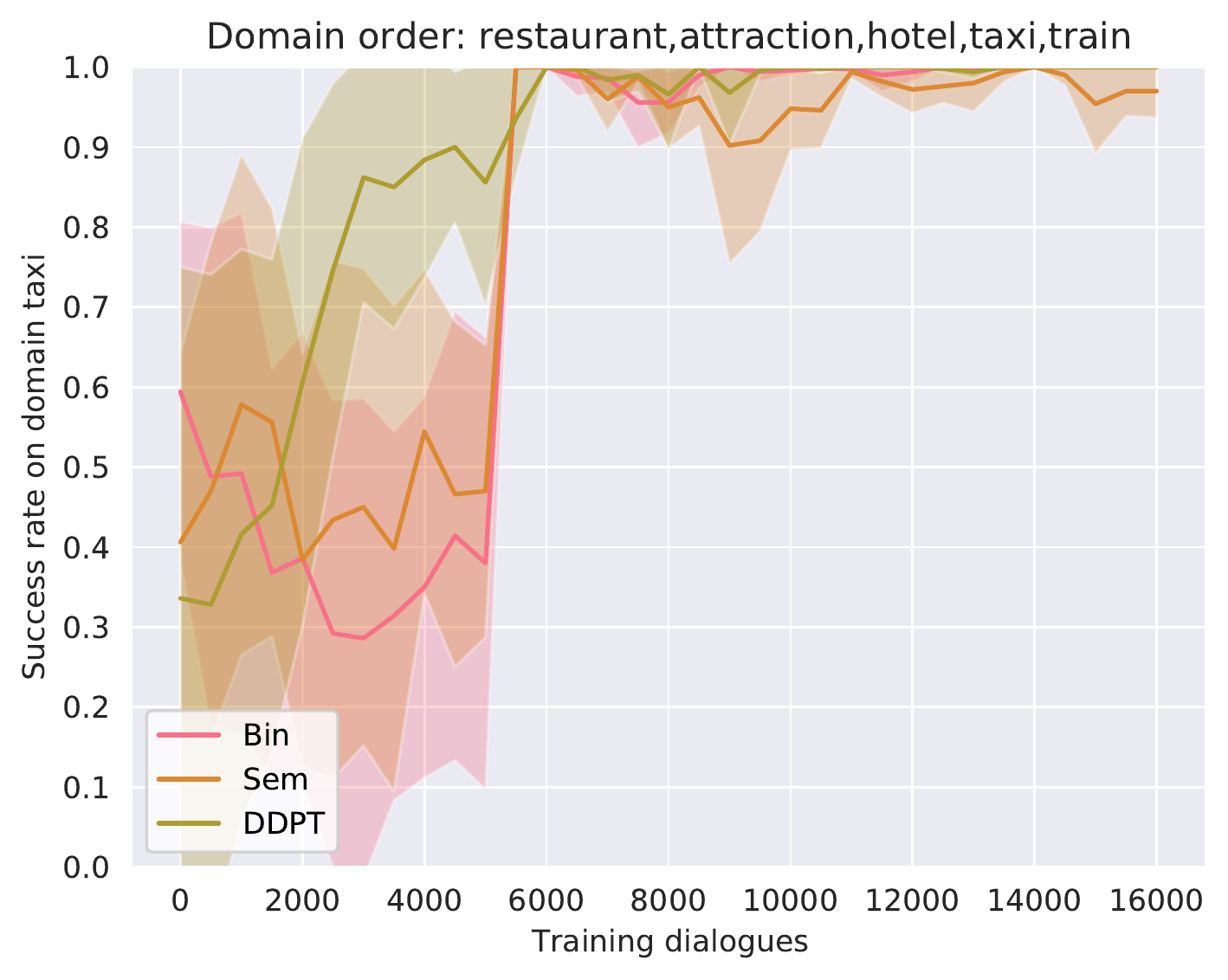}
    \caption{Success rate on taxi domain}
  \end{subfigure}
  \quad
  \begin{subfigure}[]{0.3\textwidth}
    \includegraphics[scale=0.37]{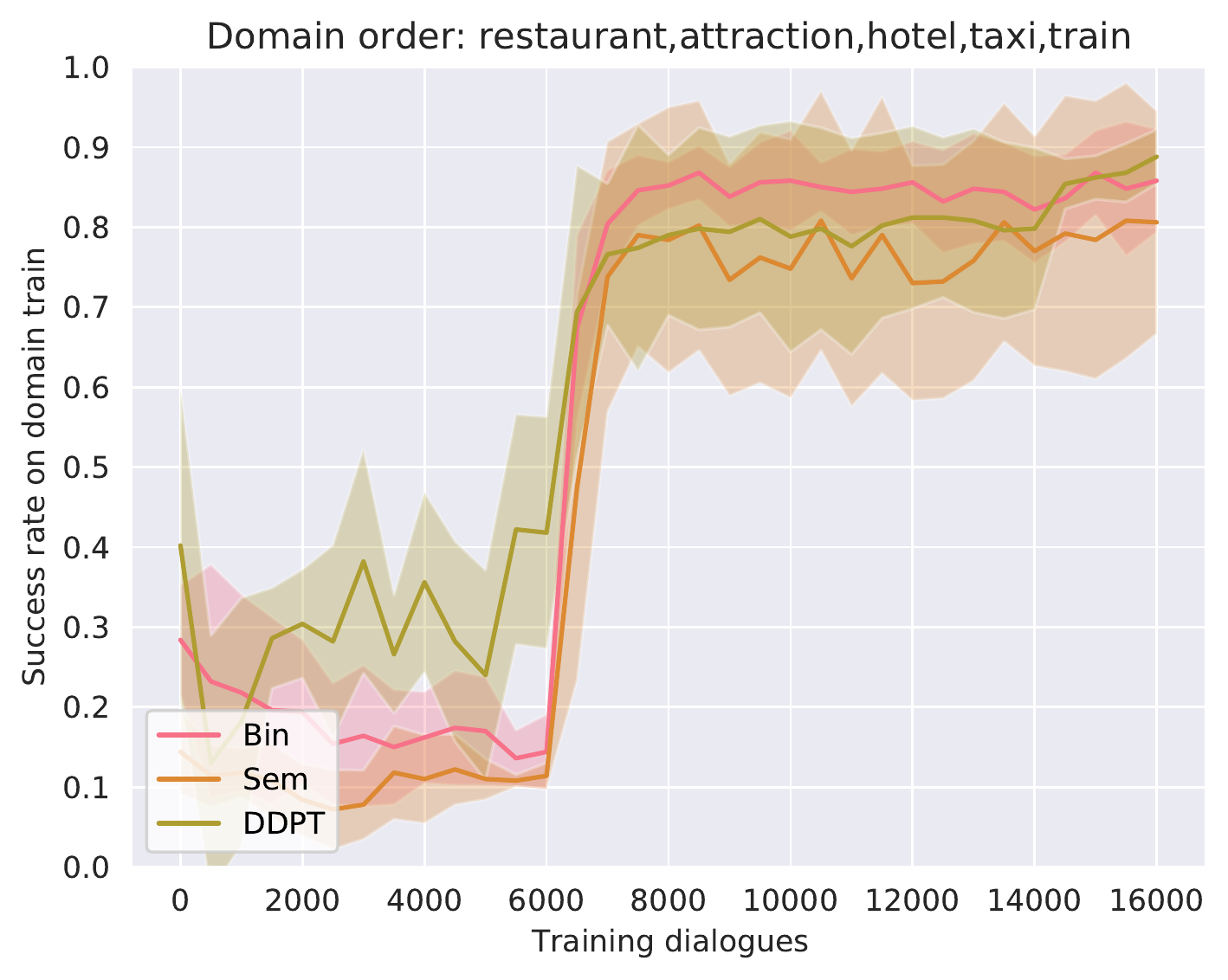}
    \caption{Success rate on train domain}
  \end{subfigure}
\\
  \begin{subfigure}[]{0.3\textwidth}
    \includegraphics[scale=0.37]{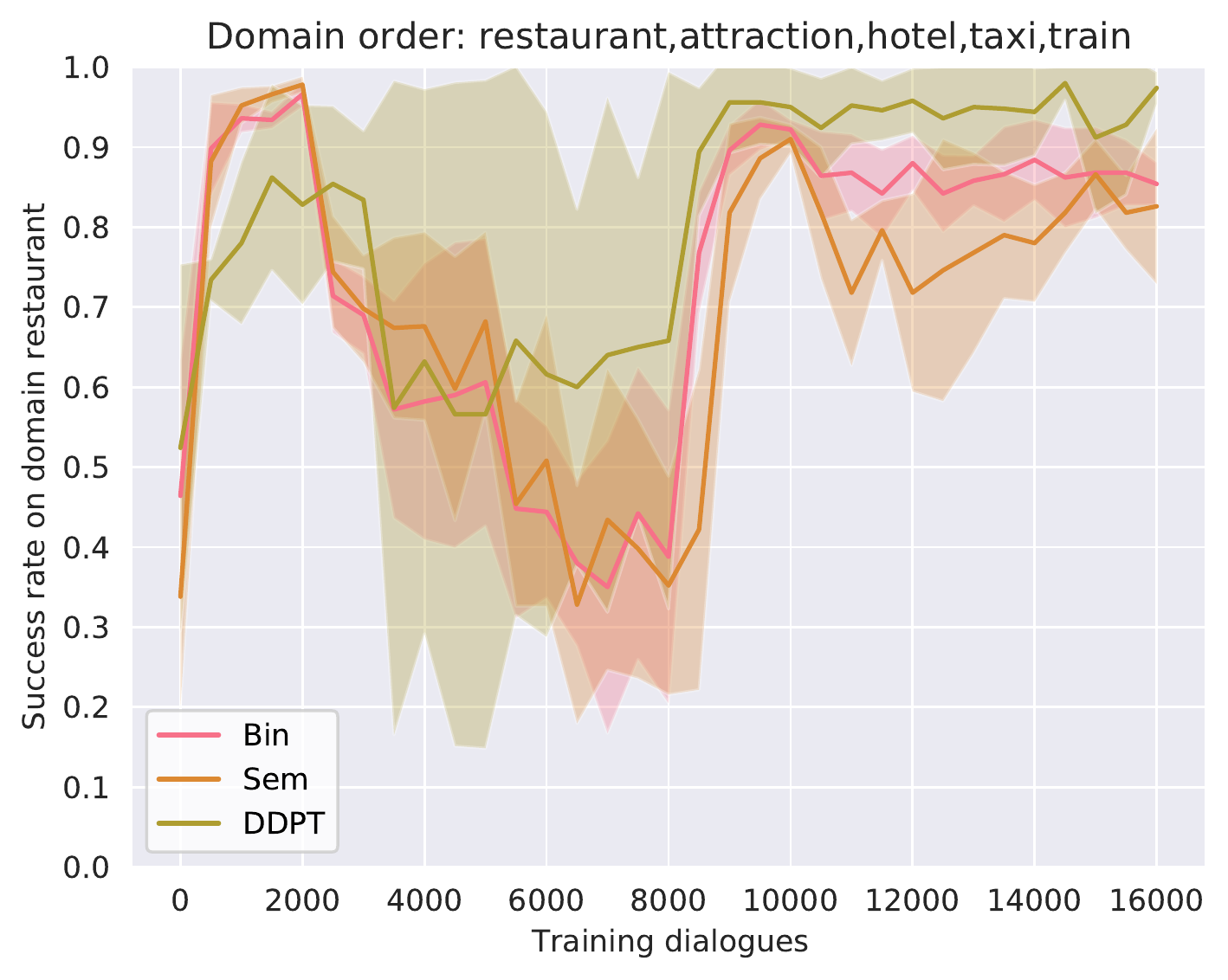}
    \caption{Success rate on restaurant domain}
  \end{subfigure}
  \quad
  \begin{subfigure}[]{0.3\textwidth}
    \includegraphics[scale=0.37]{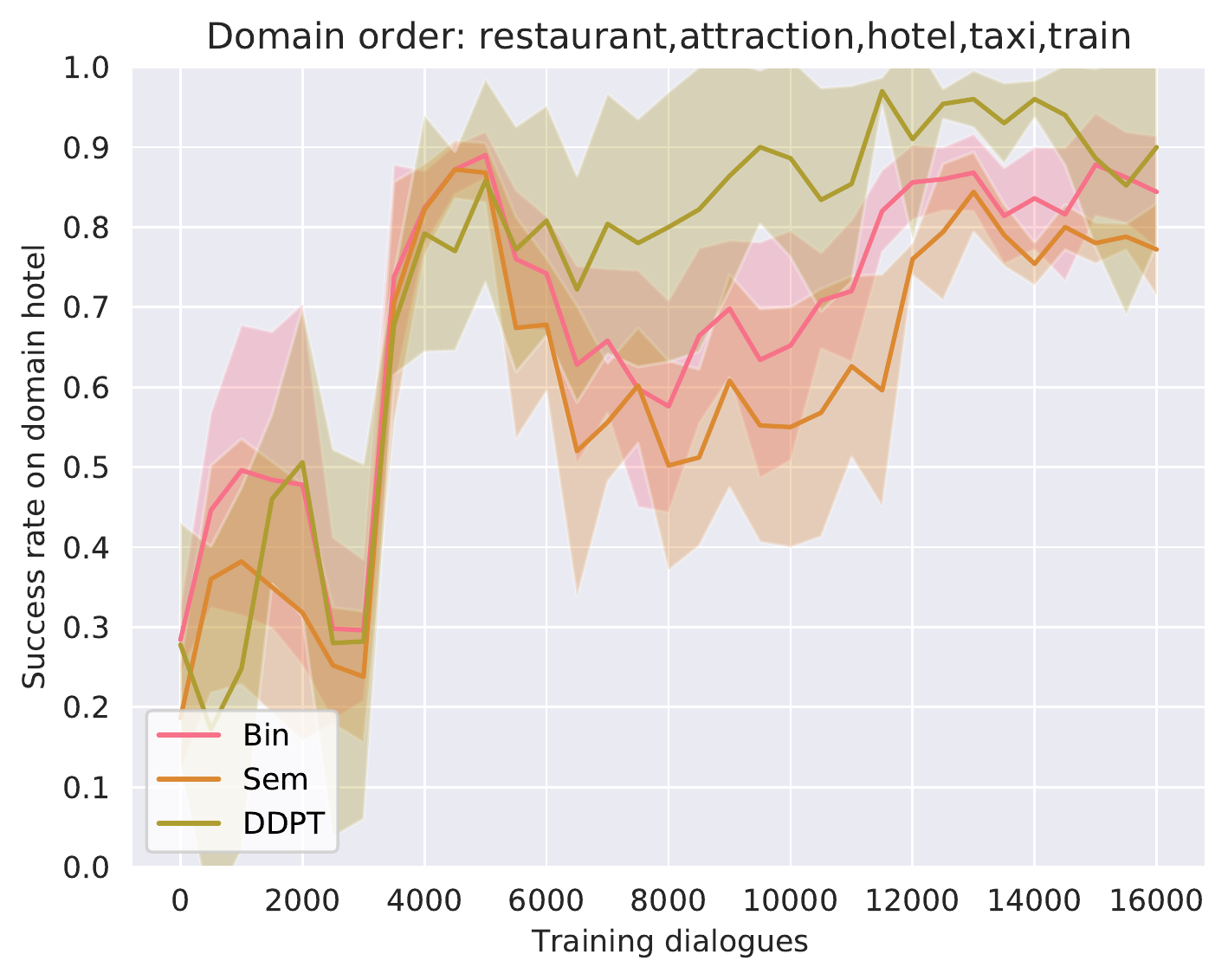}
    \caption{Success rate on hotel domain}
  \end{subfigure}

\caption{Success rate for each individual domain, where algorithms are trained in the order mixed.}
\label{}
\end{figure*}

\begin{figure*}[t!]
  \begin{subfigure}[]{0.3\textwidth}
    \includegraphics[scale=0.37]{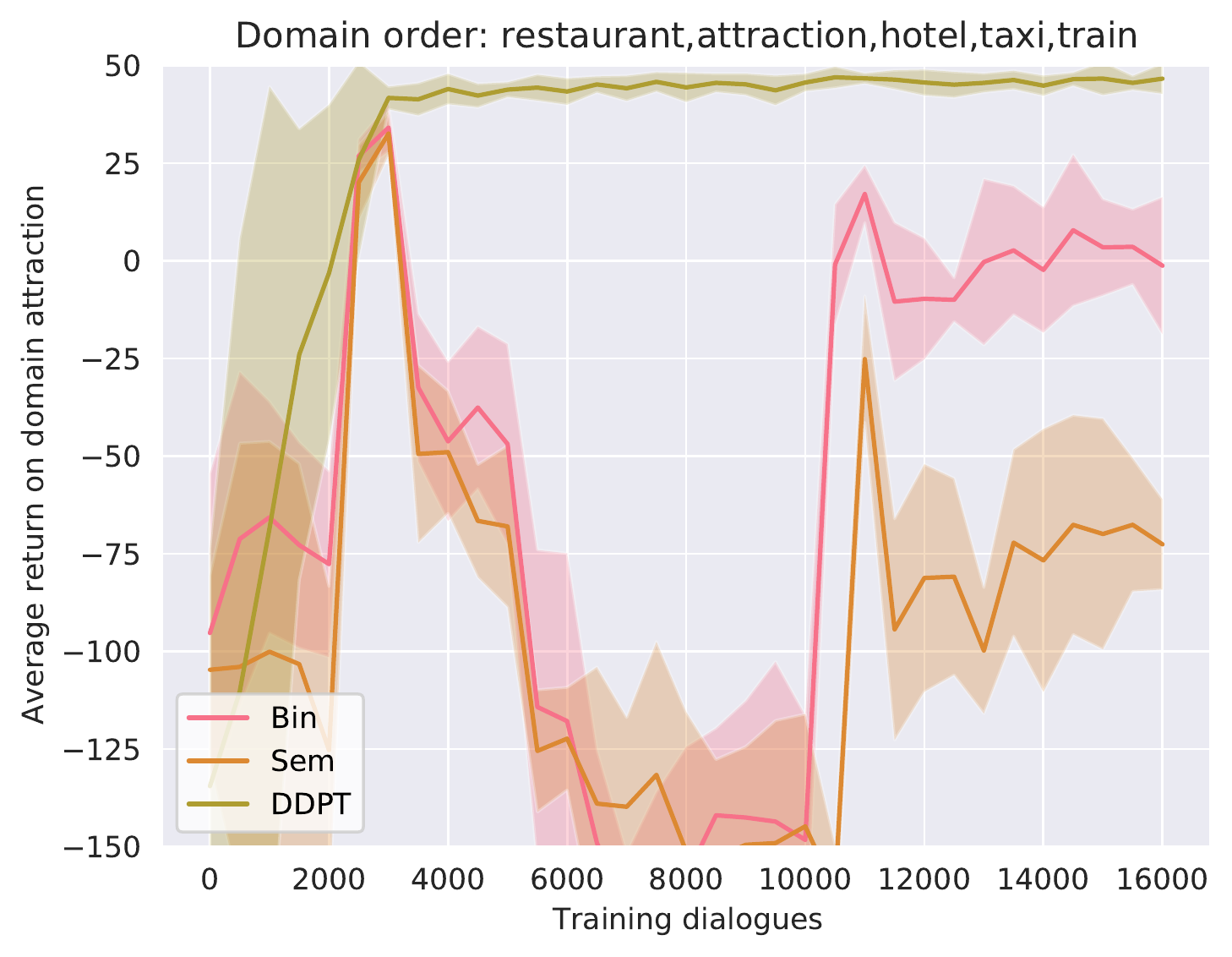}
    \caption{Average return on attraction domain}
  \end{subfigure}
  \quad
  \begin{subfigure}[]{0.3\textwidth}
    \includegraphics[scale=0.37]{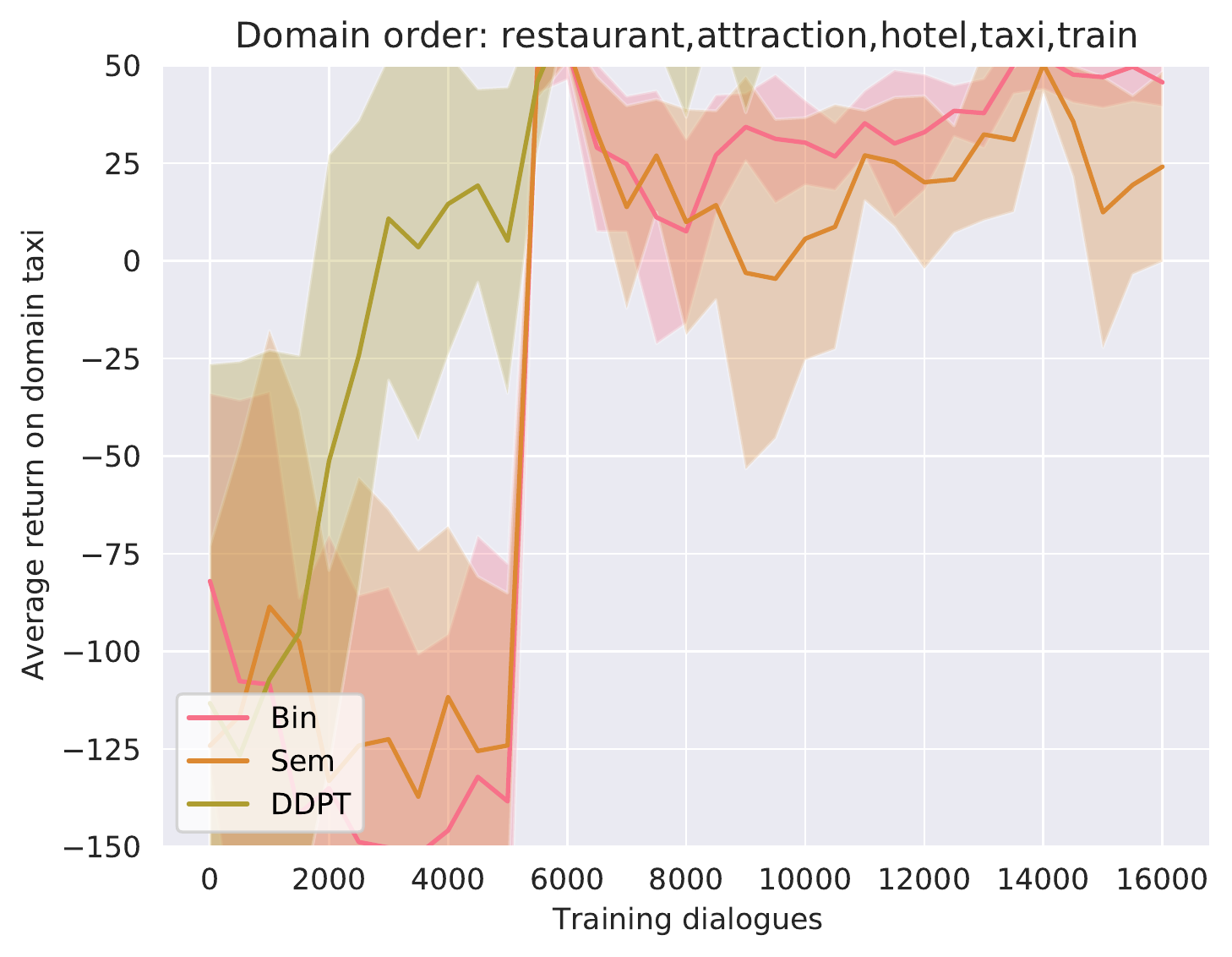}
    \caption{Average return on taxi domain}
  \end{subfigure}
  \quad
  \begin{subfigure}[]{0.3\textwidth}
    \includegraphics[scale=0.37]{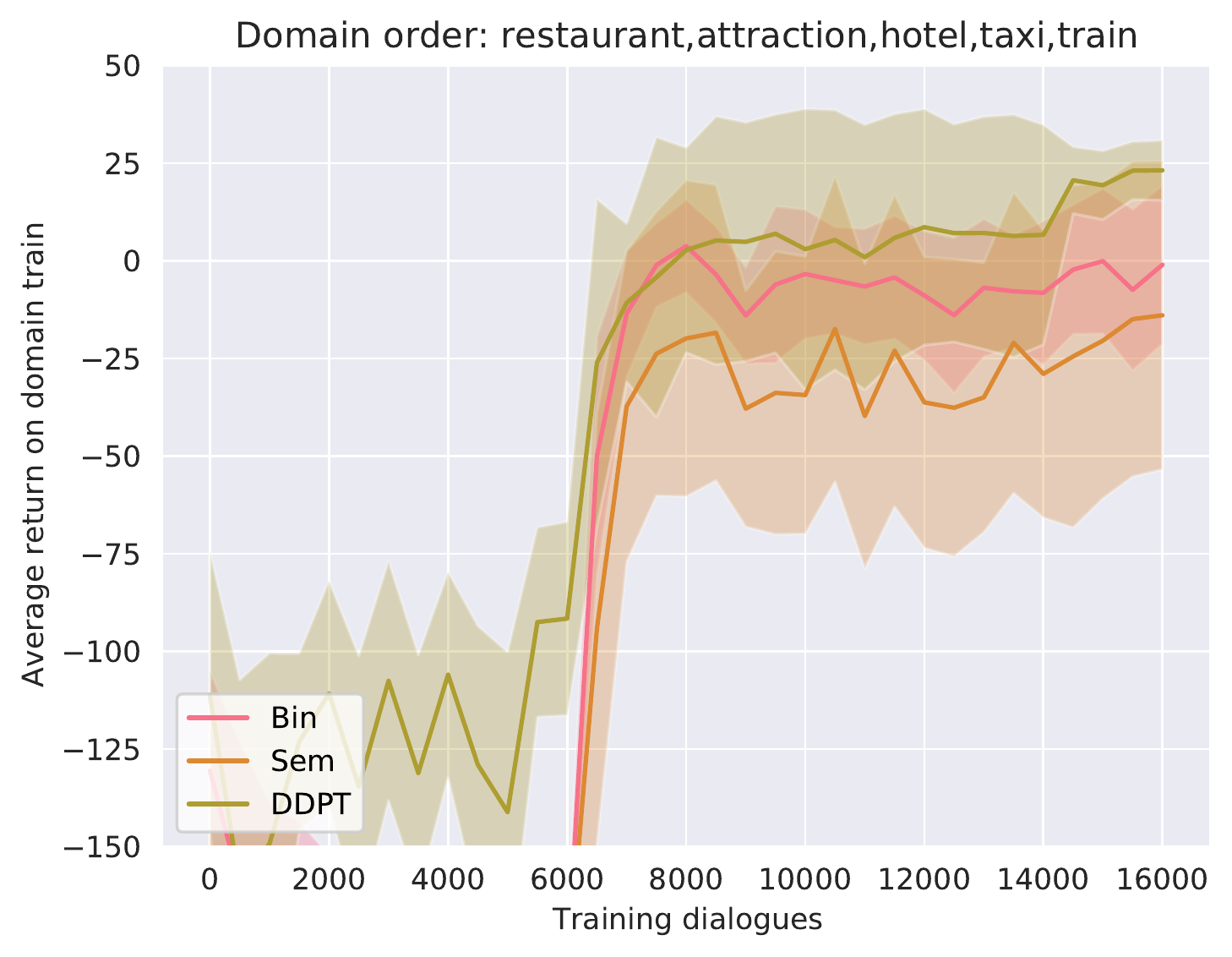}
    \caption{Average return on train domain}
  \end{subfigure}
\\
  \begin{subfigure}[]{0.3\textwidth}
    \includegraphics[scale=0.37]{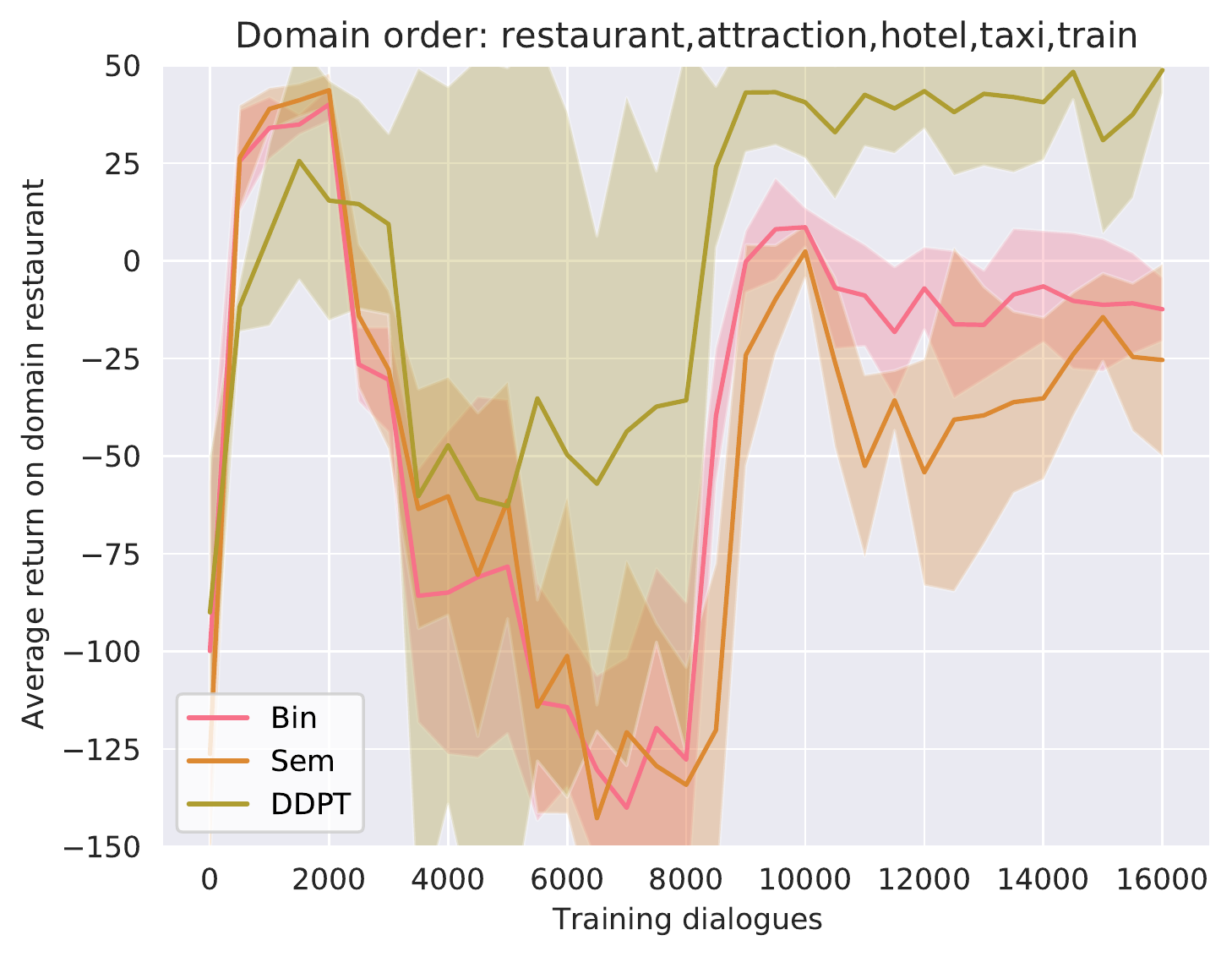}
    \caption{Average return on restaurant domain}
  \end{subfigure}
  \quad
  \begin{subfigure}[]{0.3\textwidth}
    \includegraphics[scale=0.37]{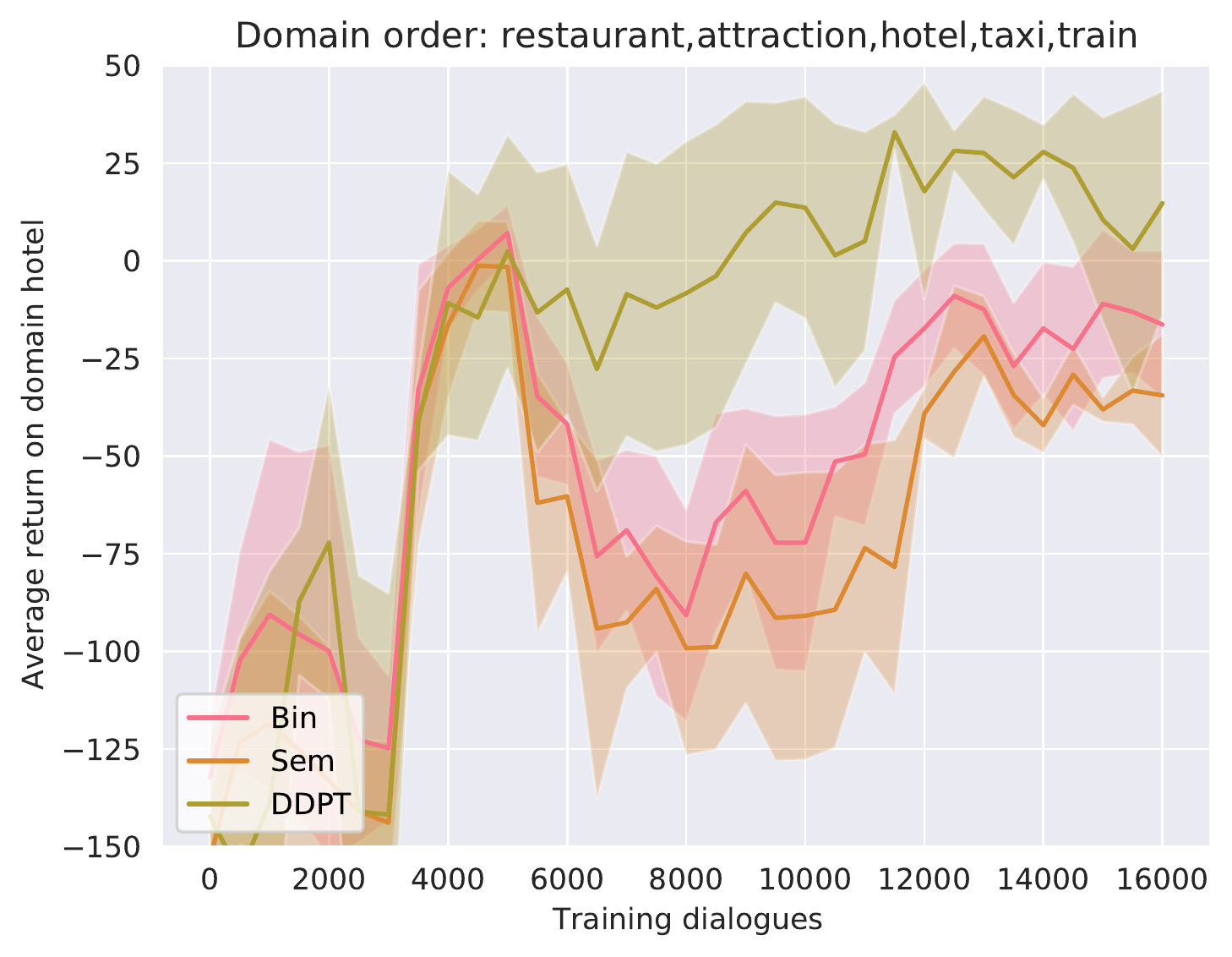}
    \caption{Average return on hotel domain}
  \end{subfigure}

\caption{Average return for each individual domain, where algorithms are trained in the order mixed.}
\label{}
\end{figure*}

\end{document}